Ain Shams University
Faculty of Engineering
Computer and Systems Engineering Department

# *Intelligent Systems for Information Security*

Submitted By

**Ayman Mohammad Bahaa Eldin Sadeq Albassal**

Master of Science
(Computer and Systems Engineering)
Ain Shams University, 1999

A THESIS
SUBMITTED IN PARTIAL FULFILMENT OF THE REQUIREMENTS
FOR THE DEGREE OF DOCTOR OF PHILOSOPHY
(Electrical Engineering)
DEPARTMENT OF COMPUTER AND SYSTEMS ENGINEERING

Supervised By

**Prof. DR. Abdel-Moneim A. Wahdan**

Cairo, Egypt
September, 2004

بسم الله الرحمن الرحيم

**بسم الله الرحمن الرحيم**

"قالوا سبحانك لا علم لنا إلا ما علمتنا إنك أنت العليم الحكيم"

**صدق الله العظيم البقرة 32**

**بسم الله الرحمن الرحيم**

"وقالوا الحمد لله الذي هدانا لهذا وما كنا لنهتدي لولا أن هدانا الله"

**صدق الله العظيم الأعراف 43**

**Faculty of Engineering**
**Computer and Systems Eng. Dept.**

## Examiners Committee

**Name** : **Ayman Mohammad Bahaa Eldin Sadeq Albassal**
**Thesis** : **Intelligent Systems for Information Security**
**Degree** : **Doctor of Philosophy**

| Name, Title, and Affiliate | Signature |
| --- | --- |
| | …………… |

Date: / /

# Abstract

**Intelligent Systems for Information Security**

**Ayman Mohammad Bahaa Eldin Sadeq Albassal**

**Doctor of Philosophy dissertation**

**Ain Shams University, 2004**

This thesis aims to use intelligent systems to extend and improve performance and security of cryptographic techniques as the cryptographic techniques are the main mechanisms used in information security. Information security goals and objectives are first studied then a historical background of cryptology is given. Focus is then turned into symmetric block ciphers. The design principles and how to measure a cipher security is then studied. Several examples of modern block ciphers are given. Cryptanalysis techniques and how they are classified are described. Details about differential cryptanalysis and interpolation cryptanalysis are presented. Genetic algorithms framework for cryptanalysis problem is addressed.

A novel extension to the differential cryptanalysis using genetic algorithm is proposed and a fitness measure based on the differential characteristics of the cipher being attacked is also proposed. The complexity of the proposed attack is shown to be less than quarter of normal differential cryptanalysis of the same cipher by applying the proposed attack to both the basic Substitution Permutation Network and the Feistel Network. The basic models of modern block ciphers are attacked instead of actual cipher to prove that the attack is applicable to other ciphers vulnerable to differential cryptanalysis.

A new attack for block cipher based on the ability of neural networks to perform an approximation of mapping is proposed. A complete problem formulation is explained and implementation of the attack on some hypothetical Feistel cipher not vulnerable to differential or linear attacks is presented.

A new block cipher based on the neural networks is proposed. A complete cipher structure is given and a key scheduling is also shown. The main properties of neural network being able to perform mapping between large dimension domains in a very fast and a very small memory compared to S-Boxes is used as a base for the cipher. A fully connected structure is also used to produce large confusion and diffusion effects. Two key dependent schemes are presented and compared.
All proposals are assessed by several experimentation and selected tests and test results are presented.

# Acknowledgements

First of all I thank Allah for all his generosity and help.

Second I would like to express my deep gratitude to Prof. Abdel-Moneim Wahdan for suggestion of the point of research, valuable advises, and his great effort revising the thesis. I can not express my thanks to him for his care and interest since I was an undergraduate student and his guiding me and my fellows as his students. Throughout the years I have worked in the department of computer and systems engineering, supervising my master thesis, he was a real model for a university professor.  He actually gave me the chance to express my capabilities and to put my first step in scientific research. All my hope is that I can follow his steps.

I would like also to thank Prof. Mohammad Adeeb Elghonaimy for his great efforts along the thesis development since it was just thoughts in mind. I would like to thank him for his great support and all the materials and references he gave to me.

I would like to thank my teachers, Staff of Computer and Systems Engineering Department for all what they taught me and also all my fellows for their support and encouragement.

I would like also to thank my parents and my mother in law and all my family for their deep prays and support for me during the thesis development.

Finally I would like to thank my beloved wife for her patience and great support and encouragement during the most important stages of the thesis.

# Statement

This dissertation is submitted to Ain Shams University for the degree of Doctor of Philosophy in Electrical Engineering, Computer and Systems

The work included in this thesis was out by the author at Computer and systems Engineering Department, Ain Shams University.

No part of this thesis has been submitted for a degree or qualification at other university or institution.

Date : / /200

Signature :

**Name** **:** Ayman Mohammad Bahaa Eldin Sadeq Albassal

## Table of Contents

# List of Tables

# List of Figures and Illustrations

# List of Symbols, Abbreviations and Nomenclature

| | |
|---|---|
| AES | Advanced Encryption Standard |
| CBC | Cipher Block Chaining Mode |
| CFB | Cipher Feedback Mode |
| Cipher | An encryption / decryption System |
| CTR | Counter Mode |
| DES | Data Encryption Standard |
| ECB | Electronic Code Book Mode |
| FIPS | Federal Information Processing Standard |
| GA | Genetic Algorithms |
| GF | Galois Fields |
| ISO | International Standardization Organization |
| ITU-T | International Telecommunication Union, Telecommunication Standardization Sector |
| NIST | National Institute for Standards and Technologies |
| NN | Neural Network |
| OFB | Output Feedback Mode |
| OSI | Open Systems Interconnection |
| S-Box | Substitution Box |

# Chapter 1: Introduction

Information security has always been a concern since old times. Secret talking was a key concept in ancient and modern wars.

In recent days, information security is needed in most commercial and legal operations performed in a modern society. Information security tools and practices change in many ways according to the situation and requirements. Information security can be categorized by the goals to be achieved during a transaction or a transmission.

The study of information security involves the study of security services, mechanisms, and attacks. Information security services are detailed in section 1.1, while a set of attacks especially on block ciphers are studied, and actually improved through novel proposals, given in the next chapters of the thesis. A mechanism is a means, a method or a system that is designed to detect, prevent, or recover from a security attack. There is no single mechanism that will support all security goals required; however one particular element underlies many of the security mechanisms in use namely cryptographic techniques.

***Cryptography*** is a Greek word that is composed of two halves, ***kryptos*** and ***graphein***. The first word, kryptos, means hidden or secret and the other word, graphein, means writing. So cryptography means secret writing. As mentioned, cryptographic techniques are the most important mechanisms to achieve information security goals. This thesis focuses on the study of cryptography and cryptanalysis that make together the science of ***cryptology***. The thesis goal is to propose new techniques using artificial intelligence to extend the cryptology mechanisms and to overcome some limitations in the classical approaches. This chapter gives a general introduction to the concepts of information security and cryptography. Then the objective and motivations of the thesis and also the methodologies used in the novel proposals of the thesis are given. Finally the structure of the thesis is outlined.

## 1.1 Cryptography and Information Security

Regardless of who is involved, to one degree or another, all parties to a transaction involving some information or raw data must have confidence that certain objectives associated with information security have been met [1]. Table 1.1 presents some possible goals, or security services, which may be involved in a secured information processing [1]. It can be noticed that information security involves many goals and hence a huge number of techniques, polices, procedures and mathematics are typically used to achieve such goals.

Over the centuries, an elaborate set of protocols and mechanisms has been created to deal with information security issues when the information is conveyed by physical documents. Often the objectives of information security cannot solely be achieved through mathematical algorithms and protocols alone, but require procedural techniques and obedience of laws to achieve the desired result.

For example, privacy of letters is provided by sealed envelopes delivered by an accepted mail service. The physical security of the envelope is, for practical necessity, limited and so laws are enforced which make it a criminal activity to open mail for which one is not authorized. It is sometimes the case that security is achieved not through the information itself but through the physical document recording it. For example, paper currency requires special inks and material to prevent counterfeiting [1].

**Table 1.1: Some information security objectives**

| | |
|---|---|
| Privacy or confidentiality | Keeping information secret from all but those who are authorized to see it. |
| Data integrity | Ensuring information has not been altered by unauthorized or unknown means. |
| Entity authentication or identification | Corroboration of the identity of an entity (e.g., a person, a computer terminal, a credit card, etc.). |
| Message authentication | Corroborating the source of information; also known as data origin authentication. |
| Signature | A means to bind information to an entity authorization |
| Validation | A means to provide timeliness of authorization to use or manipulate information. |
| Access control | Restricting access to resources to privileged entities. |
| Certification | Endorsement of information by a trusted entity. |
| Time stamping | Recording the time of creation or existence of information. |
| Witnessing | Verifying the creation or existence of information by an entity other than the creator. |
| Receipt | Acknowledgement that information has been received. |
| Confirmation | Acknowledgement that services have been provided. |
| Ownership | A means to provide an entity with the legal right to use or transfer a resource to others. |
| Anonymity | Concealing the identity of an entity involved in some process. |
| non-repudiation | Preventing the denial of previous commitments or actions. |
| Revocation | Retraction of certification or authorization. |

Most of the information security methods can be classified into two main categories

1. Carrying, or transmission, medium related methods
2. Information related methods

While the former methods were widely used until recent days, the later methods are the modern ways for information security.

It has been a great turn in information security techniques when the physical medium carrying the information changed from visual media,

like stones, leather tissues and papers to electronic media. Much of information now resides on magnetic media and is transmitted via telecommunications systems, some are wireless.

What has changed dramatically is the ability to copy and alter information. One can make thousands of identical copies of a piece of information stored electronically and each is indistinguishable from the original. With information on paper, this is much more difficult. What is needed then where information is mostly stored and transmitted in electronic form is a means to ensure information security which is independent of the physical medium recording or conveying it and such that the objectives of information security rely solely on digital information itself [1].

Among all information security methods and techniques, ***cryptography*** is always the most powerful and widely used in most application.

*Cryptography* is the study of mathematical techniques related to aspects of information security such as confidentiality, data integrity, entity authentication, and data origin authentication, and non-repudiation [2] and [1].

The previous definition involved 4 of the main services provided by information security systems and hence desire detailed definition rather than the short one presented in table 1.1. The definitions of these four terms are also taken form [1]

1. *Confidentiality* is a service used to keep the content of information from all but those authorized to have it. *Secrecy* is a term synonymous with confidentiality and privacy. Secrecy means the information is not generally known, it may not involve authorization for access, and this term is hence not widely known. Privacy means to keep the information secret from all but the source of it. There are numerous approaches to providing confidentiality, ranging from physical protection to mathematical algorithms which render data non interpretable.
2. *Data integrity* is a service which addresses the unauthorized alteration of data. To assure data integrity, one must have the

ability to detect data manipulation by unauthorized parties. Data manipulation includes such things as insertion, deletion, and substitution.

3. *Authentication* is a service related to identification. This function applies to both entities and information itself. Two parties entering into a communication should identify each other. Information delivered over a channel should be authenticated as to origin, date of origin, data content, time sent, etc. For these reasons this aspect of cryptography is usually subdivided into two major classes: *entity authentication* and *data origin authenticatio*n. Data origin authentication implicitly provides data integrity.
4. *Non-repudiation* is a service which prevents an entity from denying previous commitments or actions. When disputes arise due to an entity denying that certain actions were taken, a means to resolve the situation is necessary. For example, one entity may authorize the purchase of property by another entity and later deny such authorization was granted. A procedure involving a trusted third party is needed to resolve the dispute.

The other services mentioned in table 1.1 depend on these basic four services. The information security services concept has been standardized in the ITU-T X.800 Security Architecture for OSI. The X.800 standard defines services, mechanisms and attacks. Security service is defined as a service provided by a protocol layer of communicating open systems, which ensures adequate security of the systems of data transfer [2].

The X.800 adds another basic service other than the mentioned four that is Access Control which can be defined as controlling who has access to a resource, under what condition that access can occur and what is exactly the accessing entity can do. It can be easily seen that this service can be mixed with the authentication service, but it is not the same while it is based on the authentication service as the starting point of performing access control.

## 1.2 Thesis Motivations and Objectives

Studies in the cryptography and cryptanalysis fields are enormous and very intensive. The field is little bit complex and is a challenge, especially to engineers. Each country in the world tries to have its own information security technology as this type of technology must be local. It is not of any wise to secure your information giving the others the keys to your security facilities. This is exactly what happen when such information security is obtained from outside without a deep knowledge of how it works.

It is not only national security requirements that urge the study of information security but also many new commercial applications and new digital society approach make it very important to provide new techniques and mechanisms suitable for each application.

It was chosen to study the science itself instead of its application as this is the logical start point for a valuable contribution. Cryptology also contains a solid mathematical and theoretical foundation opening a wide prospect for research.

Most of the studies in cryptology are either built on statistical studies or very solid mathematical foundations which points out a very good place for intelligent system to be introduced. Genetic algorithms are very attractive search tools that need careful and deep study to fit them in the scope of cryptanalysis and key recovery. Also neural networks are nonlinear systems that can perform mapping from a domain to another and can be used also to approximate such mappings. Hence neural networks can be used as cryptanalysis tools and also can be used as a building block in block ciphers.

The objectives of this thesis are to propose novel techniques and suitable frameworks for using intelligent systems for information security.

### 1.3 Thesis Methodology

The thesis methodology can be summarized in the following points

1. Deep and detailed study of both cryptography and cryptanalysis.
2. Study of week points and points of improvement of both cryptanalysis and cryptography.
3. Detailed study of two successful attacks on block ciphers.
4. Introducing a framework of using genetic algorithms in cryptanalysis
5. Proposing a complete formulation of the key search cryptanalysis problem as a genetic algorithm optimal key search problem
6. Proposing a neural network based extension to the interpolation attack.
7. Testing the proposed attacks through a large number of experimentation on the two basic models of block ciphers instead of a certain category of actual block ciphers so as to be able to extend the attack to any block cipher based on those basic models.
8. Proposing a new cipher based on the use of neural network as a core function and providing details about its security and performance

### 1.4 Organization of the Thesis

The thesis is organized into 7 chapters and 2 appendices. This chapter, chapter one, gives a general introduction and details the motivations, objectives and methodology used while working through out the research.

Chapter two begins with a historical background of ciphers and then it gives details about symmetric block ciphers and their operation and introduces basic concepts of cryptology. Three common used block ciphers are detailed at the end of the chapter.

Chapter three descries the concept of cryptanalysis providing details about two major cryptanalysis techniques, namely differential and interpolation attacks.

In chapter four a proposal for using genetic algorithm as an extension to differential cryptanalysis is given. The proposed attack is applied to both the basic Substitution Permutation Network (SPN) and the Feistel Network (FN).

Chapter five details the proposal of using neural networks, namely back propagation networks, as a cryptanalysis tools that outperforms classical interpolation attacks. Example of applying the attack on a hypothetical cipher with a core function from the Rijndael cipher is given.

Chapter six describes the proposal of using neural networks as a highly nonlinear, confusing component with large input and output sizes for better security.

Chapter seven finally gives the conclusions of the thesis and a list of examples of future work points

Appendix A contains a brief introduction to genetic algorithms and neural networks while appendix B gives some  background for the mathematics used through out the thesis.

# Chapter 2: An Overview of Block Ciphers

Symmetric block ciphers have always been the most famous and important element in many cryptographic systems. Individually, they provide confidentiality paradigm but their flexibility allows them to be used as fundamental building blocks in the construction of pseudorandom number generators, stream ciphers, and hash functions. They may furthermore serve as a central component in message authentication techniques, data integrity mechanisms, entity authentication protocols, and (symmetric-key) digital signature schemes.

No block cipher is ideally suited for all applications, even one offering a high level of security. This is a result of expected tradeoffs required in practical applications, including those arising from speed requirements and memory limitations, and also constraints imposed by implementation platforms and application constraints [1].

The main design principle behind most of the block ciphers are based on Claude Shannon principles of confusion and diffusion. Also the concept of product cipher that is building complicated cipher using simpler building blocks is a key concept in block cipher design. The works of Feistel are the major starting points in modern block ciphers design.

The following sections describe the concepts and details of block ciphers. First a historical background is given then a detailed description of the basic two models used in most modern block ciphers is illustrated. Modes of operations and their effects on the security and performance of block ciphers are then detailed. Finally three examples of the most widely used modern block ciphers are detailed.

## 2.1 Historical Background of cryptology

### 2.1.1 Cryptographic Model

By cryptology we mean the study of both cryptography and cryptanalysis. Cryptography was previously defined and includes many processes, like encryption, decryption and hashing.

Cryptanalysis forms the measuring tool to judge how secure a cryptographic algorithm is.

Before going into further details let us give some definitions.

***Plaintext*** is the information that can be interpreted for the purpose it is originally intended to.

The term text should not be confusing as the definition as given here is a general for any type of information in digital format. It can include sound or video for example. The term text is inherited from the ancient use of cryptography.

***Ciphertext*** is the information in a form that can not be interpreted until it is converted back into its original form.

***Encryption or Encipherment*** is the process of converting plaintext into ciphertext. In most cases the encryption process is characterized by a key. The encryption can be viewed as a transformation from the plaint text domain into the ciphertext domain $E_e : e \in K, C = E_e(P)$ where E is the encryption transformation, P is the plaintext, C is the ciphertext, e is the key and K is the key space.

***Decryption or Decipherment*** is the process of converting ciphertext back into plaintext.

Also decryption can be viewed as a transformation from the ciphertext domain into the plaintext domain characterized by a key $D_d : d \in K, P = D_d(C)$ where D is the decryption transformation, P is the plaintext, C is the ciphertext, d is the key and K is the key space.

***Cipher*** is an algorithm that defines both encryption and decryption processes that the two processes are inverse of each other that is $P = D_d(E_e(P))$

By knowing the key it is always ***computationally easy*** to perform any of the transformations but if the key is unknown it must be very ***difficult*** or impossible to perform any of the transformations, encryption or decryption.

Figure 2.1 shows a model of using a cipher to achieve secured communication

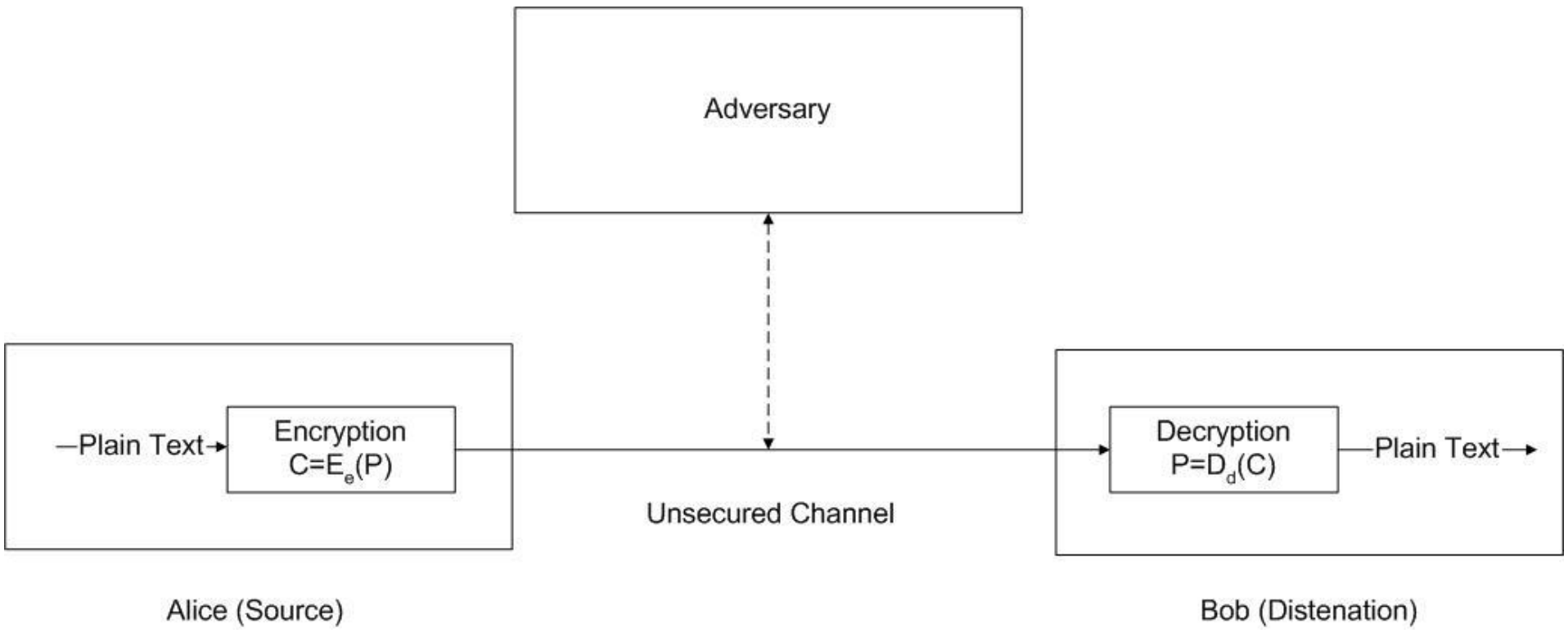

Figure 2.1: A SECURED COMMUNICATION MODEL USING A CIPHER

As can be seen from the figure, the communication is done over a public unsecured channel where others, adversary, can monitor, or even has a complete copy of, the data transferred.

The communication is secured by using a reversible encryption scheme, where the source, Alice, encrypts the output message using the cipher E characterized by the key e. The receiver, Bob, decrypts the message he receives by the inverse operation, decryption, D using the decryption key d. The adversary who does not have the key d can't decrypt the transfer and hence can not interpret it.

The entire security of the model relies on the security of the cipher. A cipher security can be defined by the effort, time and computational resources, used to break the cipher. An encryption scheme is said to be *breakable* if a third party, without prior knowledge of the key pair (e, d), can systematically recover plaintext from corresponding ciphertext within some appropriate time frame.

Formal definitions of security include two of them [2].

***Unconditional security***: A cipher is said to be unconditionally secure if it is impossible to decrypt a ciphertext generated by the cipher scheme no matter how much time and computational resources are available for an adversary. Except for the one time pad or Vernam cipher, there is no cipher scheme that is unconditionally secure.

***Computational security***: A cipher is said to be computationally secure if it is infeasible by using the available computational resources to decrypt a ciphertext generated by the cipher in a suitable time frame that the decrypted text is useful. Most of the block cipher designers seek a suitable computational security for their ciphers.

***Cryptanalysis*** is the study of mathematical techniques for attempting to defeat cryptographic techniques.

A ***cryptanalyst*** is someone who engages in cryptanalysis.

A ***cryptosystem*** is a general term referring to a set of cryptographic primitives used to provide information security services. Most often the term is used in conjunction with primitives providing confidentiality, i.e., encryption.

### 2.1.2 Symmetric Ciphers

Consider an encryption scheme consisting of the sets of encryption and decryption transformations $E_e : e \in K$ and $D_d : d \in K$ , respectively, where K is the key space. The encryption scheme is said to be *symmetric-key* if for each associated encryption/decryption key pair (e, d), it is computationally "easy" to determine d knowing only e, and to determine e from d only.

Since e = d in most practical symmetric-key encryption schemes, the term symmetric-key becomes appropriate [1].

Symmetric ciphers were the only used methods in cryptography until the discovery of public key cryptosystem in the end of 1970s.

Symmetric ciphers is also known as secret key ciphers as the key is always kept secret for the security of the system, opposed to public key cipher where one of the key pair is known to the public where the other, private key, is kept secret.

Some literature also calls symmetric key cipher as private key ciphers, this naming should not to be used as it is little bit confusing with the private key element that must be kept secret in the public key cryptographic systems and hence it is not going to be used in the scope of this work.

### 2.1.3 Classical Ciphers

Cryptography is used from very ancient ages. The Egyptians may be the first to use cryptography basically when they invented writing. The ancient written language was a secret one known only to religious men and magicians. Symbols were used to represent abstract meanings and thoughts. In literature, it is often referred to Julius Caesar as the one who invented the first cipher known as Caesar Cipher.

It is worth to introduce here two more definitions

***A block cipher*** is an encryption scheme which breaks up the plaintext messages to be transmitted into strings (called *block*s) of a fixed length t over an alphabet A, and encrypts one block at a time.

***Substitution ciphers*** are block ciphers which replace symbols (or groups of symbols) by other symbols or groups of symbols.

#### *2.1.3.1 Caesar Cipher*

Caesar cipher is basically a substitution symmetric block cipher where the alphabet is used as the domain of both the plaintext and ciphertext.

Original Caesar cipher substitutes each alphabet character with one three position after, for example the word TEXT is encrypted into WHAW,

note that the character X is mapped to A. This classical cipher can be modeled by the following equations

$$C = P + K \mod s \tag{2.1}$$

where P is the original letter from the alphabet K is the key and s is the number of letters in the alphabet. The decryption is defined as

$$P = C - K \mod s \tag{2.2}$$

In his cipher, Caesar used K=3 and s is 26 for English language.
Caesar cipher uses a block size of one letter from the alphabet.
Caesar cipher is not secure as we have only 26 mappings (one of them is trivial where k=0).
As the key space is very limited the cipher can be broken by using a ***brute force attack*** where all the possible values of the key is tested until a meaningful text is obtained.

#### *2.1.3.2 Mono-alphabetic cipher*

Caesar cipher can be extended into what is so called a simple substitution cipher or a mono-alphabetic cipher where the key space is extended to be very large

Let A be an alphabet of q symbols and M be the set of all strings of length t over A. Let K be the set of all permutations on the set A. Define for each $e \in K$ an encryption transformation $E_e$ as:

$$E_e(m) = (e(m_1)e(m_2) \ldots e(m_t)) = (c_1\ c_2 \ldots c_t) = c; \tag{2.3}$$

Where $m = (m_1\ m_2 \ldots m_t) \in M$. In other words, for each symbol in a t-tuple, replace (substitute) it by another symbol from A according to some fixed permutation e. To decrypt $c = (c_1\ c_2 \ldots c_t)$ compute the inverse permutation $d = e^{-1}$ and

$$D_d(c) = (d(c_1)d(c_2) \ldots d(c_t)) = (m_1\ m_2 \ldots m_t) = m \tag{2.4}$$

$E_e$ is called a *simple substitution cipher* or a *mono-alphabetic substitution cipher*.

So rather than just shifting the alphabet, it is possible to shuffle (jumble) the letters arbitrarily. Each plaintext letter maps to a different random ciphertext letter, hence key is 26 letters long.

For example

Plain: ABCDEFGHIJKLMNOPQRSTUVWXYZ

Cipher:DKVQFIBJWPESCXHTMYAUOLRGZN

Then a message *"if we wish to replace letters"* is going to be ciphered as follows

Plaintext: IFWEWISHTOREPLACELETTERS

Ciphertext: WIRFRWAJUHYFTSDVFSFUUFYA

Now having a total of 26! = 4 x $10^{26}$ keys which is a very large key space making brute force attack non practical. With so many keys we might think mono alphabetic cipher is secure but would be wrong. The reader is referred to chapter 3 for more details. An attack based on language characteristics can be mounted on the cipher with a complexity far below the brute force one.

Human languages are redundant, that is letters are not equally commonly used. For example in English e is by far the most common letter then T,R,N,I,O,A,S other letters are fairly rare e.g. Z,J,K,Q,X

We can build tables of single, double & triple letter frequencies in any language from large common conversations or text and have a common letter frequency chart that can be used to attack the cipher.

Although the mono alphabetic cipher has a large key space, each character is always mapped into a certain other character from the alphabet so the statistical nature of the language is not changed from the plaintext to the ciphertext.

Building a chart of the letter frequency of the ciphertext and mapping it to the original chart can be used to discover the original text. The frequency is not only for single letters but it can also be calculated for pairs of letters, in English "th" is the most common diagram, and also group of three letters frequency can be calculated, in English "the" is the most common trigram. From this information an attack can be successful. Figure 2.2 shows a frequency chart for each letter in English language [2]

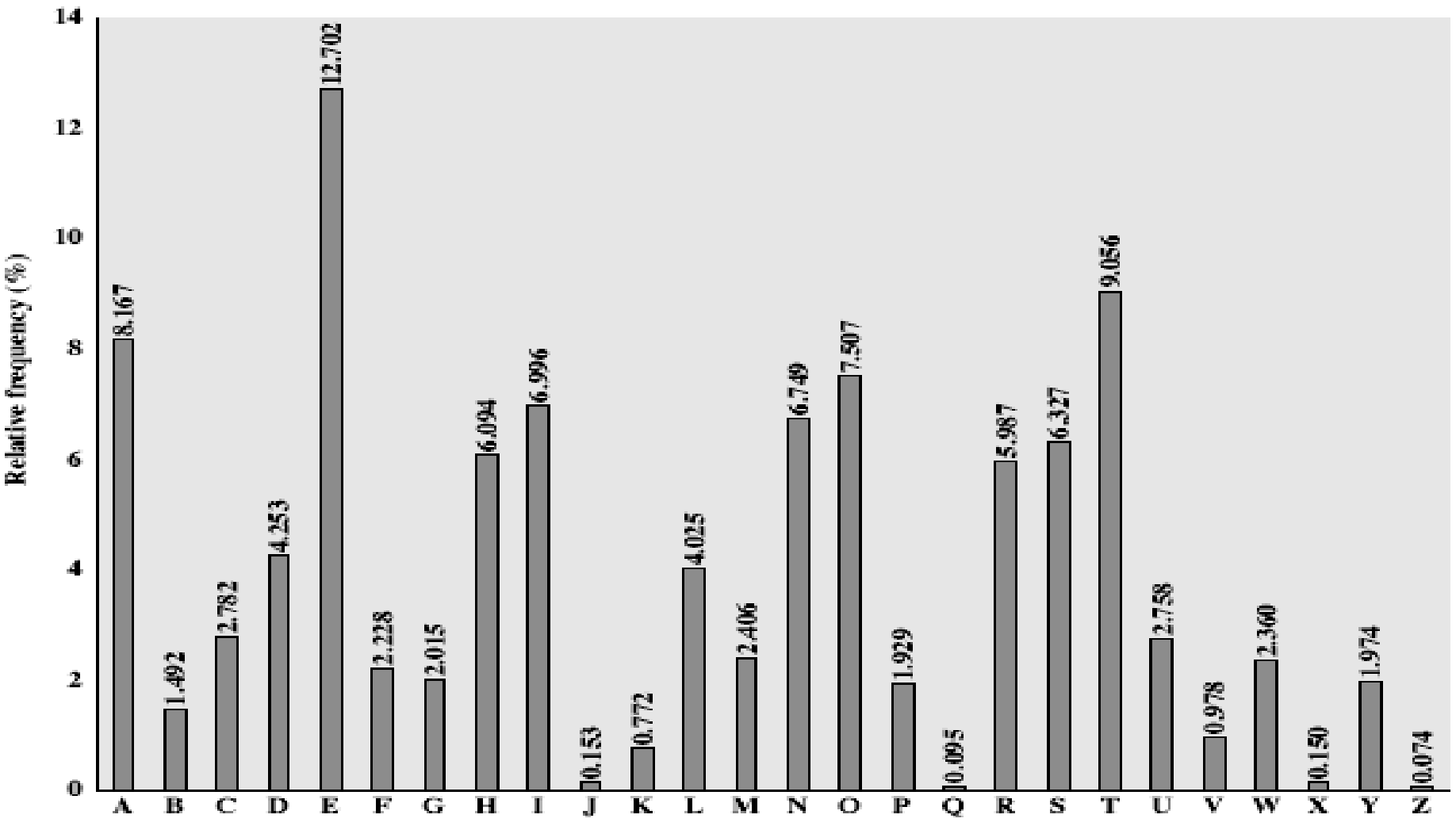

Figure 2.2: ENGLISH LETTER FREQUENCIES

The simplicity and strength of the mono alphabetic substitution cipher meant it dominated cryptographic use for the first millennium AD. It was broken by Arabic scientists. The earliest known description is in AlKindi's "A Manuscript on Deciphering Cryptographic Messages", "رسالة في استخراج المُعمَّى", published in the 9th century but only rediscovered in 1987 in Damascus, but other later works also attest to their knowledge of the field.

The language redundancy and statistical property usage in cryptanalysis was the basic tool in cryptanalysis until recent modern block ciphers.

Many ciphers have been invented until the end of World War II trying to overcome the problem of cryptanalysis arose from language redundancy.

#### *2.1.3.3 Homophonic substitution cipher*

To each symbol a $\in$ A, associate a set H(a) of strings of t symbols, with the restriction that the sets H(a), a $\in$ A, be pair wise disjoint. A *homophonic substitution cipher* replaces each symbol a in a plaintext message block with a randomly chosen string from H(a). To decrypt a string c of t symbols, one must determine an $a \in A$ such that $c \in H(a)$. The key for the cipher consists of the sets H(a).

#### 2.1.3.4 Poly alphabetic cipher

A *poly alphabetic substitution cipher* is a block cipher with block length t over an alphabet A having the following properties:

- The key space K consists of all ordered sets of t permutations ($p_1$, $p_2$, … $p_t$),where each permutation $p_i$ is defined on the set A;
- Encryption of the message $m = (m_1\ m_2 \ldots m_t)$ under the key $e = (p_1, p_2, \ldots p_t)$ is given by $E_e(m) = (p_1(m_1)p_2(m_2) \ldots p_t(m_t))$;and
- The decryption key associated with $e = (p_1, p_2, \ldots p_t)$ is $d = (p_1^{-1}, p_2^{-1}, \ldots p_t^{-1})$

#### 2.1.3.5 Transposition ciphers

Another class of symmetric-key ciphers is the simple transposition cipher, which simply permutes the symbols in a block.

Consider a symmetric-key block encryption scheme with block length t. Let K be the set of all permutations on the set {1,2,….,t}. For each $e \in K$ define the encryption function

$$E_e(m) = (m_e(1)\ m_e(2) \ldots m_e(t)) \quad (2.5)$$

where $m = (m_1 m_2 \ldots m_t) \in M$, the message space. The set of all such transformations is called a *simple transposition cipher.* The decryption key corresponding to e is the inverse permutation $d = e^{-1}$. To decrypt $c = (c_1\ c_2 \ldots c_t)$, compute

$$D_d(c) = (c_d(1)\ c_d(2) \ldots c_d(t)). \quad (2.6)$$

A simple transposition cipher preserves the number of symbols of a given type within a block, and thus is easily cryptanalyzed.

#### 2.1.3.6 Stream ciphers

Stream ciphers form an important class of symmetric-key encryption schemes. They are, in one sense, very simple block ciphers having block length equal to one. What makes them useful is the fact that the encryption transformation can change for each symbol of plain-text being encrypted. In situations where transmission errors are highly probable, stream ciphers are advantageous because they have no error propagation. They can also be used when the data must be processed one symbol at a time (e.g., if the equipment has no memory or buffering of data is limited).

Let K be the key space for a set of encryption transformations. A sequence of symbols $e_1\ e_2\ e_3 \ldots e_i \in K$, is called a *key stream*.

Let A be an alphabet of q symbols and let $E_e$ be a simple substitution cipher with block length 1 where $e \in K$. Let $m_1\ m_2\ m_3 \ldots$ be a plaintext string and let $e_1\ e_2\ e_3 \ldots$.be a key stream from K. *A stream cipher* takes the plaintext string and produces a ciphertext string $c_1\ c_2\ c_3 \ldots$.

Where $c_i = E_{ei}\ (m_i)$.If $d_i$ denotes the inverse of $e_i$, then $D_{di}\ (c_i) = m_i$ decrypts the ciphertext string.

#### 2.1.3.7 Vernam cipher

The *Vernam Cipher* is a stream cipher defined on the alphabet A = {0, 1}.A binary message $m_1\ m_2\ m_3 \ldots m_t$ is operated on by a binary key string $k_1\ k_2\ k_3 \ldots k_t$ of the same length to produce a ciphertext string $c_1\ c_2\ c_3 \ldots c_t$ where

$$c_i = m_i \oplus k_i \quad 1 \le i \le t \tag{2.7}$$

If the key string is randomly chosen and never used again, the Vernam cipher is called a *one-time system* or a *one-time pad*.

It mentioned in [1] that to realize an unbreakable system requires a random key of the same length as the message. This reduces the practicality of the system in all but a few specialized situations.

## 2.2 Confusion and Diffusion

Most of the old attacks on simple ciphers were based on statistical analysis of the output of the cipher compared to the statistical characteristics of the used language of the cipher-text. For example knowing that the most frequent letter in English language is the letter E, results in full discovery of the key used in Caesar cipher, which is a simple substitution cipher [1].

Alternating confusion and diffusion was proposed in [3] to build a product cipher [2]. The idea behind this proposal is to protect the cipher against statistical analysis of output of the cipher system.

In Diffusion, the statistical information of the plaintext is dissipated in a longer range of statistical structure of the cipher-text by making a single plaintext bit affects many bits in the cipher-text, or equivalently, making each cipher-text bit a function of many plaintext bits. This diffusion can be achieved by applying some permutation on the input bits and applying some function on the output of the permutation. This makes bits from different positions in the plaintext contribute in the production of a single bit in the cipher-text. So diffusion makes the relation between the statistics of the input text and the output cipher-text as complex as possible.

Confusion does the same function but between the output bits and the key bits. Thus knowing any information about the statistics of the cipher-text gives the attacker no idea about how the key bits were used to produce that output. This confusion can be achieved by using complex non-linear substitution mapping keyed by the key bits [2].

## 2.3 Iterated Block Ciphers

### 2.3.1 Ideal Block Cipher

As has been shown previously Secret-key algorithms can be divided into block and stream ciphers.

For a block cipher the encryption transformation, for a randomly chosen key, is often modeled as a random permutation, that is, a permutation chosen uniformly at random from the set of all permutations of the plaintext space. The main design criterion for a (secure) block cipher is:

A block cipher should be indistinguishable from a random permutation, for every key; in that case it is called an ideal block cipher [2]. Figure 2.3 shows a permutation for a 4 bit block cipher

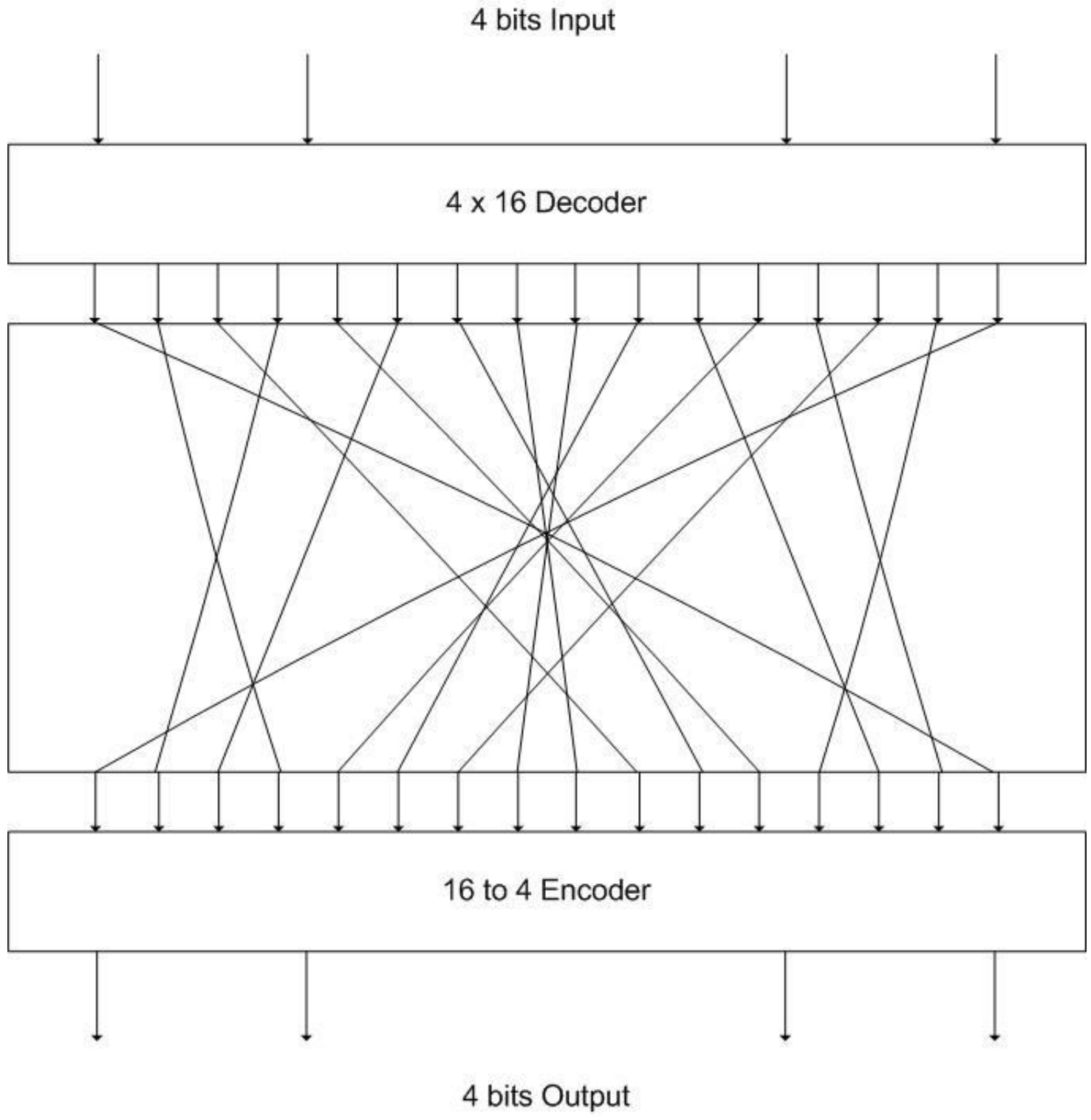

Figure 2.3: GENERAL 4 TO 4 BIT BLOCK PERMUTATION

In practice, for a random n-bit block cipher there are $2^n!$ possible permutations of the plaintexts, which means that the amount of key bits needed to provide all possible permutations is $\log_2(2^n!) \approx n \cdot 2^n$. Key size of most block ciphers is not more than a small multiple of the block size. Therefore, these block ciphers can provide only a small fraction of the total amount of possible permutations. For example, in, Data Encryption Standard, DES the key size is k = 56 bits and the block size is n = 64 bits. Therefore, out of the $2^{64}!$ permutations of 64-bit blocks, only $2^{56}$ transformations can be selected. This example is shows how far DES is from an ideal block cipher. The key size of 56 bits is much less than the ideal key size which is $2^{70}$ as $2^{70}>>>56$.

Every attack on a block cipher starts by distinguishing the given cipher from a random permutation. The next step is usually a key-recovery attack if the resulting complexity is lower than that of a brute-force attack. The proposed alternation of confusion and diffusion introduced the concept of a ***product cipher*** that combines simple operations, which individually do not provide sufficient protection, but jointly, results in a complex and more secure cipher. Further, in order to simplify the implementation, many product ciphers have an iterated structure.

An ***iterated block cipher*** consists of the composition of a key-dependent and invertible structure called a round transformation. Two kinds of round structure are most commonly found in modern iterated block ciphers: the Substitution-Permutation Network (SPN) and the Feistel Network (FN). The round transformation $T_i$ of both types of iterated ciphers has the following general form: $X_i = T_i(X_{i-1};K_i)$, where $X_{i-1}$ is the $i^{th}$ round input block, $X_i$ is the output of the $i^{th}$ round and $K_i$ is the round sub key. The initial input, $X_0$, is the plaintext block, and the output after r rounds, $X_r$, is the ciphertext block.

### 2.3.2 Basic Substitution Permutation Networks

Feistel [4] was the first to suggest that an SPN architecture consisting of rounds of nonlinear substitutions (S-boxes) connected by bit position permutations was a simple, effective implementation of Shannon's concept of a "mixing transformation" based on the principles of "confusion" and "diffusion" [3]. Many modern block ciphers, including Rijndael cipher [5], [6], which is selected as the new AES, Advanced Encryption Standard, by the National Institute of Standards and Technology, NIST are based on the basic SPN, so understanding such architecture is very helpful for the other modern block ciphers.

#### *2.3.2.1 Substitution*

The description of the basic SPN is best declared in [7]. In this cipher, the 16-bit data block is broken into four 4-bit sub-blocks. Each sub-block forms an input to a 4×4 S-box (a substitution with 4 input and 4 output bits), which can be easily implemented with a lookup table, like the one presented in table 2.1, of sixteen 4-bit values, indexed by the integer represented by the 4 input bits. The most fundamental property of an S-box is that it is a nonlinear mapping, i.e., the output bits cannot be represented as a linear operation on the input bits. The mapping chosen for this cipher, given in Table 2.2, is chosen from the S-boxes of DES. In the table, the most significant bit of the hexadecimal notation represents the leftmost bit of the S-box in Figure 2.4.

| Input | 0 | 1 | 2 | 3 | 4 | 5 | 6 | 7 | 8 | 9 | A | B | C | D | E | F |
|---|---|---|---|---|---|---|---|---|---|---|---|---|---|---|---|---|
| Output | E | 4 | D | 1 | 2 | F | B | 8 | 3 | A | 6 | C | 5 | 9 | 0 | 7 |

**Table 2.1. S-box Representation (in hexadecimal)**

| Input | 0 | 1 | 2 | 3 | 4 | 5 | 6 | 7 | 8 | 9 | A | B | C | D | E | F |
|---|---|---|---|---|---|---|---|---|---|---|---|---|---|---|---|---|
| $S_{11}$ | E | 4 | D | 1 | 2 | F | B | 8 | 3 | A | 6 | C | 5 | 9 | 0 | 7 |
| $S_{12}$ | 0 | F | 7 | 4 | E | 2 | D | 1 | A | 6 | C | B | 9 | 5 | 3 | 8 |
| $S_{13}$ | 4 | 1 | E | 8 | D | 6 | 2 | B | F | C | 9 | 7 | 3 | A | 5 | 0 |
| $S_{14}$ | F | C | 8 | 2 | 4 | 9 | 1 | 7 | 5 | B | 3 | E | A | 0 | 6 | D |
| $S_{21}$ | F | 1 | 8 | E | 6 | B | 3 | 4 | 9 | 7 | 2 | D | C | 0 | 5 | A |
| $S_{22}$ | 3 | D | 4 | 7 | F | 2 | 8 | E | C | 0 | 1 | A | 6 | 9 | B | 5 |
| $S_{23}$ | 0 | E | 7 | B | A | 4 | D | 1 | 5 | 8 | C | 6 | 9 | 3 | 2 | F |
| $S_{24}$ | D | 8 | A | 1 | 3 | F | 4 | 2 | B | 6 | 7 | C | 0 | 5 | E | 9 |
| $S_{31}$ | A | 0 | 9 | E | 6 | 3 | F | 5 | 1 | D | C | 7 | B | 4 | 2 | 8 |
| $S_{32}$ | D | 7 | 0 | 9 | 3 | 4 | 6 | A | 2 | 8 | 5 | E | C | B | F | 1 |
| $S_{33}$ | D | 6 | 4 | 9 | 8 | F | 3 | 0 | B | 1 | 2 | C | 5 | A | E | 7 |
| $S_{34}$ | 1 | A | D | 0 | 6 | 9 | 8 | 7 | 4 | F | E | 3 | B | 5 | 2 | C |
| $S_{41}$ | 7 | D | E | 3 | 0 | 6 | 9 | A | 1 | 2 | 8 | 5 | B | C | 4 | F |
| $S_{42}$ | D | 8 | B | 5 | 6 | F | 0 | 3 | 4 | 7 | 2 | C | 1 | A | E | 9 |
| $S_{43}$ | A | 6 | 9 | 0 | C | B | 7 | D | F | 1 | 3 | E | 5 | 2 | 8 | 4 |
| $S_{44}$ | 3 | F | 0 | 6 | A | 1 | D | 8 | 9 | 4 | 5 | B | C | 7 | 2 | E |

**Table 2.2 16 S-Boxes used in the SPN**

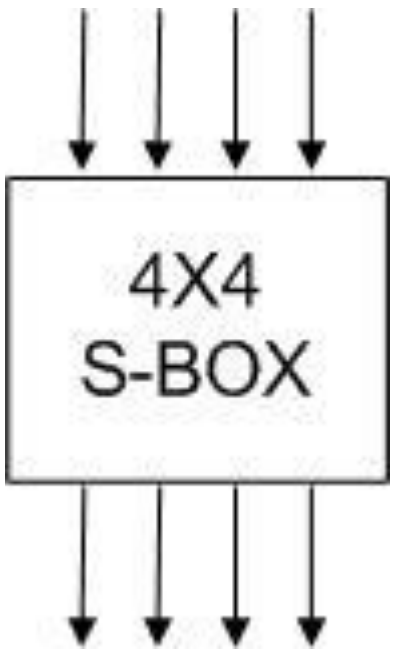

Figure 2.4: S-BOX MAPPING

#### *2.3.2.2 Permutation*

The permutation portion of a round is simply the transposition of the bits or the permutation of the bit positions. The permutation of Figure 2.5 is given in Table 2.3 (where the numbers represent bit positions in the block, with 1 being the leftmost bit and 16 being the rightmost bit) and can be simply described as: the output *i* of S-box *j* is connected to input *j* of S-box *i*. Note that there would be no purpose for a permutation in the last round and, hence, this cipher does not need a permutation.

| Input | 1 | 2 | 3 | 4 | 5 | 6 | 7 | 8 | 9 | 10 | 11 | 12 | 13 | 14 | 15 | 16 |
|---|---|---|---|---|---|---|---|---|---|---|---|---|---|---|---|---|
| Output | 1 | 5 | 9 | 13 | 2 | 6 | 10 | 14 | 3 | 7 | 11 | 15 | 4 | 8 | 12 | 16 |

**Table 2.3 Permutation**

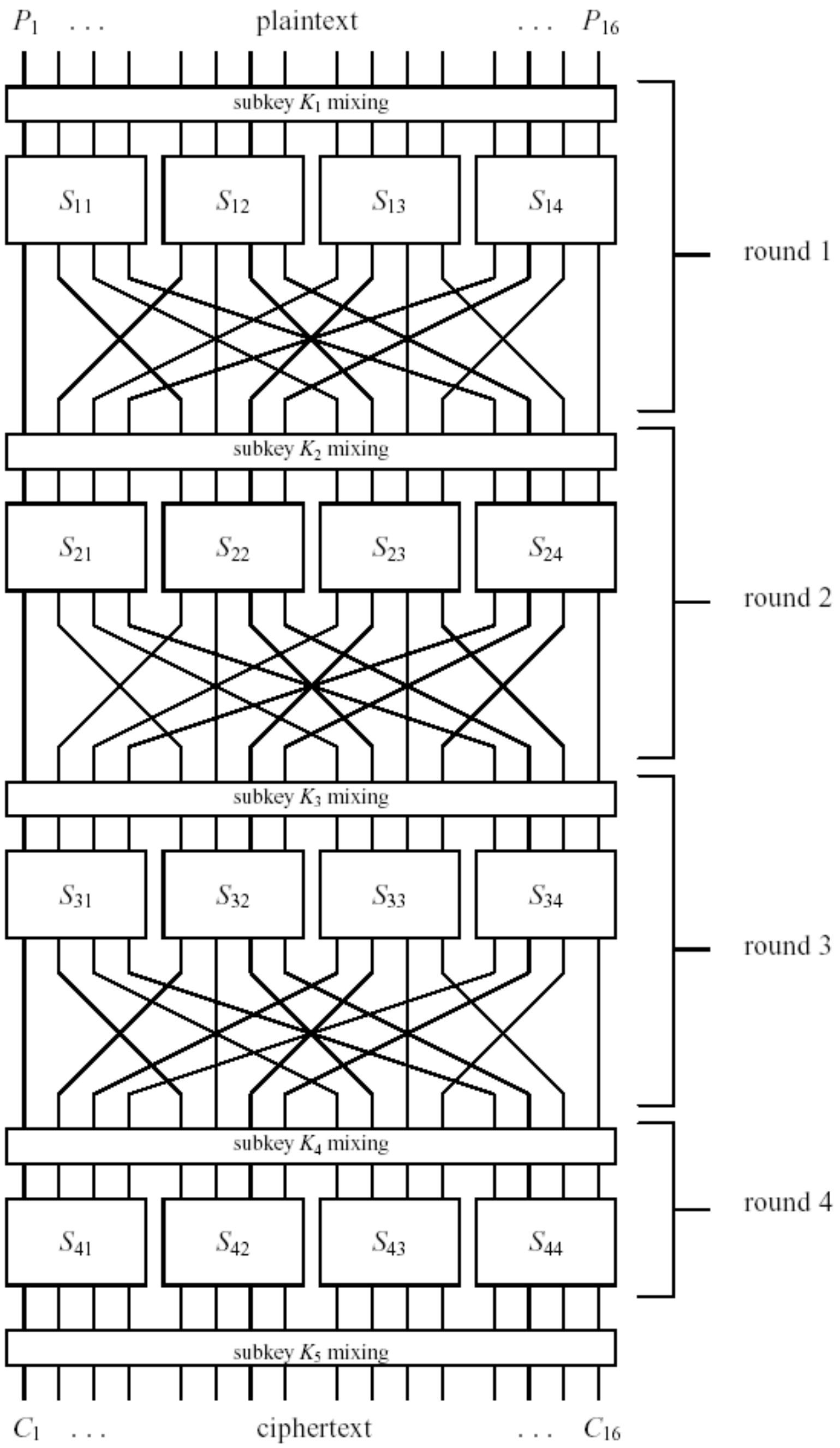

Figure 2.5: BASIC SUBSTITUTION-PERMUTATION NETWORK (SPN) CIPHER

#### *2.3.2.3 Key Mixing*

To achieve the key mixing, a simple bit-wise exclusive-OR between the key bits associated with a round (referred to as a sub-key) and the data block input to the round is used. As well, a sub-key is applied following the last round, ensuring that the last layer of substitution cannot be easily ignored by a cryptanalyst which simply works backward through the last round's substitution. Normally, in a cipher, the sub-key for a round is derived from the cipher's master key through a process known as the key schedule. In this cipher, all bits of the sub-keys are independently generated and unrelated.

#### *2.3.2.4 Decryption*

To perform decryption, ciphertext is essentially passed backwards through the network. Hence, decryption is also of the form of an SPN as illustrated in Figure 2.5. However, the mappings used in the S-boxes of the decryption network are the inverse of the mappings in the encryption network (i.e., input becomes output, output becomes input). This implies that for an SPN to allow for decryption, all S-boxes must be bijective, that is, a one-to-one mapping with the same number of input and output bits. For the network to properly decrypt, the sub-keys are applied in reverse order and the bits of the sub-keys must be moved around according to the permutation, if the SPN is to look similar to Figure 2.5. Note also that the lack of the permutation after the last round ensures that the decryption network can be of the same structure as the encryption network.

### 2.3.3 Feistel Ciphers

Feistel also proposed another product cipher widely known as Feistel cipher or Feistel network [2]. Unlike the SPN, Feistel network can be composed of non-reversible components while the entire cipher remains reversible. The idea is also based on performing successive rounds of substitution followed by some sort of halves swapping and XOR. Feistel cipher is very popular and many famous block ciphers use this structure like DES and Blowfish.

The following sections will describe the basic Feistel network structure

#### *2.3.3.1 Basic Feistel Cipher Structure*

The basic Feistel cipher with six rounds is shown in figure 2.6

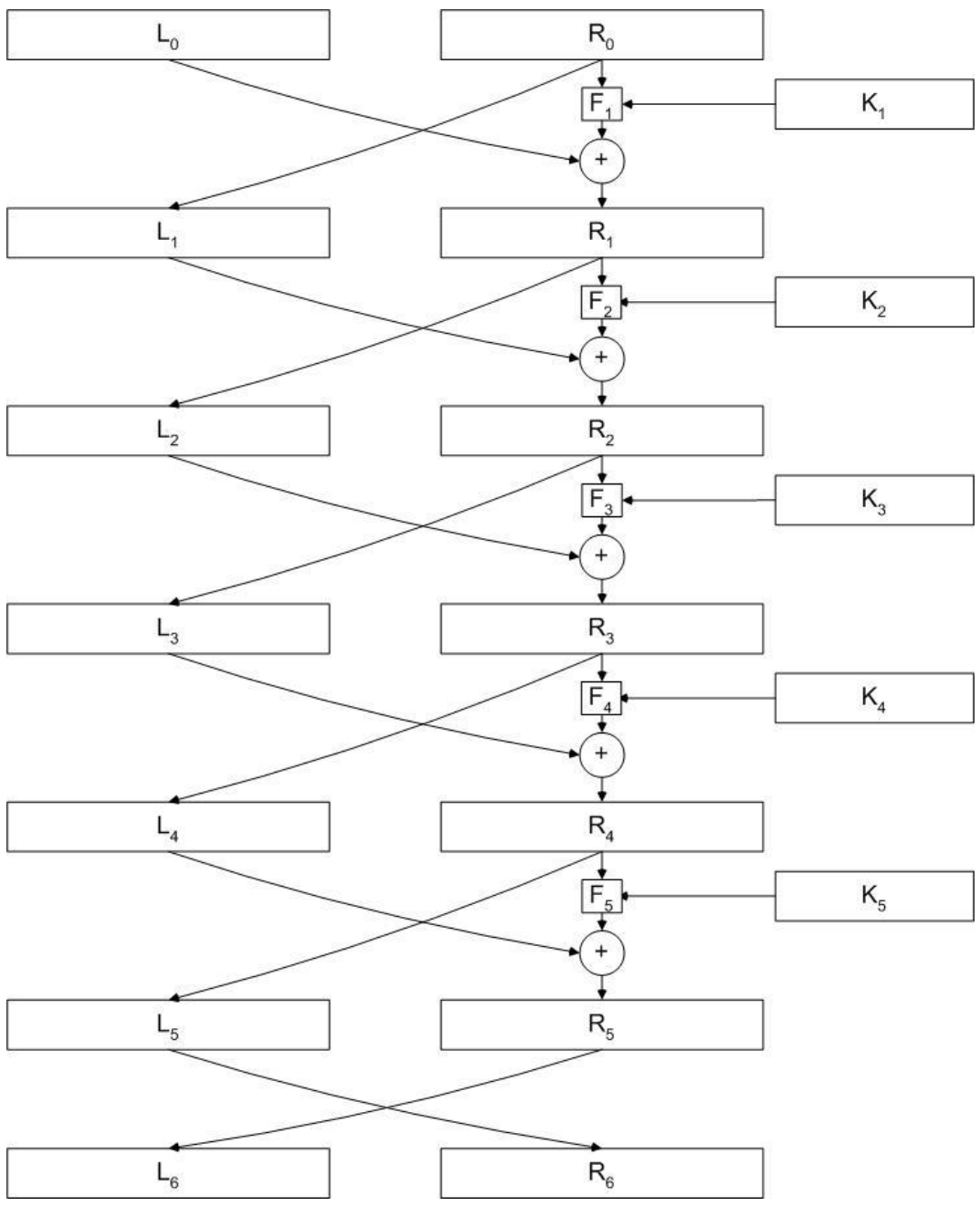

Figure 2.6: A 6 ROUND FEISTEL NETWORK

As seen from figure 2.6, Feistel cipher gets a block of 2n bits of plaintext and produces a 2n bits block of cipher-text.

Each block is divided into two halves, a left (L) and a right (R) ones n bits each.

The two halves block passes through a number of rounds (in our case they are five complete rounds and a final one). In each round an initial permutation of the right half into the output left half is done, then the right half is passed through a keyed function using an m bits key (usually m=n) and finally the output is mixed with the left half with an XOR. Equations 2.8 and 2.9 describe the input output relation of a round.

Consider the input block to round i is $L_{i-1}R_{i-1}$ then the output of the round is

$$L_i = R_{i-1} \qquad (2.8)$$

$$R_i = F_i(R_{i-1}, K_i) \oplus L_i \qquad (2.9)$$

Where $F_i$ is a keyed function that maps n bits input to n bits output.

In the final round, there is a block permutation. This round gives the Feistel cipher a very interesting phenomenon, which is decryption is basically an encryption with reversed order keys and functions ($K_i$, $F_i$).

To show this criterion, consider figure 2.7 of one round of a Feistel cipher

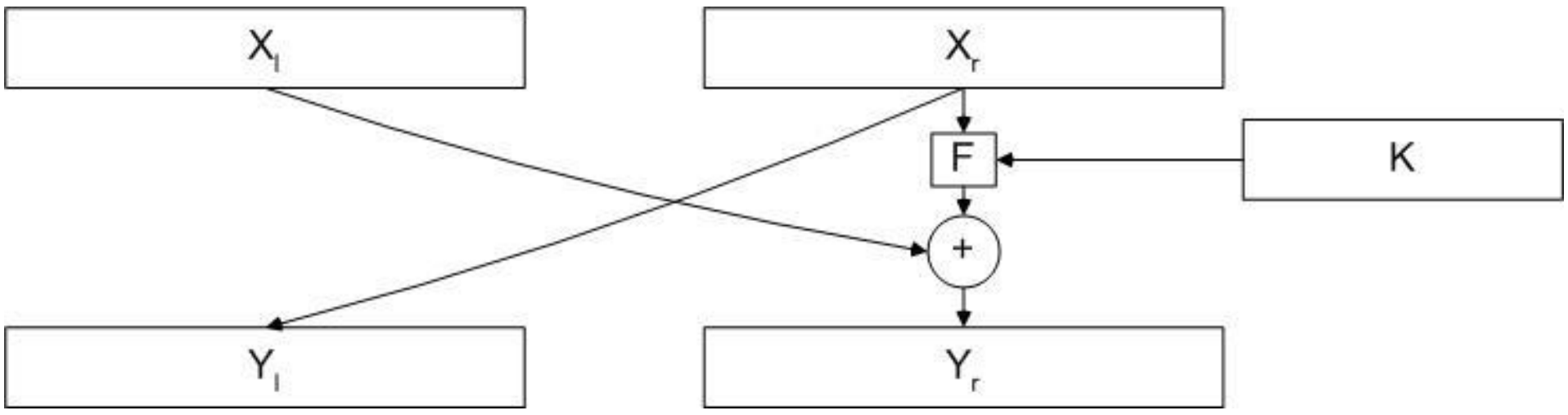

Figure 2.7: ONE ROUND OF A FEISTEL NETWORK

Now applying equations 2.8 and 2.9 we find that

$$Y_l = X_r \tag{2.10}$$

$$Y_r = F(X_r, K) \oplus X_l \tag{2.11}$$

Now consider swapping the outputs and feeding them back again into the same round, see figure 2.8

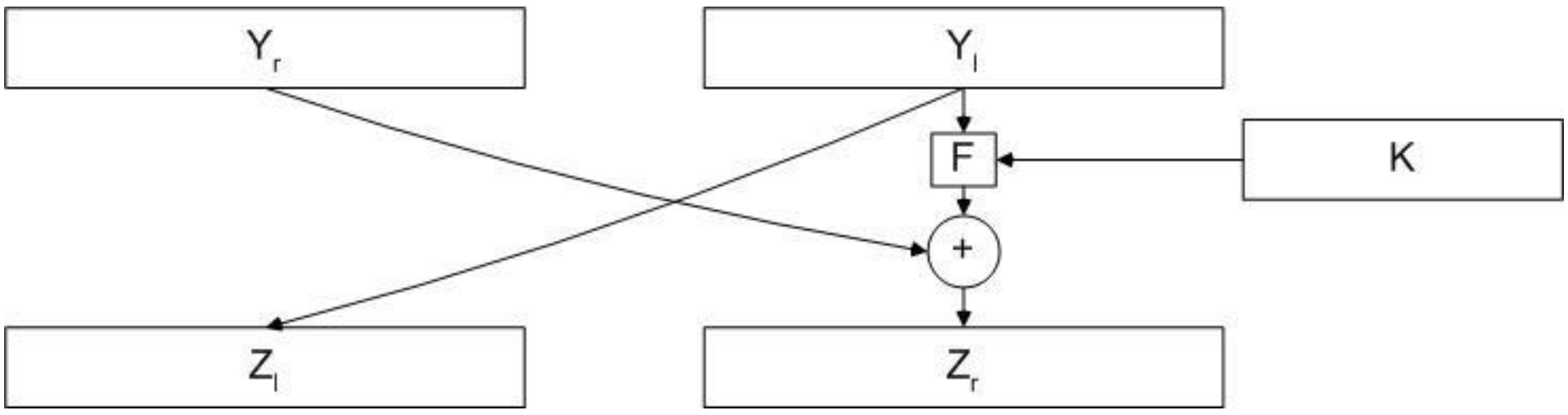

Figure 2.8: OUTPUT SWAPPED AND FED AGAIN TO THE ROUND

Applying the equation again will result the following

$$Z_l = Y_l = X_r \tag{2.12}$$

$$\begin{aligned} Z_r &= F(Y_l, K) \oplus Y_r \\ &= F(X_r, K) \oplus Y_r \end{aligned} \tag{2.13}$$

Substituting 2.11 into 2.13. results

$$Z_r = F(X_r, K) \oplus F(X_r, K) \oplus X_l = X_l \tag{2.14}$$

And from 2.12 and 2.14 we notice that the original block is restored but in reverse order. The last block permutation round give the Fiestel cipher its nature of being reversible

#### *2.3.3.2 Encryption*

Encryption is performed by passing the 2n bits block through rounds in forward order. Each round involves a keyed function that has two inputs, a half block and a round key. This function provides the confusion effect of the cipher and it may also provide diffusion if permutation is used inside it.

The round function generally acts on two inputs, the half block input and key, but usually bit wise XOR is used to mix the key with the half block and then the result is passed to a complex function. Another approach is to use a key dependent round function that is selected by the key itself. Blowfish is an example of using key dependent round function.

In DES the round function F is a very complex one with substitution and bit position permutation which involves expansion and compression of bits.

#### *2.3.3.3 Decryption*

Decryption is as mentioned before uses the same encryption structure by passing the cipher-text through the system but with reversed order keys ($K_n$, $K_{n-1}$, … $K_1$) and functions. This means that it is not needed to implement separate encryption and decryption algorithms.

### 2.3.4 Key Scheduling

In both of the two models described earlier, each sub round needs an m bit key. The most secured way is to have independent round keys but this is impractical, since in many ciphers, the number of rounds is very big and also the key size, for example in DES we have 16 rounds with 56 bit round key and in Rijndael 128 variant we have 11 rounds with 128 bits round key.

Instead of using independent round keys, a scheduling algorithm is used to generate each round key from the original cipher key. Key scheduling is a major component in the overall cipher design and may be a week point from which attacker may compromise the entire cipher.

## 2.4 Modes of Operation

### 2.4.1 Definition and characteristics of a mode of operation

As discussed in the previous section, it is concluded that a block cipher is basically a transformation or mapping between two domains, the plaintext domain and the ciphertext domain.

The block cipher as described is said to be working in the Electronic Code Book mode, ECB,

In this mode, the plaintext must be subdivided into n bits blocks, padding for smaller blocks may be needed.

Also if the same block is fed to the cipher more than one time, the result blocks of ciphertext will always be the same.

Of course the statistical analysis of large block size is very hard but it is still going to increase the security if the output ciphertext block depends not only on the input plaintext block but also on the entire plian text blocks or at least the once preceded this block.

Moreover, depending on the cryptographic application, the plaintext messages may need to be processed in segments smaller than the block size, characters for example,

Modes of operations are a form of cryptographic protocol that provides solutions to the mentioned problems.

Each mode of operation can be characterized by the following parameters [8],

- Plaintext Redundancy: means that the mode allows the plain-text statistics to propagate to the ciphertext, for example, if equal plaintext blocks are being encrypted to the same ciphertext block (Knudsen [9]).

- Random Access: means that the mode allows an arbitrary ciphertext block to be read or modified without the need to access or modify (update) other blocks.
- Parallel Processing: means that the mode allows simultaneous processing of neighbouring blocks.
- Error Expansion: concerns garbled text blocks due to bit-flipping errors (bits are complemented but there is no change in the amount of bits) or slip errors (arbitrary insertion or deletion of bits) in the ciphertext which results in one of the following cases:
  - No error expansion (errors limited to the faulty block or segment)
  - Finite error expansion (a limited number of blocks are affected),
  - Infinite error expansion.
- Counter or Initial Value: means that the mode requires additional values, independent of the text and the key, for each new message or for resynchronization. Such values may be (for security reasons):
  - Random
  - Unpredictable
  - Non-repeating.

### 2.4.2 The Electronic Code Book (ECB) mode

The Electronic Code Book (ECB) mode has been standardized by NIST [10] and ISO [11].

ECB consists of encrypting each individual plaintext block independently of the other blocks.

Encryption is defined as $E_k : k \in K, C = E_k(P)$ where E is the encryption transformation, P is the plaintext block, C is the ciphertext block, e is the key and K is the key space. Decryption is defined as $D_k : k \in K, P = D_k(C)$. Figure 2.9 demonstrates the ECB mode

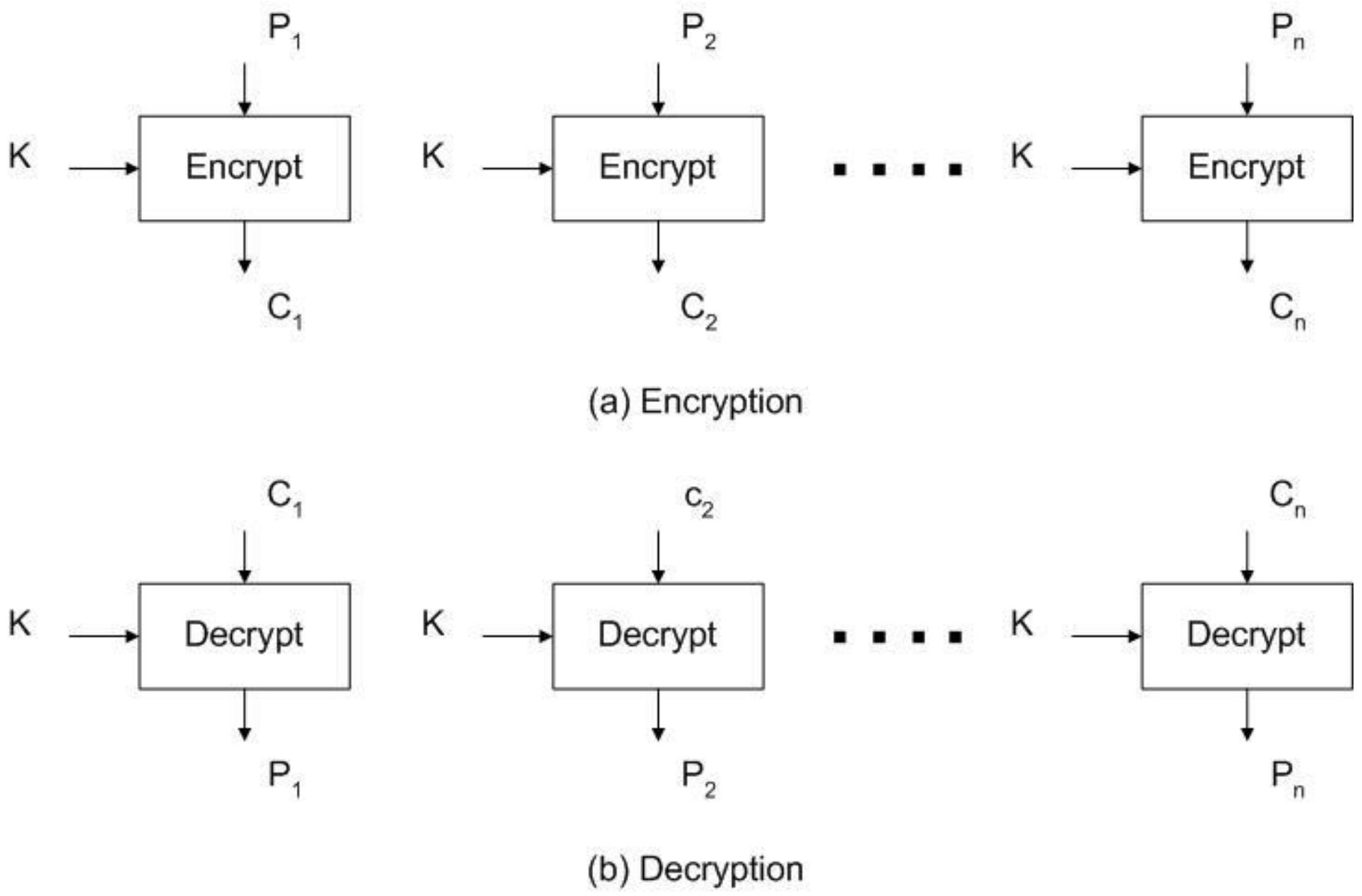

Figure 2.9: ECB MODE

Table 2.4 shows the characteristics of the ECB mode based on the measuring parameters introduced earlier

| | |
|---|---|
| Encryption | $C_i = E_k(P_i)$ |
| Decryption | $P_i = D_k(C_i)$ |
| Plaintext redundancy | Not concealed in ECB mode because each block is encrypted separately, and equal plaintext blocks, under the same key, result in equal ciphertext blocks. |
| Random access | Since each block is considered separately, Random access is possible |
| Parallel processing | Possible due to independent encryption (decryption) of each block |
| Bit-flipping errors propagation | The error does not propagate from a block to the next and is limited only to the block it happened in. |
| Slip errors propagation | The slip error propagates infinitely because block boundaries are totally lost |
| Counter or Initial Value | No extra value other than the key is needed for this ECB mode to operate |

**Table 2.4: ECB Mode Characteristics.**

Because of the independent processing of each text block, an attacker can insert, delete or change the order of ciphertext blocks without affecting decryption, unless there is enough plaintext redundancy or some separate integrity mechanism to allow detection of active tampering.

### 2.4.3 The Cipher Block Chaining (CBC) mode

The Cipher Block Chaining (CBC) mode has been also standardized by NIST [10] and ISO [11]. CBC consists of processing plaintext blocks combined with a feedback of the previous ciphertext block.

Figure 2.10 demonstrates the CBC mode

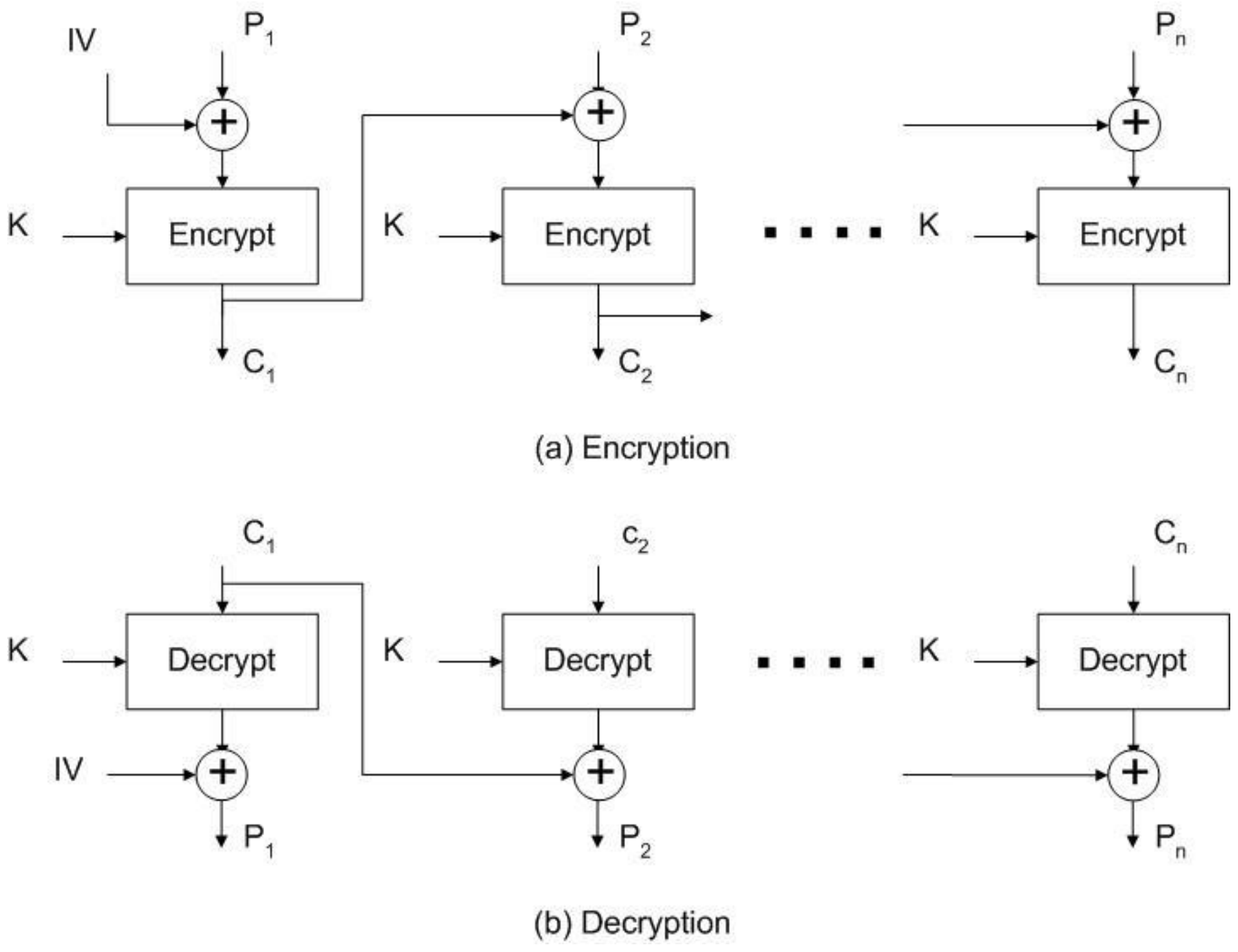

Figure 2.10: CBC MODE

Table 2.5 shows the characteristics of the CBC mode

| | |
|---|---|
| Encryption | $C_i = E_k(P_i \oplus C_{i-1})$<br>$C_0 = IV$ |
| Decryption | $C_0 = IV$<br>$P_i = D_k(C_i) \oplus C_{i-1}$ |
| Plaintext redundancy | Concealed in CBC mode because each block depends on the previous blocks |
| Random access | Since each block is dependent on the previous block, Random access is possible if the previous block is known. |
| Parallel processing | Chaining is sequential |
| Bit-flipping errors propagation | The bit flipping error of a ciphertext block affects the block and the one after, finite propagation. Upon receiving two correct blocks the mode resynchronizes itself. |
| Slip errors propagation | The slip error propagates infinitely because block boundaries are totally lost |
| Counter or Initial Value | Initial value is needed. The choice of the initial value affects the security of the mode. It should be randomly selected and key independent [12] |

**Table 2.5: CBC Mode Characteristics.**

### 2.4.4 The Cipher Feedback (CFB) mode

The Cipher Feedback (CFB) mode has been standardized by NIST [10] and ISO [11]. CFB generates a key-and data-dependent stream of segments of r bits ≤ n bits, block size that is combined via XOR to the plaintext.

Figure 2.11 demonstrates the CFB mode

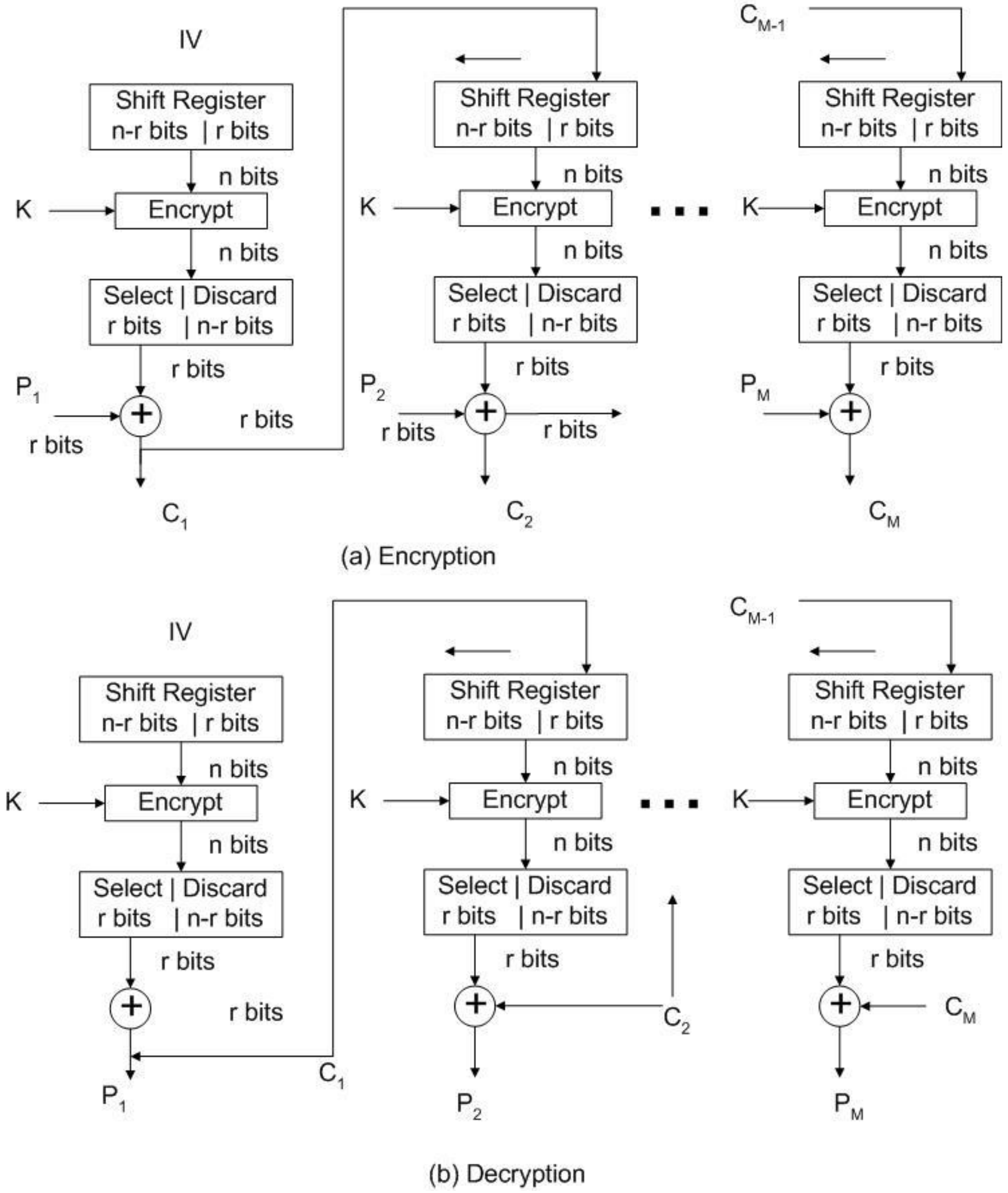

Figure 2.11: CFB MODE

Table 2.6 shows the characteristics of the CFB mode

| | |
|---|---|
| Encryption | $I_0 = IV$<br>$C_i = P_i \oplus left_r(E_k(I_{i-1}))$<br>$I_i = right_{n-r}(I_{i-1}) \| c_i$ |
| Decryption | $I_0 = IV$<br>$P_i = C_i \oplus left_r(E_k(I_{i-1}))$<br>$I_i = right_{n-r}(I_{i-1}) \| c_i$ |
| Plaintext redundancy | Concealed in CFB mode because each block depends on the previous blocks |
| Random access | Random access requires $\left\lceil \frac{n}{r} \right\rceil$ segments or the initial value. |
| Parallel processing | Sequential |
| Bit-flipping errors propagation | The bit flipping error of a ciphertext block affects the block and next blocks until the r bits containing the error are shifted out. |
| Slip errors propagation | The slip error propagates infinitely because block boundaries are totally lost if the slip is not bounded to exactly r bits. If r bits slip occurs, the propagation is like bit flipping |
| Counter or Initial Value | Initial value is needed. The choice of the initial value affects the security of the mode. It should be randomly selected and key independent. The initial value needn't be secret. |

**Table 2.6: CFB Mode Characteristics.**

Note that only encryption operation is considered. The CFB mode can be used to implement stream ciphers using n bits block ciphers.

### 2.4.5 Output Feedback (OFB) mode

The Output Feedback (OFB) mode has been standardized by NIST [10] and ISO [11]. OFB generates a key stream [13] in steps of n bits, of which $r \le n$ bits can be combined to the plaintext segments via exclusive-or. Figure 2.12 demonstrates the OFB mode

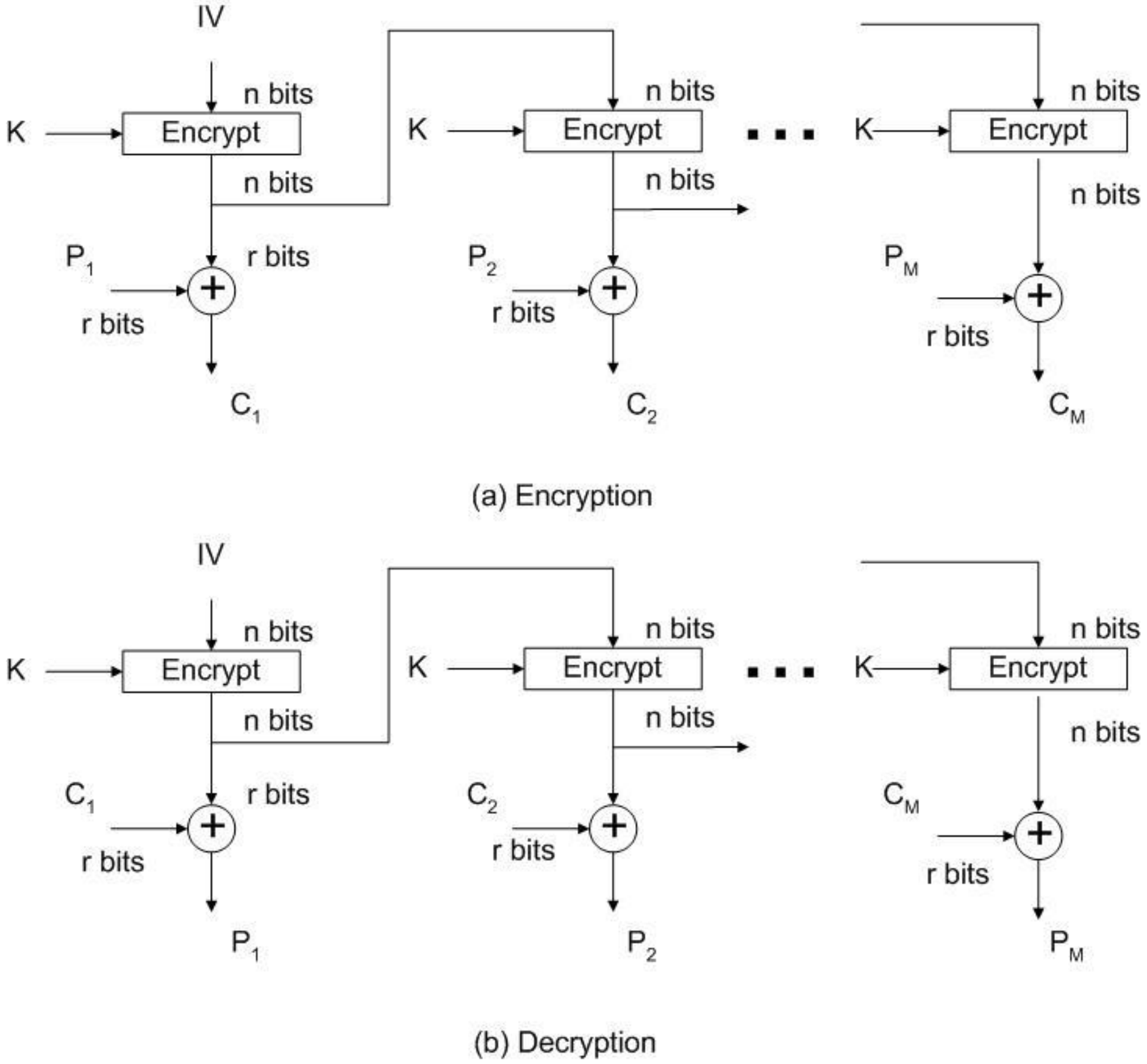

Figure 2.12: OFB MODE

Table 2.7 shows the characteristics of the OFB mode

| | |
|---|---|
| Encryption | $I_0 = IV$<br>$C_i = P_i \oplus left_r(I_i)$<br>$I_i = E_k(I_{i-1})$ |
| Decryption | $I_0 = IV$<br>$P_i = C_i \oplus left_r(I_i)$<br>$I_i = E_k(I_{i-1})$ |
| Plaintext redundancy | Concealed in OFB mode because each block depends on the previous blocks |
| Random access | Since each block is considered separately, Random access is possible |
| Parallel processing | Also parallel processing is possible (if all the I elements are generated previously knowing the initial value) due to independent encryption (decryption) of each block |
| Bit-flipping errors propagation | The error does not propagate from a block to the next and is limited only to the block it happened in. |
| Slip errors propagation | The slip error propagates infinitely because initial value is totally lost and resynchronization is needed. |
| Counter or Initial Value | Initial value is needed. The choice of the initial value affects the security of the mode. It should be randomly selected and key independent. It should also be changed rapidly |

**Table 2.7: OFB Mode Characteristics.**

As in CFB, Note that only encryption operation is considered. The OFB mode can be used to implement stream ciphers using n bits block ciphers.

### 2.4.6 The Counter (CTR) mode

The Counter (CTR) mode was originally suggested by Diffie and Hellman [14]. More recently it was reviewed and submitted by Lipmaa et al. [15] for the NIST Modes of Operation Workshop [16]. This mode is also recommended in a NIST Special Publication [17], and in ISO [11]. The CTR mode consists of processing plaintext blocks with a counter-dependent encrypted output [18]. Figure 2.13 demonstrates the CTR mode

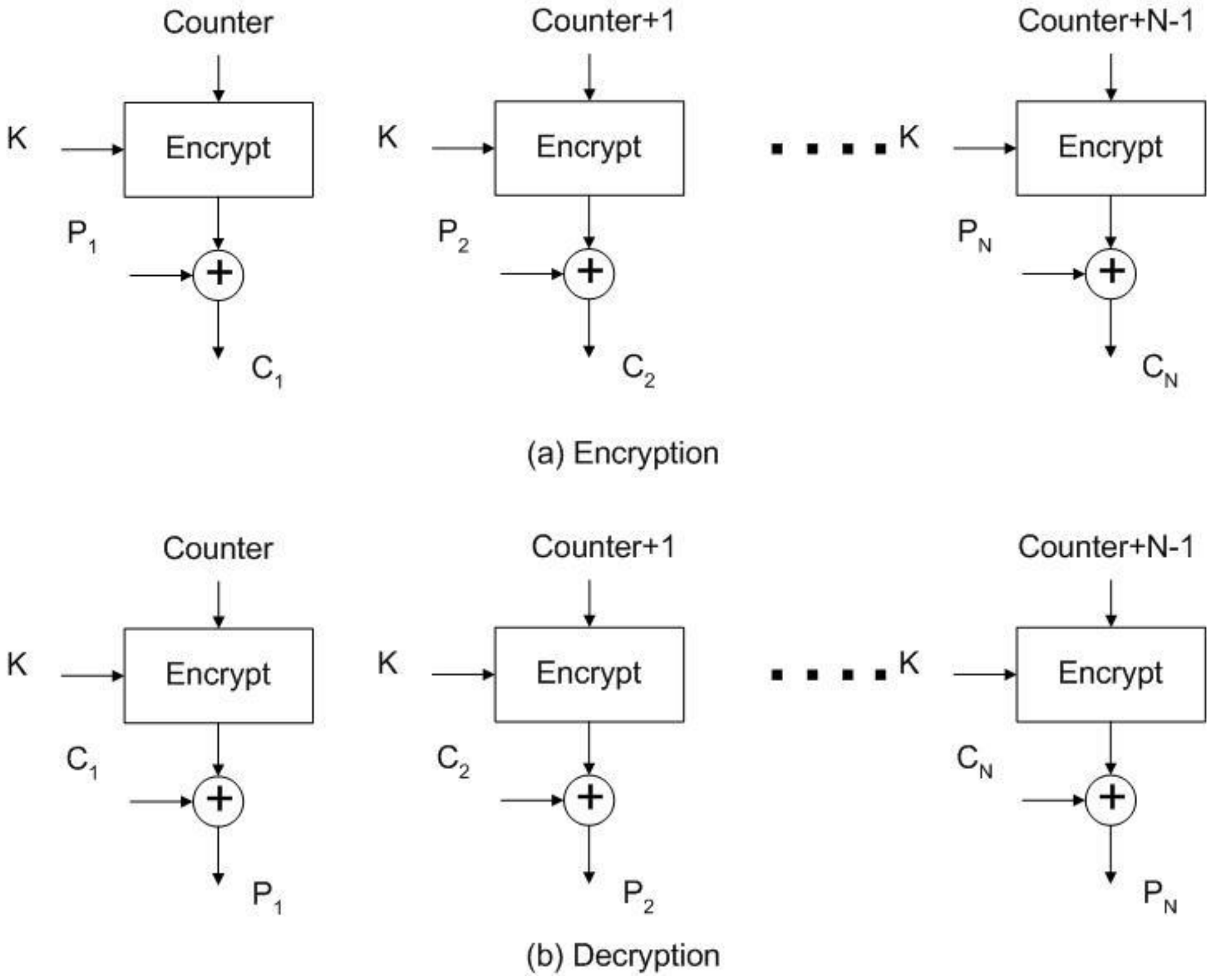

Figure 2.13: CTR MODE

Table 2.8 shows the characteristics of the CTR mode

| | |
|---|---|
| Encryption | $I_0 = IV$<br>$C_i = P_i \oplus E_k(I_{i-1})$<br>$I_i = (I_{i-1} + 1) \bmod 2^n$ |
| Decryption | $I_0 = IV$<br>$P_i = C_i \oplus E_k(I_{i-1})$<br>$I_i = (I_{i-1} + 1) \bmod 2^n$ |
| Plaintext redundancy | Concealed in CTR mode because each block depends on the time (cont) at which it was encoded. |
| Random access | Since each block is considered separately, Random access is possible |
| Parallel processing | Also parallel processing is possible (if all the I elements are generated previously knowing the initial value) due to independent encryption (decryption) of each block |
| Bit-flipping errors propagation | The error does not propagate from a block to the next and is limited only to the block it happened in. |
| Slip errors propagation | The slip error propagates infinitely because initial value is totally lost and resynchronization is needed. |
| Counter or Initial Value | Initial value is needed. The choice of the initial value affects the security of the mode. It should be randomly selected and key independent. It should also be changed rapidly and should have a very long period. |

**Table 2.8: CTR Mode Characteristics.**

After the AES Development Process [19], NIST organized two workshops to discuss new and existing modes of operation for the AES in particular, and for block ciphers in general. In 2001, NIST issued a special publication [17] suggesting the five modes described previously. Nonetheless, several other modes were submitted for evaluation, including: the ABC (Accumulated Block Chaining) by Knudsen [20], the IACBC (Integrity Aware CBC) and IAPM (Integrity Aware Parallelizable

Mode) by Jutla [21], the KFB (Key Feedback Mode) by Hastad and Naslund [22], the OCB (Offset Code Book) by Rogaway [23], the PCFB (Propagating CFB) by Hellstrom [24], the XCBC-XOR (Extended CBC) and XECB-XOR (Extended ECB) by Gligor and Donescu [25], and the 2DEM (2D Encryption Mode) by Belal and Abdel-Gawad [26].

## 2.5 Examples

### 2.5.1 DES

Data Encryption Standard (DES) [27], [2], and [1] is the world most used block cipher. It was the standard of non confidential use in the United States government and in many commercial applications for more than 20 years, 1977 until October 2000. The most advances in cryptanalysis techniques were primarily for breaking DES. Most new block ciphers are designed to defeat those attacks and to outperform DES.

DES encrypts 64-bit data block using 56-bit key using a standard Feistel structure. It is a complicated cipher and its design steps were classified especially its S-Box design.

#### *2.5.1.1 DES History*

By the efforts of IBM to produce commercial cipher products that fit in one chip, IBM developed Lucifer cipher by a team led by Feistel. Lucifer cipher utilizes 64-bit data blocks with 128-bit key. A new version of Lucifer was developed for commercial interests. The new cipher was more resistive to cryptanalysis but with a smaller key size, to fit on single chip. The new key size was 56 bits.

In 1973 the National Bureau of Standards, NBS, issued a request for proposals for a national cipher standard. IBM submitted their revised Lucifer to the NBS which was eventually accepted as the DES.

#### *2.5.1.2 DES structure*

DES is basically a 16 round Feistel cipher with extra round before the first one and its inverse after the 16$^{th}$ round.

##### *2.5.1.2.1 Initial Permutation*

The extra round is an initial permutation which rearranges the bits of the input 64 block of data. The initial permutation is as defined in table 2.9 where each cell indicates where a bit from the input should be placed after permutation. Bit positions are started from 1to 64.

| IBP | OBP | IPB | OBP | IPB | OPB | IPB | OBP |
|---|---|---|---|---|---|---|---|
| 1 | 58 | 17 | 62 | 33 | 57 | 49 | 61 |
| 2 | 50 | 18 | 54 | 34 | 49 | 50 | 53 |
| 3 | 42 | 19 | 46 | 35 | 41 | 51 | 45 |
| 4 | 43 | 20 | 38 | 36 | 33 | 52 | 37 |
| 5 | 26 | 21 | 30 | 37 | 25 | 53 | 29 |
| 6 | 18 | 22 | 22 | 38 | 17 | 54 | 21 |
| 7 | 10 | 23 | 14 | 39 | 9 | 55 | 13 |
| 8 | 2 | 24 | 6 | 40 | 1 | 56 | 5 |
| 9 | 60 | 25 | 64 | 41 | 59 | 57 | 63 |
| 10 | 52 | 26 | 56 | 42 | 51 | 58 | 55 |
| 11 | 44 | 27 | 48 | 43 | 43 | 59 | 47 |
| 12 | 36 | 28 | 40 | 44 | 35 | 60 | .39 |
| 13 | 28 | 29 | 32 | 45 | 27 | 61 | 31 |
| 14 | 20 | 30 | 24 | 46 | 19 | 62 | 23 |
| 15 | 12 | 31 | 16 | 47 | 11 | 63 | 15 |
| 16 | 4 | 32 | 8 | 48 | 3 | 64 | 7 |

**Table 2.9, DES Initial Permutation,**

**(IBP: Input Bit Position, OBP: Output Bit Position)**

The initial permutation is just reversed after the 16th round.

### *2.5.1.2.2 DES Round Structure*

DES round is a little bit complicated. It deals with the input 64 bits block as two 32 bits blocks labelled L and R for left and right halves. As in all Feistel structure, the output from the round is defined as in equations 2.8 and 2.9 that is rewritten here

$$L_i = R_{i-1}$$

$$R_i = F_i(R_{i-1}, K_i) \oplus L_i$$

Where i subscript is the round order.

#### *2.5.1.2.3 DES round function F*

The round function of DES accepts 32 bits block of data and 48 bits key. The round function is performed on 4 stages,

1. Expansion permutation (E) of the input block form 32 bits into 48 bits by repeating 16 bits of the input block.
2. Add round key using XOR
3. Substitution choice using S-Boxes, the output of this stage is 32 bit block.
4. Permutation, (P) which rearranges the final 32 bits.

Figure 2.14 shows the details of one round of DES

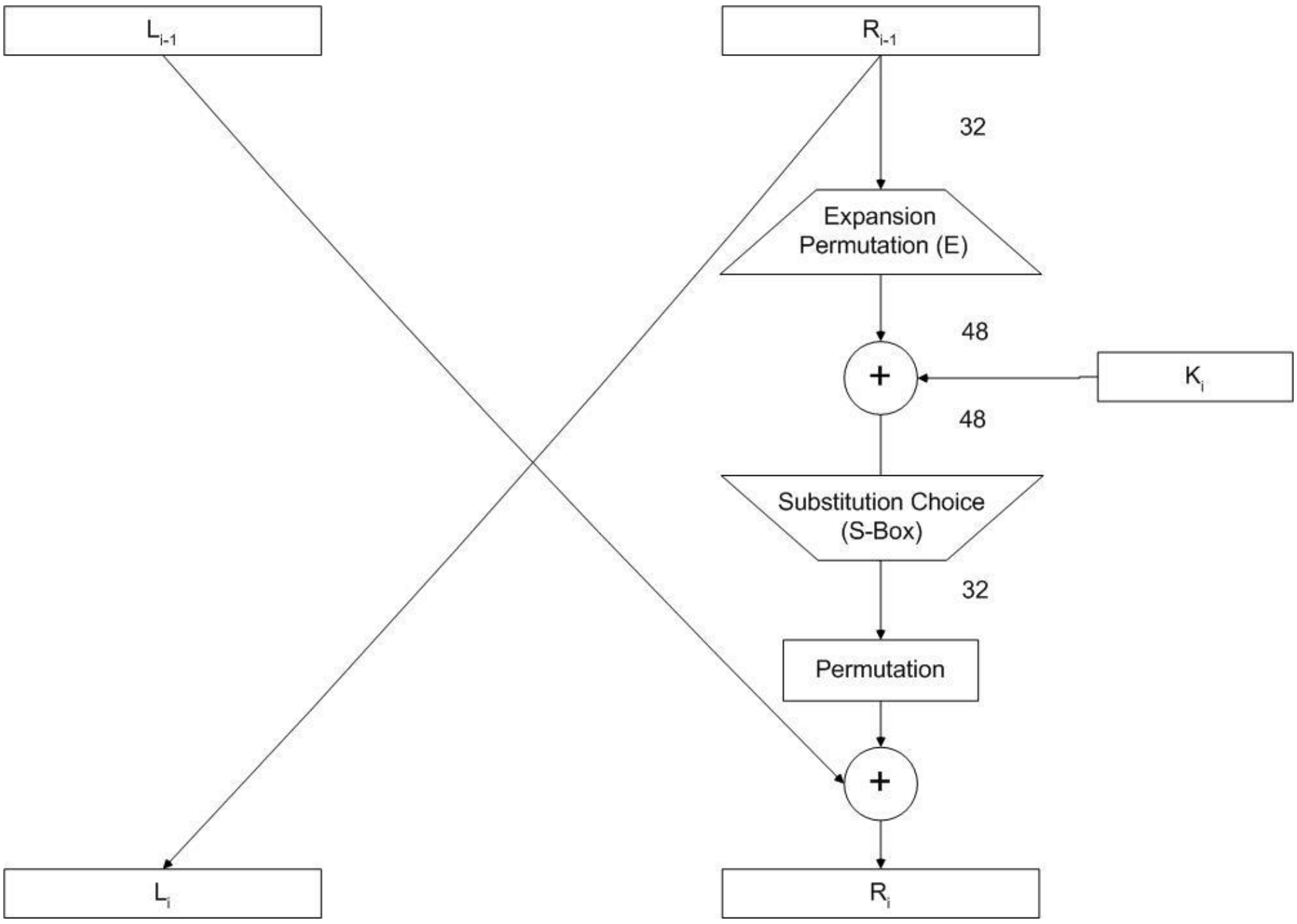

Figure 2.14: SINGLE DES ROUND

Each permutation above introduces some diffusion effect. The Expansion permutation (E) and the Permutation (P) are shown in tables 2.11 and 2.12 respectively

| OBP | 1 | 2 | 3 | 4 | 5 | 6 | 7 | 8 | 9 | 10 | 11 | 12 |
|---|---|---|---|---|---|---|---|---|---|---|---|---|
| IBP | 32 | 1 | 2 | 3 | 4 | 5 | 4 | 5 | 6 | 7 | 8 | 9 |

| OBP | 13 | 14 | 15 | 16 | 17 | 18 | 19 | 20 | 21 | 22 | 23 | 24 |
|---|---|---|---|---|---|---|---|---|---|---|---|---|
| IPB | 8 | 9 | 10 | 11 | 12 | 13 | 12 | 13 | 14 | 15 | 16 | 17 |

| OBP | 25 | 26 | 27 | 28 | 29 | 30 | 31 | 32 | 33 | 34 | 35 | 36 |
|---|---|---|---|---|---|---|---|---|---|---|---|---|
| IPB | 16 | 17 | 18 | 19 | 20 | 21 | 20 | 21 | 22 | 23 | 24 | 25 |

| OBP | 37 | 38 | 39 | 40 | 41 | 42 | 43 | 44 | 45 | 46 | 47 | 48 |
|---|---|---|---|---|---|---|---|---|---|---|---|---|
| IPB | 24 | 25 | 26 | 27 | 28 | 29 | 28 | 29 | 30 | 31 | 32 | 1 |

**Table 2.10 Expansion permutation.**

While the Permutation P is as follows

| OBP | 1 | 2 | 3 | 4 | 5 | 6 | 7 | 8 | 9 | 10 | 11 | 12 | 13 | 14 | 15 | 16 |
|---|---|---|---|---|---|---|---|---|---|---|---|---|---|---|---|---|
| IBP | 16 | 7 | 20 | 21 | 29 | 12 | 28 | 17 | 1 | 15 | 23 | 26 | 5 | 18 | 31 | 10 |

| OBP | 17 | 18 | 19 | 20 | 21 | 22 | 23 | 24 | 25 | 26 | 27 | 28 | 29 | 30 | 31 | 32 |
|---|---|---|---|---|---|---|---|---|---|---|---|---|---|---|---|---|
| IBP | 2 | 8 | 24 | 14 | 32 | 27 | 3 | 9 | 19 | 13 | 30 | 6 | 22 | 11 | 4 | 25 |

**Table 2.11 Permutation.**

The heart of the DES is its S-Box structure. S-Box is a nonlinear mapping between its input bits and its output bits. S-Box implementation was always a cheap way to implement complex function in GF($2^n$), an introduction to finite fields arithmetic is given in appendix A, for small values of n. It is simply can be implemented in hardware or software using a lookup table or in other words, indexed successive memory locations. In DES the 48 bits input is sub divided into 8 sub blocks 6 bits each. Each sub block is used to produce a 4 bit output using 6 x 4 s-box. The structure of the S-box is little bit complex. Each s-box is divided into 4 s-boxes 4 x 4 each. The first and last input bits select which sub s-box to use, and then the remaining 4 bits selects the output of the s-box. The result is a block of 32 bits length. Table 2.12 shows the S-Boxes of DES

| Input | | 0 | 1 | 2 | 3 | 4 | 5 | 6 | 7 | 8 | 9 | A | B | C | D | E | F |
|---|---|---|---|---|---|---|---|---|---|---|---|---|---|---|---|---|---|
| $S_1$ | $S_{11}$ | E | 4 | D | 1 | 2 | F | B | 8 | 3 | A | 6 | C | 5 | 9 | 0 | 7 |
| | $S_{12}$ | 0 | F | 7 | 4 | E | 2 | D | 1 | A | 6 | C | B | 9 | 5 | 3 | 8 |
| | $S_{13}$ | 4 | 1 | E | 8 | D | 6 | 2 | B | F | C | 9 | 7 | 3 | A | 5 | 0 |
| | $S_{14}$ | F | C | 8 | 2 | 4 | 9 | 1 | 7 | 5 | B | 3 | E | A | 0 | 6 | D |
| $S_2$ | $S_{21}$ | F | 1 | 8 | E | 6 | B | 3 | 4 | 9 | 7 | 2 | D | C | 0 | 5 | A |
| | $S_{22}$ | 3 | D | 4 | 7 | F | 2 | 8 | E | C | 0 | 1 | A | 6 | 9 | B | 5 |
| | $S_{23}$ | 0 | E | 7 | B | A | 4 | D | 1 | 5 | 8 | C | 6 | 9 | 3 | 2 | F |
| | $S_{24}$ | D | 8 | A | 1 | 3 | F | 4 | 2 | B | 6 | 7 | C | 0 | 5 | E | 9 |
| $S_3$ | $S_{31}$ | A | 0 | 9 | E | 6 | 3 | F | 5 | 1 | D | C | 7 | B | 4 | 2 | 8 |
| | $S_{32}$ | D | 7 | 0 | 9 | 3 | 4 | 6 | A | 2 | 8 | 5 | E | C | B | F | 1 |
| | $S_{33}$ | D | 6 | 4 | 9 | 8 | F | 3 | 0 | B | 1 | 2 | C | 5 | A | E | 7 |
| | $S_{34}$ | 1 | A | D | 0 | 6 | 9 | 8 | 7 | 4 | F | E | 3 | B | 5 | 2 | C |
| $S_4$ | $S_{41}$ | 7 | D | E | 3 | 0 | 6 | 9 | A | 1 | 2 | 8 | 5 | B | C | 4 | F |
| | $S_{42}$ | D | 8 | B | 5 | 6 | F | 0 | 3 | 4 | 7 | 2 | C | 1 | A | E | 9 |
| | $S_{43}$ | A | 6 | 9 | 0 | C | B | 7 | D | F | 1 | 3 | E | 5 | 2 | 8 | 4 |
| | $S_{44}$ | 3 | F | 0 | 6 | A | 1 | D | 8 | 9 | 4 | 5 | B | C | 7 | 2 | E |
| $S_5$ | $S_{51}$ | 2 | C | 4 | 1 | 7 | A | B | 6 | 8 | 5 | 3 | F | D | 0 | E | 9 |
| | $S_{52}$ | E | B | 2 | C | 4 | 7 | D | 1 | 5 | 0 | F | A | 3 | 9 | 8 | 6 |
| | $S_{53}$ | 4 | 2 | 1 | B | A | D | 7 | 8 | F | 9 | C | 5 | 6 | 3 | 0 | E |
| | $S_{54}$ | B | 8 | C | 7 | 1 | E | 2 | D | 6 | F | 0 | 9 | A | 4 | 5 | 3 |
| $S_6$ | $S_{61}$ | C | 1 | A | F | 9 | 2 | 6 | 8 | 0 | D | 3 | 4 | E | 7 | 5 | B |
| | $S_{62}$ | A | F | 4 | 2 | 7 | C | 9 | 5 | 6 | 1 | D | E | 0 | B | 3 | 8 |
| | $S_{63}$ | 9 | E | F | 5 | 2 | 8 | C | 3 | 7 | 0 | 4 | A | 1 | D | B | 6 |
| | $S_{64}$ | 4 | 3 | 2 | C | 9 | 5 | F | A | B | E | 1 | 7 | 6 | 0 | 8 | D |
| $S_7$ | $S_{71}$ | 4 | B | 2 | E | F | 0 | 8 | D | 3 | C | 9 | 7 | 5 | A | 6 | 1 |
| | $S_{72}$ | D | 0 | B | 7 | 4 | 9 | 1 | A | E | 3 | 5 | C | 2 | F | 8 | 6 |
| | $S_{73}$ | 1 | 4 | B | D | C | 3 | 7 | E | A | F | 6 | 8 | 0 | 5 | 2 | 9 |
| | $S_{74}$ | 6 | B | D | 8 | 1 | 4 | A | 7 | 9 | 5 | 0 | F | E | 2 | 3 | C |
| $S_8$ | $S_{81}$ | D | 2 | 8 | 4 | 6 | F | B | 1 | A | 9 | 3 | E | 5 | 0 | C | 7 |
| | $S_{82}$ | 1 | F | D | 8 | A | 3 | 7 | 4 | C | 5 | 6 | B | 0 | F | 9 | 2 |
| | $S_{83}$ | 7 | B | 4 | 1 | 9 | C | E | 2 | 0 | 6 | A | D | F | 3 | 5 | 8 |
| | $S_{84}$ | 2 | 1 | E | 7 | 4 | A | 8 | D | F | C | 9 | 0 | 3 | 5 | 6 | B |

**Table 2.12 32 (8 x 4) DES S-Boxes**

Figure 2.15 shows how the blocks are subdivided and S-boxes used

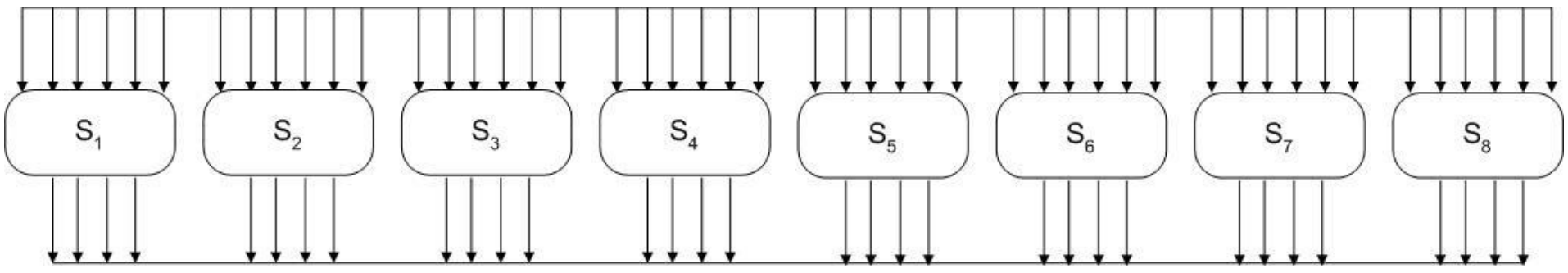

Figure 2.15: DES S-BOXES

#### *2.5.1.3 DES Key Generation*

DES uses a 48 bits round key generated from the original 56 bits key of the cipher for each round. The key is entered in the form of 64 bits. By ignoring the last bit of each byte, 64 bits are 8 bytes, of the user supplied key, a 56 bits key is obtained. The 56 bits key is first subjected to a permutation called permutation choice 1 (PC1). Then for each round, the 56 bit key is divided into 2 halves 28 bits each. Each half is rotated left 1 or 2 bits depending on the round. These shifted halves are the input to the next round. The 2 halves are then subjected to another permutation where only 48 bits are selected. This permutation is called permutation choice 2 (PC2) shown in table 2.13.

| OBP | 1 | 2 | 3 | 4 | 5 | 6 | 7 | 8 | 9 | 10 | 11 | 12 |
|---|---|---|---|---|---|---|---|---|---|---|---|---|
| IBP | 14 | 17 | 11 | 24 | 1 | 5 | 3 | 28 | 15 | 6 | 21 | 10 |

| OBP | 13 | 14 | 15 | 16 | 17 | 18 | 19 | 20 | 21 | 22 | 23 | 24 |
|---|---|---|---|---|---|---|---|---|---|---|---|---|
| IPB | 23 | 19 | 12 | 4 | 26 | 8 | 16 | 7 | 27 | 20 | 13 | 2 |

| OBP | 25 | 26 | 27 | 28 | 29 | 30 | 31 | 32 | 33 | 34 | 35 | 36 |
|---|---|---|---|---|---|---|---|---|---|---|---|---|
| IPB | 41 | 52 | 31 | 37 | 47 | 55 | 30 | 40 | 51 | 45 | 33 | 48 |

| OBP | 37 | 38 | 39 | 40 | 41 | 42 | 43 | 44 | 45 | 46 | 47 | 48 |
|---|---|---|---|---|---|---|---|---|---|---|---|---|
| IPB | 44 | 49 | 39 | 56 | 34 | 35 | 46 | 42 | 50 | 36 | 29 | 32 |

**Table 2.13: Permutation Choice 2.**

Figure 2.16 shows one round of key generation.

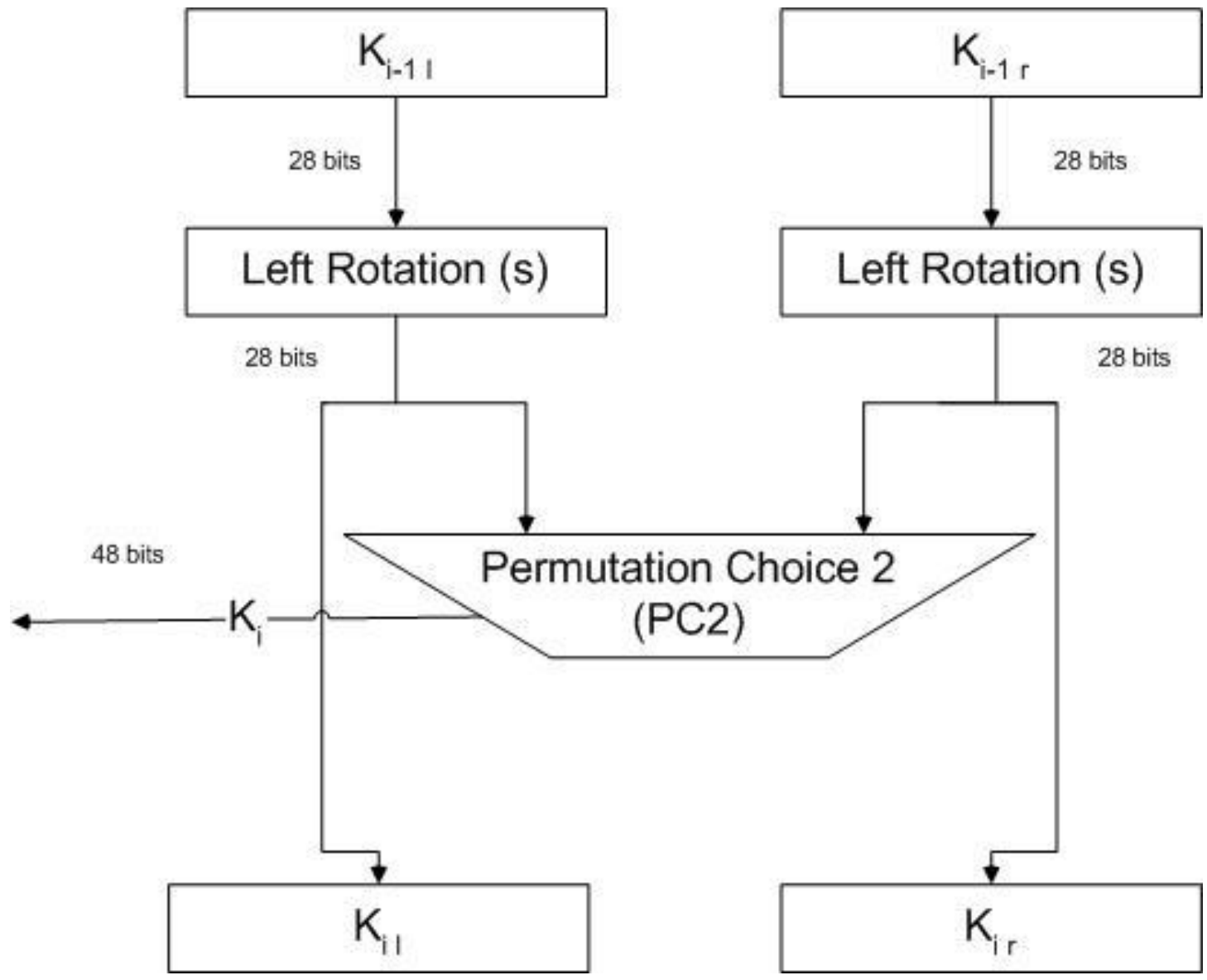

Figure 2.16: DES ROUND KEY SCHEDULING

DES uses 16 rounds and hence 16 round sub keys are needed.

#### *2.5.1.4 DES Decryption*

DES decryption is done by passing the ciphertext through the structure with reversed order key. The initial and final permutations are inverse of each other and hence they do not affect the reversibility of the Feistel structure used for DES.

#### *2.5.1.5 DES Security and the need for a replacement*

DES was believed to be very secure until the early 1990s. It was not until that date when new attacks namely differential cryptanalysis (for details refer to the next chapter) were discovered, the only possible attack to DES before was brute force.

By the end of the 1990s, it was noticed that computer power is increasing and brute force against DES is becoming feasible. It was clear that a replacement for DES was needed. Rijndael was selected as the AES in Oct-2000 and issued as FIPS, Federal Information Processing Standard, publication 197 in Nov-2001 (Rijndael will be presented at the end of this chapter). A cipher which is widely accepted and used especially in open source environments is blowfish by Bruce Schneier, one of the cryptography leaders [28].

### 2.5.2 Blowfish

Blowfish was an attempt from Schneier to replace DES with an open, unpatented free, cipher for usage by the entire world. It is an urgent need for such a cipher as the US government restricts export of products using high key length ciphers to certain countries and regions in the world. Also patented ciphers are not to be used without permission from their owners. Hence Blowfish gained a great interest as being one of the first attempts to produce a free cipher. It also presents a new design method which will be described in section 8.2.5.6.

#### *2.5.2.1 Blowfish cipher description*

Blowfish is a variable-length key, 64-bit block cipher. The algorithm consists of two parts: a key expansion part and a data- encryption part. Key expansion converts a key of at most 448 bits into several sub key arrays totaling 4168 bytes.

Data encryption occurs via a 16-round Feistel network. Each round consists of a key-dependent permutation, and a key- and data-dependent substitution. All operations are XORs and additions on 32-bit words. The only additional operations are four indexed array data lookups per round.

Blowfish uses a large number of sub keys. These keys must be pre-computed before any data encryption or decryption. The key field is represented as an array called P array which holds 18 sub key ($P_1$, $P_2$,..., $P_{18}$) 32 bits each. There are 4 substitution boxes ($S_1$, $S_2$, $S_3$, $S_4$) 8 x 32 bits each.

### 2.5.2.2 Blowfish Encryption:

Blowfish is a Feistel network consisting of 16 rounds. The input is a 64-bit data element, x which is divided into two 32-bit halves: $x_L$, $x_R$. The flowing pseudo code shows how blow fish encryption is performed

```
For i = 1 to 16
    begin
        xL = xL XOR Pi
        xR = F(xL) XOR xR
        Swap xL and xR
    end
Swap xL and xR (Undo the last swap.)
xR = xR XOR P17
xL = xL XOR P18
Recombine xL and xR
```

### 2.5.2.3 Blow fish core Function F

The core function is applied on the left half as follows, Divide $x_L$ into four eight-bit quarters: a, b, c, and d and then the operations of XOR and addition mod $2^{32}$ are performed as shown in figure 2.17

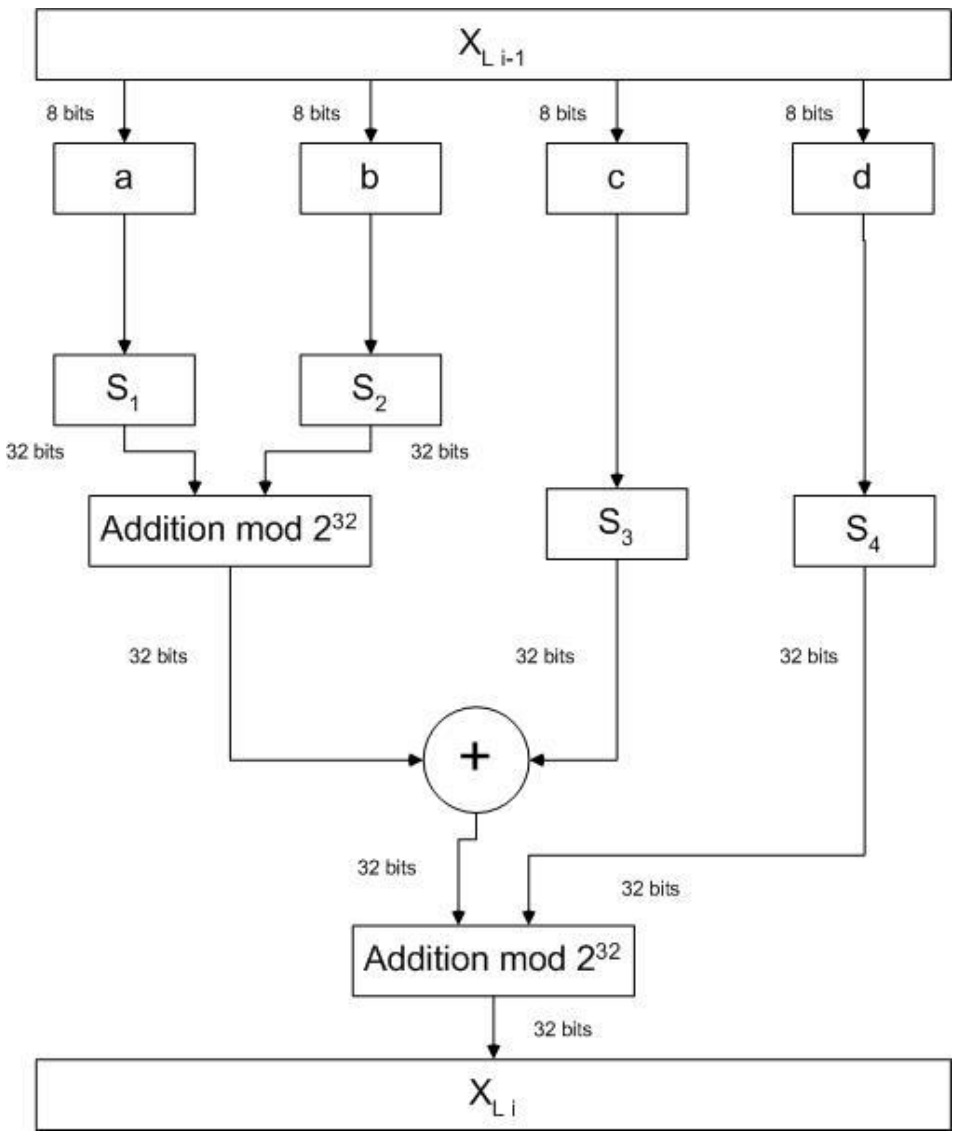

Figure 2.17: BLOWFISH CORE FUNCTION

#### *2.5.2.4 Blowfish Decryption*

Decryption is exactly the same as encryption, except that $P_1$, $P_2$,..., $P_{18}$ are used in the reverse order.

Implementations of Blowfish that require the fastest speeds should unroll the loop and ensure that all sub keys are stored in cache.

#### *2.5.2.5 Blowfish key scheduling and S-Box generation*

The sub keys are calculated using the Blowfish algorithm. The exact method is as follows:

1. Initialize first the P-array and then the four S-boxes, in order, with a fixed string. This string consists of the hexadecimal digits of pi (less the initial 3). For example:

   $P_1$ = 0x243f6a88

   $P_2$ = 0x85a308d3

   $P_3$ = 0x13198a2e

   $P_4$ = 0x03707344
2. XOR P1 with the first 32 bits of the key, XOR P2 with the second 32-bits of the key, and so on for all bits of the key (possibly up to P14). Repeatedly cycling through the key bits until the entire P-array has been XORed with key bits. (For every short key, there is at least one equivalent longer key; for example, if A is a 64-bit key, then AA, AAA, etc., are equivalent keys.)
3. Encrypt the all-zero string with the Blowfish algorithm, using the sub keys described in steps (1) and (2).
4. Replace $P_1$ and $P_2$ with the output of step (3).
5. Encrypt the output of step (3) using the Blowfish algorithm with the modified sub keys.
6. Replace $P_3$ and $P_4$ with the output of step (5).
7. Continue the process, replacing all entries of the P- array, and then all four S-boxes in order, with the output of the continuously-changing Blowfish algorithm.

In total, 521 iterations (9, 18/2, for the P array + 512, 256*4/2, for the s-boxes) are required to generate all required sub keys. Applications can store the sub keys rather than execute this derivation process multiple times.

S-Box design in blow fish follows a random key dependent approach yielding a structure very hard to analyze for attacks like linear or differential attacks.

#### *2.5.2.6 Blowfish security*

The key dependent S-boxes and sub keys, generated using cipher itself, makes analysis very difficult.

Provided key is large enough, brute-force key search is not practical, especially with the high key scheduling cost.

### 2.5.3 Rijndael

Rijndael [5], [6], and [29] is the new AES standard block cipher which replaced DES. The cipher is simple in structure and yet very secure. The designers of the cipher used a methodology to ensure great confusion and diffusion effects and their strategy was to use byte operations which are very fast to implement both in Hardware and Software. The structure of Rijndael follows a substitution permutation network structure but first the input block is arranged into a square of bytes. The structure is inspired by a previous cipher designed by the authors of Rijndael and Knudsen called square cipher [30].

#### *2.5.3.1 Rijndael Cipher Structure*

Rijndael is an iterated cipher that accepts data block of 128 bits and a key of 128, 192, or 256 bits. The number of rounds depends on the key length. The basic version of 128 bits key is described here.

In this version the cipher consists of 9 rounds plus an extra last round.

Each regular round involves four steps. First is the Byte Substitution step, where each byte of the block is replaced by its substitute in an S-box.

The S-box itself was very cleverly designed to overcome known attacks exploiting S-Boxes of DES and similar ciphers, namely differential, linear, and interpolation attacks.

The S-box here is a $GF(2^8) \rightarrow GF(2^8)$ mapping. They first used the inverse of input in $GF(2^8)$.

$a(x)=x^{-1}$ in $GF(2^8)$ with polynomial basis with the irreducible polynomial

$$m(x) = x^8 + x^4 + x^3 + x + 1$$

0 is mapped to itself as it does not have an inverse.

This mapping is already shown to be non vulnerable to linear or differential cryptanalysis by Nyberg [31].

Then an affine transform is performed to rise up the degree of the overall system to overcome the interpolation attack [32].

The transform used is

$$b(x) = (x^7 + x^6 + x^2 + x) + a(x)(x^7 + x^6 + x^5 + x^4 + 1) \bmod (x^8 + 1)$$

Rijndael designers performed the above transform on all possible values of 8 bits and came up with am S-Box which is 8 x 8 mapping for the sake of performance. The S-box and it inverse are shown in tables 2.14 and 2.15 respectively (hexadecimal values are presented). The tables should be read from left to right up down for the input output mapping performed by s-boxes for inputs in sequence from 00 to ff (in hexadecimal).

| | | | | | | | | | | | | | | | |
|---|---|---|---|---|---|---|---|---|---|---|---|---|---|---|---|
| 63 | 7C | 77 | 7B | F2 | 6B | 6F | C5 | 30 | 01 | 67 | 2B | FE | D7 | AB | 76 |
| CA | 82 | C9 | 7D | FA | 59 | 47 | F0 | AD | D4 | A2 | AF | 9C | A4 | 72 | C0 |
| B7 | FD | 93 | 26 | 36 | 3F | F7 | CC | 34 | A5 | E5 | F1 | 71 | D8 | 31 | 15 |
| 04 | C7 | 23 | C3 | 18 | 96 | 05 | 9A | 07 | 12 | 80 | E2 | EB | 27 | B2 | 75 |
| 09 | 83 | 2C | 1A | 1B | 6E | 5A | A0 | 52 | 3B | D6 | B3 | 29 | E3 | 2F | 84 |
| 53 | D1 | 00 | ED | 20 | FC | B1 | 5B | 6A | CB | BE | 39 | 4A | 4C | 58 | CF |
| D0 | EF | AA | FB | 43 | 4D | 33 | 85 | 45 | F9 | 02 | 7F | 50 | 3C | 9F | A8 |
| 51 | A3 | 40 | 8F | 92 | 9D | 38 | F5 | BC | B6 | DA | 21 | 10 | FF | F3 | D2 |
| CD | 0C | 13 | EC | 5F | 97 | 44 | 17 | C4 | A7 | 7E | 3D | 64 | 5D | 19 | 73 |
| 60 | 81 | 4F | DC | 22 | 2A | 90 | 88 | 46 | EE | B8 | 14 | DE | 5E | 0B | DB |
| E0 | 32 | 3A | 0A | 49 | 06 | 24 | 5C | C2 | D3 | AC | 62 | 91 | 95 | E4 | 79 |
| E7 | C8 | 37 | 6D | 8D | D5 | 4E | A9 | 6C | 56 | F4 | EA | 65 | 7A | AE | 08 |
| BA | 78 | 25 | 2E | 1C | A6 | B4 | C6 | E8 | DD | 74 | 1F | 4B | BD | 8B | 8A |
| 70 | 3E | B5 | 66 | 48 | 03 | F6 | 0E | 61 | 35 | 57 | B9 | 86 | C1 | 1D | 9E |
| E1 | F8 | 98 | 11 | 69 | D9 | 8E | 94 | 9B | 1E | 87 | E9 | CE | 55 | 28 | DF |
| 8C | A1 | 89 | 0D | BF | E6 | 42 | 68 | 41 | 99 | 2D | 0F | B0 | 54 | BB | 16 |

**Table 2.14: Rijndael S-Box**

| | | | | | | | | | | | | | | | |
|---|---|---|---|---|---|---|---|---|---|---|---|---|---|---|---|
| 52 | 09 | 6A | D5 | 30 | 36 | A5 | 38 | BF | 40 | A3 | 9E | 81 | F3 | D7 | FB |
| 7C | E3 | 39 | 82 | 9B | 2F | FF | 87 | 34 | 8E | 43 | 44 | C4 | DE | E9 | CB |
| 54 | 7B | 94 | 32 | A6 | C2 | 23 | 3D | EE | 4C | 95 | 0B | 42 | FA | C3 | 4E |
| 08 | 2E | A1 | 66 | 28 | D9 | 24 | B2 | 76 | 5B | A2 | 49 | 6D | 8B | D1 | 25 |
| 72 | F8 | F6 | 64 | 86 | 68 | 98 | 16 | D4 | A4 | 5C | CC | 5D | 65 | B6 | 92 |
| 6C | 70 | 48 | 50 | FD | ED | B9 | DA | 5E | 15 | 46 | 57 | A7 | 8D | 9D | 84 |
| 90 | D8 | AB | 00 | 8C | BC | D3 | 0A | F7 | E4 | 58 | 05 | B8 | B3 | 45 | 06 |
| D0 | 2C | 1E | 8F | CA | 3F | 0F | 02 | C1 | AF | BD | 03 | 01 | 13 | 8A | 6B |
| 3A | 91 | 11 | 41 | 4F | 67 | DC | EA | 97 | F2 | CF | CE | F0 | B4 | E6 | 73 |
| 96 | AC | 74 | 22 | E7 | AD | 35 | 85 | E2 | F9 | 37 | E8 | 1C | 75 | DF | 6E |
| 47 | F1 | 1A | 71 | 1D | 29 | C5 | 89 | 6F | B7 | 62 | 0E | AA | 18 | BE | 1B |
| FC | 56 | 3E | 4B | C6 | D2 | 79 | 20 | 9A | DB | C0 | FE | 78 | CD | 5A | F4 |
| 1F | DD | A8 | 33 | 88 | 07 | C7 | 31 | B1 | 12 | 10 | 59 | 27 | 80 | EC | 5F |
| 60 | 51 | 7F | A9 | 19 | B5 | 4A | 0D | 2D | E5 | 7A | 9F | 93 | C9 | 9C | EF |
| A0 | E0 | 3B | 4D | AE | 2A | F5 | B0 | C8 | EB | BB | 3C | 83 | 53 | 99 | 61 |
| 17 | 2B | 04 | 7E | BA | 77 | D6 | 26 | E1 | 69 | 14 | 63 | 55 | 21 | 0C | 7D |

**Table 2.15: Rijndael Inverse S-Box**

The input block which is 128 bits or 16 bytes is arranged in the form of a 4 x 4 square byte matrix and substitution is made on each byte.

The next stage of Rijndael is a shift row operation where each row of the square 4 x 4 substitution is rotated as follows

| From | | | | | To | | | |
|---|---|---|---|---|---|---|---|---|
| 1 | 5 | 9 | 13 | | 1 | 5 | 9 | 13 |
| 2 | 6 | 10 | 14 | | 6 | 10 | 14 | 2 |
| 3 | 7 | 11 | 15 | | 11 | 15 | 3 | 7 |
| 4 | 8 | 12 | 16 | | 16 | 4 | 8 | 12 |

Where the first row is not shifted while the second is rotated one byte to the left, the third two bytes, and the fourth 3 bytes.

This operation introduces permutation and hence diffusion.

The third stage is a Mix Column operation where addition and multiplication in GF($2^8$) of each column is performed resulting that each byte is a function in all the column four bytes. The mix column is defined as a matrix multiplication in GF($2^8$) with polynomial basis, the same polynomial used to generate the s-boxes.

The result of the shift row matrix is multiplicated with the following matrix

| | | | |
|---|---|---|---|
| 2 | 3 | 1 | 1 |
| 1 | 2 | 3 | 1 |
| 1 | 1 | 2 | 3 |
| 3 | 1 | 1 | 2 |

And the result is finally mixed with the round key, 128 bits, using XOR.

Before introducing the plaintext block to the first round it is mixed with the user input key and after the last round an extra round without the mix column stage is applied.

The entire cipher components are reversible and hence the entire cipher is reversible.

Figure 2.18 represents a 3D illustration of Rijndael round, while figure 2.19 is a conventional illustration which represents the SPN structure of Rijndael.

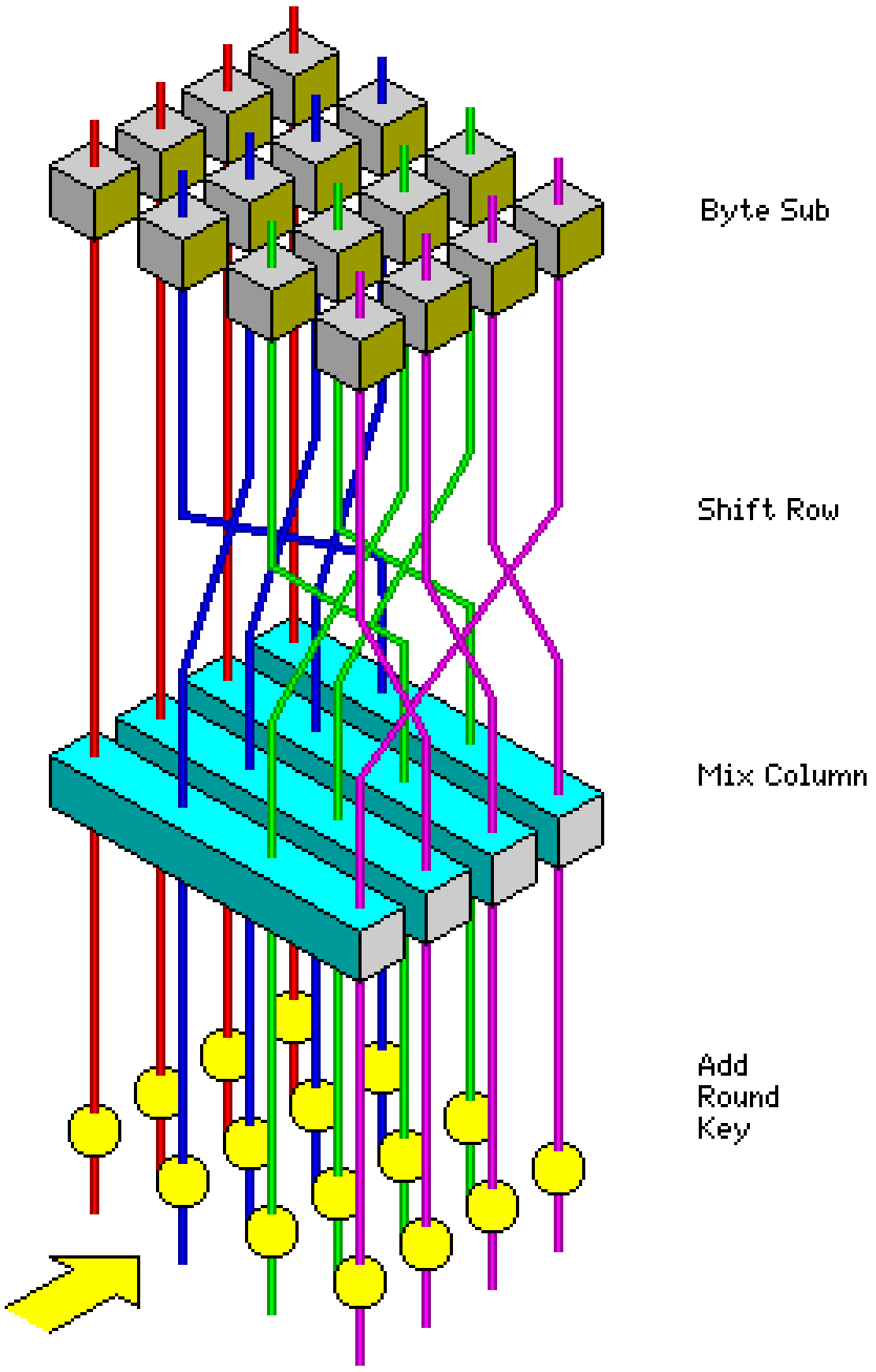

Figure 2.18: 3D ILLUSTRATION OF A RIJNDAEL ROUND

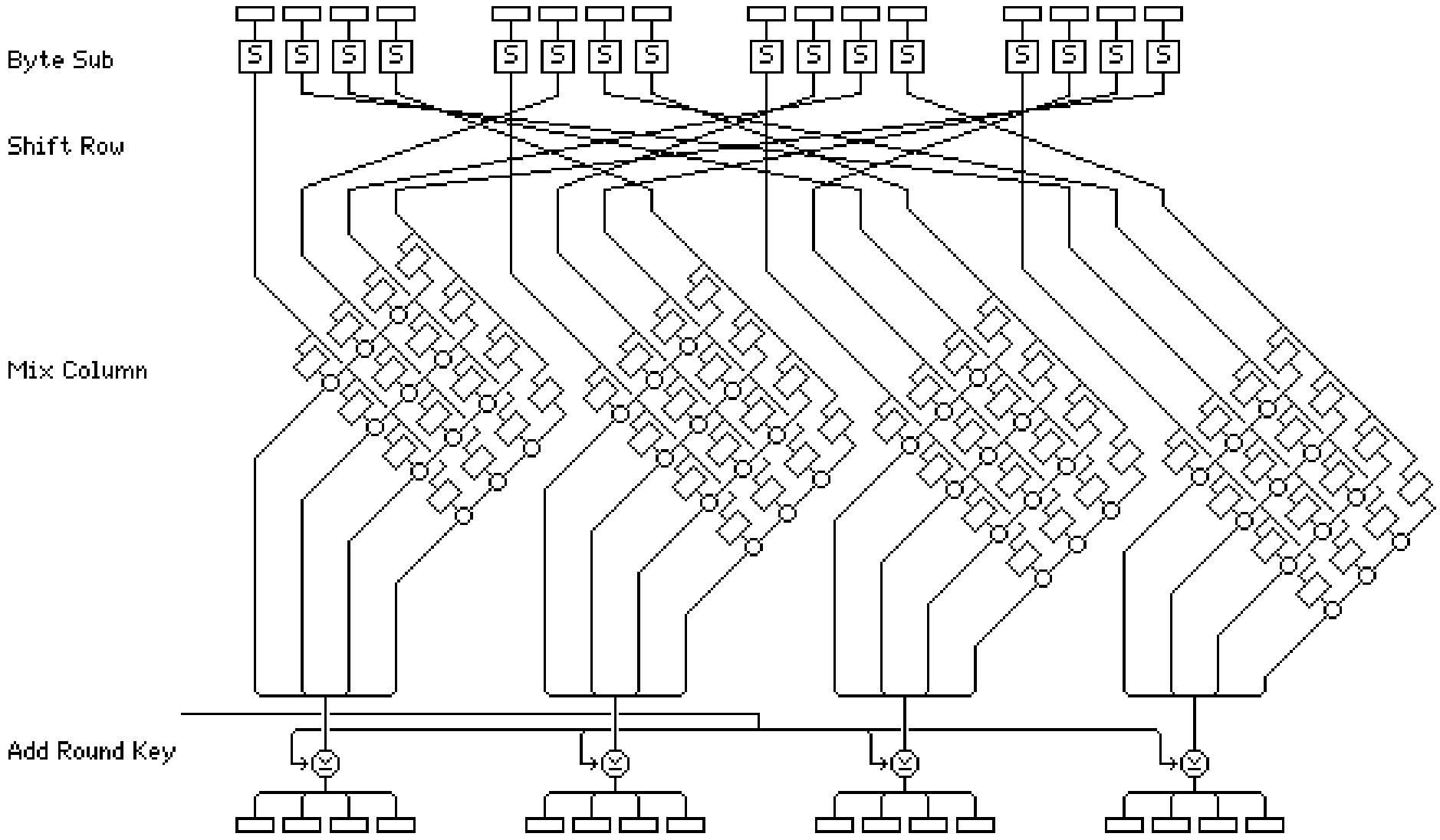

Figure 2.19: RIJNDAEL ROUND

#### *2.5.3.2 Rijndael Key Scheduling*

Rijndael needs a 128 bit key for each round. The sub key material, which consists of all the round keys in order, consists of the original key, followed by stretches.

Each stretch is of the length of the original key, consisting of 4 words four-byte each such that each word is the XOR of the preceding four byte word and either corresponding word in the previous stretch for all stretch word except the first one.

For the first word in a stretch, the last word of the previous stretch is first rotated one byte to the left, and then its bytes are transformed using the S-box from the Byte Substitution step, and then a round-dependent constant is XORed to its first byte. The result is XORed with the first word of thje previous stretch, see figure 2.20

The round constants are generated as successive powers of 3 in GF($2^8$):

| | | | | | | | |
|---|---|---|---|---|---|---|---|
| 1 | 2 | 4 | 8 | 16 | 32 | 64 | 128 |
| 27 | 54 | 108 | 216 | 171 | 77 | 154 | 47 |
| 94 | 188 | 99 | 198 | 151 | 53 | 106 | 212 |
| 179 | 125 | 250 | 239 | 197 | 145 | 57 | 114 |
| 228 | 211 | 189 | 97 | ... | | | |

Successive powers of 3 in GF($2^8$), unlike 2, include all the values from 1 to 255, 3 is called primitive root, and thus several implementations of Rijndael use tables of the powers of 3, and the inverse table giving the discrete logarithm in that field, to facilitate calculations, but the round constants are still the powers of 2.

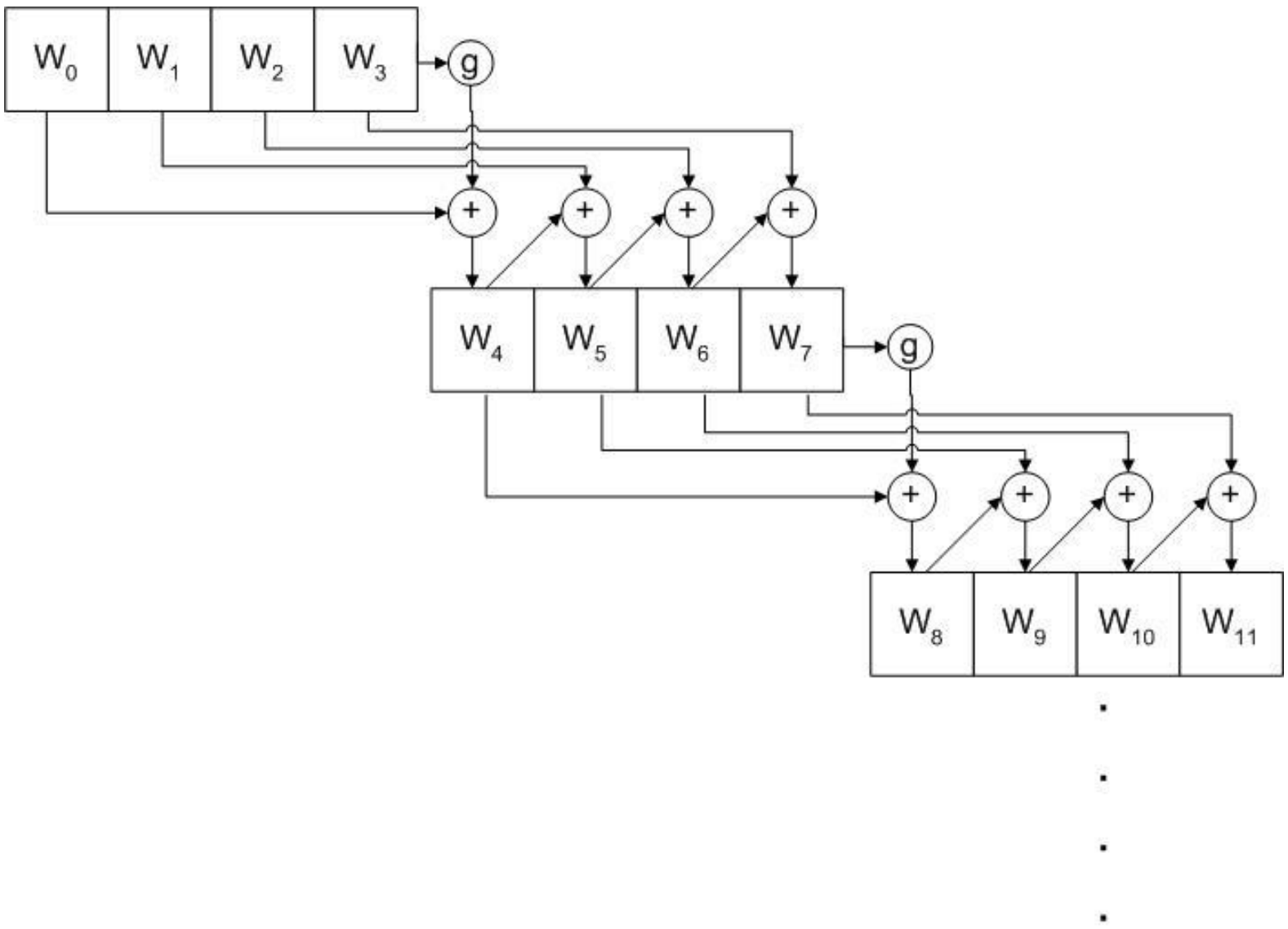

Figure 2.20: RIJNDAEL KEY SCHEDULING

#### *2.5.3.3 Rijndael Decryption*

Decryption in Rijndael is performed by going backward in the structure with each operation reversed and keys in reverse order just like the basic SPN described.

#### *2.5.3.4 Rijndael Security*

Rijndael is believed for the best knowledge to be very secure. Its design was based to defeat known cryptanalysis techniques known. A dedicated attack is the square attack developed by the authors of Rijndael, but it seems to attack only reduced versions up to 4 rounds.

The main threat to Rijndael may be from the discovery of that Rijndael can be presented as a set of algebraic equations. Hence this property can be used with known plaintext to attack Rijndael. But the number of equations to be solved are quit large and no way has been discovered to solve them yet.

# Chapter 3: Attacks on Block Ciphers

## 3.1 Introduction

Major advances in the field of cryptanalysis have been developed. The DES cipher which was believed to be very secure faced a major challenge by the discovery of the differential cryptanalysis in 1991 [33]. Later linear cryptanalysis was discovered and applied to DES. More and more complicated and successful attacks were developed and applied to block cipher previously known to be secure. In the following sections, a detailed over view of attacks and some examples are given. An extension to those attacks using artificial intelligence techniques will be addressed in the following chapters.

## 3.2 Attack Complexity

The complexity of the attack is a measure of how much resources is needed to mount the attack. The resources here mean the computational power or time and the amount of information needed in the form of plaintext, ciphertext, or both. It is always assumed that he algorithm of the encryption is known. Complexity can be measured by the number of encryptions and/ or decryptions needed to mount the attack.

For example a cipher with b bits secret key can be attacked by trying all the possible values of the key which is called a brute force attack. Of course at least one plaintext – ciphertext pair has to be known to verify each value of the key. This attack is said to have $2^b$ complexity.

The required information is always used to classify the type of the attack as the following section explains.

## 3.3 Types of Attacks

Attacks on block ciphers can be classified based on the amount of required information to mount the attack. The following is a list of these types of attacks [2].

### 3.3.1 Known ciphertext only attack

In this attack, the cryptanalyst only knows the encryption algorithm and the ciphertext to be decoded. The statistical attacks on classical ciphers described in chapter 2 are examples of known ciphertext only attacks. All modern block ciphers defeat this attack.

### 3.3.2 Known plaintext attack

In this attack, the cryptanalyst knows the encryption algorithm the ciphertext to be decoded, and one ore more plaintext – ciphertext pairs generated using the secret key. This type of information can be easily gained by previous knowledge of a certain pattern to occur in the ciphertext like in the case of encrypting an image file with predetermined structure and headers. Linear cryptanalysis is an example of known plaintext attack. Interpolation attack is another example, detailed in section 3.5.

### 3.3.3 Chosen plaintext attack

In this type of attack, the cryptanalyst is able to select certain plaintext and obtain the corresponding ciphertext using the secret key to be revealed. A good example of this type of attack is deferential cryptanalysis described later in section 3.4. The attacker chooses plaintext patterns that exploit the cipher structure and help to discover the key.

### 3.3.4 Chosen ciphertext and chosen text attacks

These types of attacks are less likely to be used or to obtain the information needed in an easy way, but there still a possibility to exploit cipher structure more efficiently if such information can be available.

A cipher is believed to be secure if it passes the known plaintext attack.

## 3.4 Differential Cryptanalysis

In this section, the application of differential cryptanalysis to the basic SPN cipher is described [7].

### 3.4.1 Overview of the Differential Attack

Differential cryptanalysis was developed by Biham and Shamir in early 1990s primarily for DES [34], [35], and [33].

Differential cryptanalysis exploits the high probability of certain occurrences of plaintext differences and differences into the last round of an iterated cipher.

For example, consider a system with input $X = [X1\ X2\ ...\ Xn]$ and output $Y = [Y1\ Y2\ ...\ Yn]$. Let two inputs to the system be $X'$ and $X''$ with the corresponding outputs $Y'$ and $Y''$, respectively. The input difference is given by $\Delta X = X' \oplus X''$ where "$\oplus$" represents a bit-wise exclusive-OR of the $n$-bit vectors and, hence,

$\Delta X = [\Delta X_1\ \Delta X_2\ ...\ \Delta X_n]$

Where $\Delta X_i = X_i' \oplus X_i''$ representing the $i^{th}$ bit of $X'$ and $X''$, respectively.

Similarly, $\Delta Y = Y' \oplus Y''$ is the output difference and

$\Delta Y = [\Delta Y_1\ \Delta Y_2\ ...\ \Delta Y_n]$

Where $\Delta Y_i = Y_i' \oplus Y_i''$

In an ideally randomizing cipher, the probability that a particular output difference $\Delta Y$ occurs given a particular input difference $\Delta X$ is $1/2^n$ where $n$ is the number of bits of $X$. Differential cryptanalysis seeks to exploit a scenario where a particular $\Delta Y$ occurs given a particular input difference $\Delta X$ with a very high probability $P_D$ (i.e., much greater than $1/2^n$). The pair ($\Delta X$, $\Delta Y$) is referred to as a *differential*. Differential cryptanalysis is a chosen plaintext attack, meaning that, the attacker is able to select inputs and examine outputs in an attempt to derive the key.

### 3.4.2 Differential Cryptanalysis of the basic SPN

In this section, the construction of a differential ($\Delta X$, $\Delta Y$) involving plaintext bits as represented by $X$ and the input to the last round of the cipher as represented by $Y$ for the basic SPN described in chapter 2 is investigated.

For the SPN structure, the only nonlinear component is the S-Box, and the focus of the attack is on that component. Each S-Box is studied for its *differential distribution.* Then by combining all the *active S-box*es differential distributions based on the SPN structure, a cipher *differential characteristics* is constructed.

Using the high likely differential characteristic gives the opportunity to exploit information coming into the last round of the cipher to derive bits from the last round sub-keys.

To find an S-Box difference distribution, the input and output differences of the S-boxes is considered in order to determine a high probability difference pair.

#### *3.4.2.1 Analyzing the Cipher Components*

Consider the 4×4 S-box representation of Figure 3.1 with input $X = [X1\ X2\ X3\ X4]$ and output $Y = [Y1\ Y2\ Y3\ Y4]$.

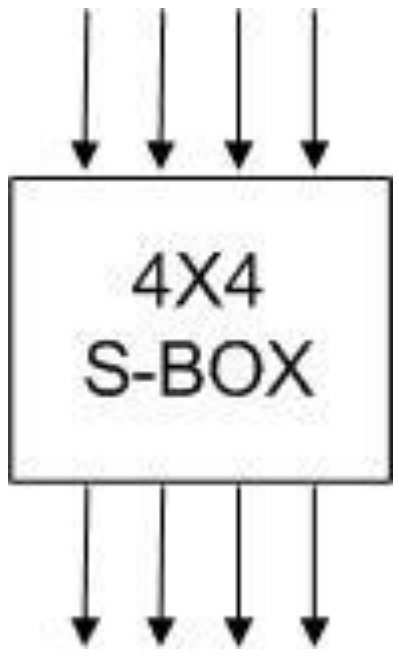

Figure 3.1: S-BOX MAPPING

All difference pairs of an S-box, ($\Delta X$, $\Delta Y$), can be examined and the probability of $\Delta Y$ given $\Delta X$ can be derived by considering input pairs ($X'$, $X''$) such that $X' \oplus X'' = \Delta X$. Since the ordering of the pair is not relevant, for a 4×4 S-box only consider all 16 values for $X'$ and then the value of $\Delta X$ constrains the value of $X''$ to be $X'' = X' \oplus \Delta X$.

The complete data for an S-box is tabulated in a *difference distribution table* in which the rows represent $\Delta X$ values (in hexadecimal) and the columns represent $\Delta Y$ values (in hexadecimal). The difference distribution table for the S-box $S_{11}$ from chapter 2 is given in Table 3.1. Each element of the table represents the number of occurrences of the corresponding output difference $\Delta Y$ value given the input difference $\Delta X$. Note that, besides the special case of ($\Delta X = 0$, $\Delta Y = 0$), the largest value in the table is 8, corresponding to $\Delta X$ = B and $\Delta Y = 2$. Hence, the probability that $\Delta Y = 2$ given an arbitrary pair of input values that satisfy $\Delta X$ = B is 8/16. The smallest value in the table is 0 and occurs for many difference pairs. In this case, the probability of the $\Delta Y$ value occurring given the $\Delta X$ value is 0.

| Input Difference \ Output Difference | 0 | 1 | 2 | 3 | 4 | 5 | 6 | 7 | 8 | 9 | A | B | C | D | E | F |
|---|---|---|---|---|---|---|---|---|---|---|---|---|---|---|---|---|
| 0 | 16 | 0 | 0 | 0 | 0 | 0 | 0 | 0 | 0 | 0 | 0 | 0 | 0 | 0 | 0 | 0 |
| 1 | 0 | 0 | 0 | 2 | 0 | 0 | 0 | 2 | 0 | 0 | 4 | 0 | 4 | 2 | 0 | 0 |
| 2 | 0 | 0 | 0 | 2 | 0 | 6 | 2 | 2 | 0 | 2 | 0 | 0 | 0 | 0 | 2 | 0 |
| 3 | 0 | 0 | 2 | 0 | 2 | 0 | 0 | 0 | 0 | 4 | 2 | 0 | 2 | 0 | 0 | 4 |
| 4 | 0 | 0 | 0 | 2 | 0 | 0 | 6 | 0 | 0 | 2 | 0 | 4 | 2 | 0 | 0 | 0 |
| 5 | 0 | 4 | 0 | 0 | 0 | 2 | 2 | 0 | 0 | 0 | 4 | 0 | 2 | 0 | 0 | 2 |
| 6 | 0 | 0 | 4 | 0 | 4 | 0 | 0 | 0 | 0 | 0 | 0 | 0 | 2 | 2 | 2 | 2 |
| 7 | 0 | 0 | 2 | 2 | 2 | 0 | 2 | 0 | 0 | 2 | 2 | 0 | 0 | 0 | 0 | 4 |
| 8 | 0 | 0 | 0 | 0 | 0 | 0 | 2 | 2 | 0 | 0 | 0 | 4 | 0 | 4 | 2 | 2 |
| 9 | 0 | 2 | 0 | 0 | 2 | 0 | 0 | 4 | 2 | 0 | 2 | 2 | 2 | 0 | 0 | 0 |
| A | 0 | 2 | 2 | 0 | 0 | 0 | 0 | 0 | 6 | 0 | 0 | 2 | 0 | 0 | 4 | 0 |
| B | 0 | 0 | 8 | 0 | 0 | 2 | 0 | 2 | 0 | 0 | 0 | 0 | 0 | 2 | 0 | 2 |
| C | 0 | 2 | 0 | 0 | 2 | 2 | 2 | 0 | 0 | 0 | 0 | 2 | 0 | 6 | 0 | 0 |
| D | 0 | 4 | 0 | 0 | 0 | 0 | 0 | 4 | 2 | 0 | 2 | 0 | 2 | 0 | 2 | 0 |
| E | 0 | 0 | 2 | 4 | 2 | 0 | 0 | 0 | 6 | 0 | 0 | 0 | 0 | 0 | 2 | 0 |
| F | 0 | 2 | 0 | 0 | 6 | 0 | 0 | 0 | 0 | 4 | 0 | 2 | 0 | 0 | 2 | 0 |

**Table 3.1 Difference table for the first S-Box**

There are several general properties of the difference distribution table that should be mentioned. First, it should be noted that the sum of all elements in a row is $2^n$ =16; similarly the sum of any column is $2^n$ =16. Also, all element values are even (because if X', and X" are reversed, the same result is given since $\Delta X = X' \oplus X'' = X'' \oplus X'$). As well, the input difference of $\Delta X = 0$ must lead to an output difference of $\Delta Y = 0$ for the one-to-one mapping of the S-box. Hence, the top right corner of the table has a value of $2^n$ = 16 and all other values in the first row and first column are 0. Finally, if an ideal S-box can be constructed, which gives no differential information about the output given the input value, the S-box would have all elements in the table equal to 1 and the probability of occurrence of a particular value for $\Delta Y$ given a particular value of $\Delta X$ would be $1/2^n = 1/16$. However, as the properties discussed above must hold, this is clearly not achievable.

Before proceeding to discuss the combining of S-box difference pairs to derive a good differential to use in the attack, the influence of the key on the S-box differential must be discussed.

Let the input to the "un-keyed" S-box is $X$ and the output is $Y$.

However, in the cipher structure the keys applied at the input of each S-box must be considered. In this case, if let the input to the "keyed" S-box be $W = [W1\ W2\ W3\ W4]$, and consider the input difference to the keyed S-box to be

$$\begin{aligned} &\Delta W = [w_1^{'} \oplus w_1^{''} \quad w_2^{'} \oplus w_2^{''} \ldots\ldots w_n^{'} \oplus w_n^{''}] \\ &where w^{'} = [w_1^{'}\, w_2^{'} \ldots . w_n^{'}] \quad and \quad w^{''} = [w_1^{''}\, w_2^{''} \ldots . w_n^{''}] \end{aligned} \tag{3.1}$$

represent the two keyed input values. Since the key bits remain the same for w' and w",

$$\begin{aligned} \Delta w_i &= w_i^{'} \oplus w_i^{''} = (x_i^{'} \oplus k_i) \oplus (x_i^{''} \oplus k_i) \\ &= x_i^{'} \oplus x_i^{''} = \Delta x_i \qquad Since k_i \oplus k_i = 0 \end{aligned} \tag{3.2}$$

Hence, the key bits have no influence on the input difference value and can be ignored. In other words, the keyed S-box has the same difference distribution table as the un-keyed S-box.

Using the above equations, complete difference characteristics can be calculated for the SPN and hence differential attack can be mounted.

#### ***3.4.2.2 Constructing Differential Characteristics***

After navigating through the SPN, with an input difference 00F0 [0000,0000,1111,0000], difference in all bits for the input of the third S-

Box, an input difference for the last round is expected to be 2157 [0010,0001,0101,0111] with a probability of 3/1024, for details see chapter 4.

#### *3.4.2.3 Extracting Key Bits*

Once an $R-1$ round differential characteristic is discovered for a cipher of $R$ rounds with a suitably large enough probability, it is possible to attack the cipher by recovering bits from the last sub-key. In the case of our example, it is possible to extract bits from sub-key $K5$. The process followed involves partially decrypting the last round of the cipher and examining the input to the last round to determine if a right pair has probably occurred or not. The sub-key bits following the last round at the output of S-boxes in the last round influenced by non-zero differences in the differential output is referred as the *target partial sub-key*.

A partial decryption is executed for each pair of cipher-text corresponding to the pairs of plaintext used to generate the input difference $\Delta X$ for all possible target partial sub-key values. A count is kept for each value of the target partial sub-key value. The count is incremented when the difference for the input to the last round corresponds to the value expected from the differential characteristic. The partial sub-key value which has the largest count is assumed to indicate the correct value of the sub-key bits. This works because it is assumed that the correct partial sub-key value will result in the expected differential characteristics. An incorrect sub-key is assumed to result in a relatively random effect at the bits entering the S-boxes of the last round then, the difference will not be as expected from the characteristic.

Results from implementing the attack are shown below for last round Sub-Key=1E4F

| Candidate Key | Right Pairs out of 5000 |
|---|---|
| 174F | 5 |
| 190F | 5 |
| 194D | 5 |
| 194F | 8 |
| 199F | 5 |
| 1E0F | 6 |
| 1E4D | 6 |
| 1E4F | 12 |
| 1E7F | 5 |
| 1E8F | 6 |
| 1E9F | 9 |
| 1ECF | 8 |
| 1EDF | 7 |
| 1EEF | 8 |
| 1EFF | 7 |
| 6ECF | 5 |
| 7ECF | 5 |
| 7EFF | 6 |
| 8ECF | 5 |
| 8EFF | 6 |
| D944 | 5 |
| DE44 | 5 |
| EECF | 5 |
| EEFF | 7 |
| FE9F | 5 |
| FEDF | 5 |

As can be seen, the correct key gives a very close match to the differential probability 0.24% (12/5000) compared to 0.293% (3/1024) as the calculated differential probability.

#### *3.4.2.4 Complexity of the Attack*

For differential cryptanalysis, the S-boxes involved in a characteristic which have a non-zero input difference (and hence a non-zero output difference) are called *active* S-boxes. In general the larger the probabilities of differentials to occur (of the active S-boxes), the larger the differential characteristic probability for the complete cipher. Also, the fewer active S-boxes, the larger the differential characteristic probability.

The data required to mount the attack is considered a measure of the complexity of the cryptanalysis.

In general it is very complex to determine exactly the number of chosen plaintext pairs required to mount the attack. However, it can be shown that a good rule-of-thumb for the number of chosen plaintext pairs, $N_D$, required to distinguish right pairs when trying sub-key candidates is

$$N_D = c / P_D$$

Where $P_D$ is the differential characteristic probability for the R-1 rounds of the R-round cipher and c is a small positive integer.

The above equation simply indicates that a few occurrences of the right pair are enough to give a count to the correct target partial sub-key value that is significantly greater than the counts for the incorrect target partial sub-key values. Since a right pair is expected to occur for about every $1/P_D$ pairs examined, in practice, it is generally reasonable to use some small multiple of $1/P_D$ chosen plaintext pairs to successfully mount the attack.

Approaches to provide resistance to differential cryptanalysis have been focused on the S-box properties (i.e., minimizing the difference pair probability of an S-box) and finding structures to maximize the number of active S-boxes. Rijndael is a good example of a cipher designed to provide high resistance to differential cryptanalysis.

Probability $P_D$ is an estimate only. In practice, in many ciphers it has proven to be reasonably accurate [7].

## 3.5 Interpolation Cryptanalysis

As deferential cryptanalysis gained so much interest, many works tried to extend the idea of differential cryptanalysis to attack more block ciphers not vulnerable to normal deferential cryptanalysis.

In [37] Lai proposed a definition for higher order derivative of discrete functions, Later Knudsen used this to produce an attack known as higher deferential attack [32].

In [32] Jakobsen et al demonstrated the idea of interpolation attack. The system proposed can be used to attack block ciphers of low algebraic degree by approximating the R-1 rounds of the cipher with a polynomial of some degree d. Using Lagrange or Newton interpolation methods (see appendix B), the attacker can calculate the coefficients of the interpolating polynomial and uses this polynomial to decide if a key value is eligible to be a correct one. The entire attack is based on the approximation of that a wrong guessed value of the last round key will produce a randomly distributed bit values that disagree with other known plaintext- ciphertext pairs not used to produce the interpolating polynomial. In the following sections, a brief description of the basic interpolation attack as presented in [32] and [38] is given.

### 3.5.1 Description of the attack

The interpolation attack is described as follows,

The output of an R-1 round reduced cipher can be expressed as a polynomial $p(x) \in GF(2^m)[x]$ of the plaintext. Assume that this polynomial has a degree d, hence has d+1 unknown coefficients, and also assume that d+2 plaintext/ciphertext pairs are available. For each value of the last round key, one can perform a decryption assuming that value of the key and constructs the interpolation polynomial using d+1 pairs using Lagrange or Newton interpolation. Then check for the output of the polynomial with the remaining pair. If the output matches then the correct value of the last round key has been discovered with some high probability by reasoning. The attack depends on the following theorem,

*Theorem 3.1*

*Assume that the interpolating polynomial obtained from a wrongly guessed value of the last round key will produce a random output and uniformly distributed over the output bits when tried with other known plaintext – partially decrypted text pair, then there exist an interpolation attack to the cipher at hand such that the last round key value is obtained by dropping out the values of the wrong key.*

A simplified proof to the above theorem is given here,

***Proof***.

Now let y be the output of the cipher given a particular input x and $\tilde{y}$ be the partially decrypted cipher-text with the correct value of the key K and let $\tilde{y}'$ be the one obtained by a wrong guess of the key, and assume that the last round function mapping is g(.) , note that the round as hole is considered not the core function and is reversible, then

$$\tilde{y} = g^{-1}(k, y) \tag{3.3}$$

$$\tilde{y}' = g^{-1}(k', y) = g^{-1}(k', g(k, \tilde{y})) \tag{3.4}$$

Then the difference between the two values is the difference between applying the last round function twice, one with a correct key and the inverse with a wrong one, then from the nature of the cipher round mapping and its strong confusion and diffusion characteristics, this difference is expected to occur randomly and uniformly distributed over the output bits and hence one will unlikely be able to find an interpolating polynomial that finds the correct mapping for the R-1 rounds block cipher.

It was demonstrated how to use of the interpolation attack on many cipher proven to be secure against the linear or differential cryptanalysis [32].

The interpolation attack is only applicable to ciphers which can be represented with low algebraic degree polynomials, and hence its round function must not be very complicated. One of the ciphers attacked in [32] was the Knudsen-Nyberg cipher with a round function described as $F(x,k) = d(h(e(x) \oplus k))$, where $h(x) = x^3$ in $GF(2^{33})$ and e is an affine extension combination of the input bits $e : GF(2^{32}) \rightarrow GF(2^{33})$ and d is a discarding transform as $d : GF(2^{33}) \rightarrow GF(2^{32})$.

As can be seen the degree of the round function is a little bit small.

In Rijndael [6], Daemen and Rjimen used an affine transform of degree 7 to overcome the interpolation attack and they have succeeded (see chapter 2 and appendix B for details).

### 3.5.2 Complexity of the attack

The complexity of the attack is based on the number of pairs needed to perform the interpolation. For each interpolation, to be completed, n+1 pairs of plaintext- ciphertext are needed where n is the number of non-zero coefficients in the target polynomial and generally $n \leq d$ . If the last round key is b bits, then the attack complexity is $2^b(n+1)$. The attack can be extended into a chosen plaintext attack where some of the input bits may be set to zero and the polynomial is only representing a function of the other input bits.

There are much more attacks developed along the years, such as Square Attacks, Higher and Truncated differential attacks and more. Refer to [2], [38] and [9] for details

# Chapter 4: Genetic Algorithm Cryptanalysis

The cryptanalysis problem can be modeled as a search problem, in principle, through the key space. Brute force attack can be viewed as a linear search for the target key. The security of a cipher is measured based on the complexity of the type of attacks to this particular cipher system as described in the previous chapter. Many researches have been developed in the direction of the reduction of the complexity of the attacks on different cipher systems, differential cryptanalysis is an example.

Genetic algorithms (GA), see appendix A, is an attractive scope for cryptanalysis. Genetic algorithms can be modeled as a guided random search in the problem space, where upon the completion of each generation in a genetic algorithm process, better solutions to the problem at hand are found, and by the end of the process almost the best solution is found.

The main problem of using a genetic algorithm in cryptanalysis is to find a fitness measure of how close a candidate key is to the target key.

***A candidate key*** is a key value under judgment in a key search attack on a cipher.

***A target key*** is the correct key value that needs to be found as a target of an attack on a particular cipher.

The possibility of using the genetic algorithm in key search is very attractive due to the ability of genetic algorithm to reduce the complexity of the search problem.

The problem again is to find such fitness measure. Of course one of the direct measures, may be the inverse of the Euclidian distance between the candidate key and the target key, but since the target key is unknown, then this distance can not be calculated. Also due to the confusion effect of modern block ciphers, there is no simple relation of a candidate key and the number of correct bits in a ciphertext.

This work tries to find a fitness measure based on the differential catachrestic of a cipher. In other words, the differential cryptanalysis technique will be enhanced, performance wise, using genetic algorithms. The following sections introduce a frame work of how to use genetic algorithms in cryptanalysis, an overview of previous work in this field and proposal for new technique. The proposed solution is applied to both of the generic block ciphers models, the SPN and the FN.

## 4.1 Genetic Algorithm in Cryptanalysis, a Framework

The cryptanalysis problem, as mentioned before, can be modeled as the fittest key, the most close to the target key, search in the scope of genetic algorithms.

Three problems have to be addressed first before a framework can be achieved.

1. Candidate key representation as a chromosome in the scope of GA
2. Fitness Measure of how close a candidate key is to the target key
3. A stooping condition to stop the genetic algorithms evolutionary process.

In section 4.3, a proposal for the solution of these three problems is given. Figure 4.1 shows a flowchart of the cryptanalysis problem as a genetic algorithm fitness maximization problem, assuming the mentioned three problems had been addressed.

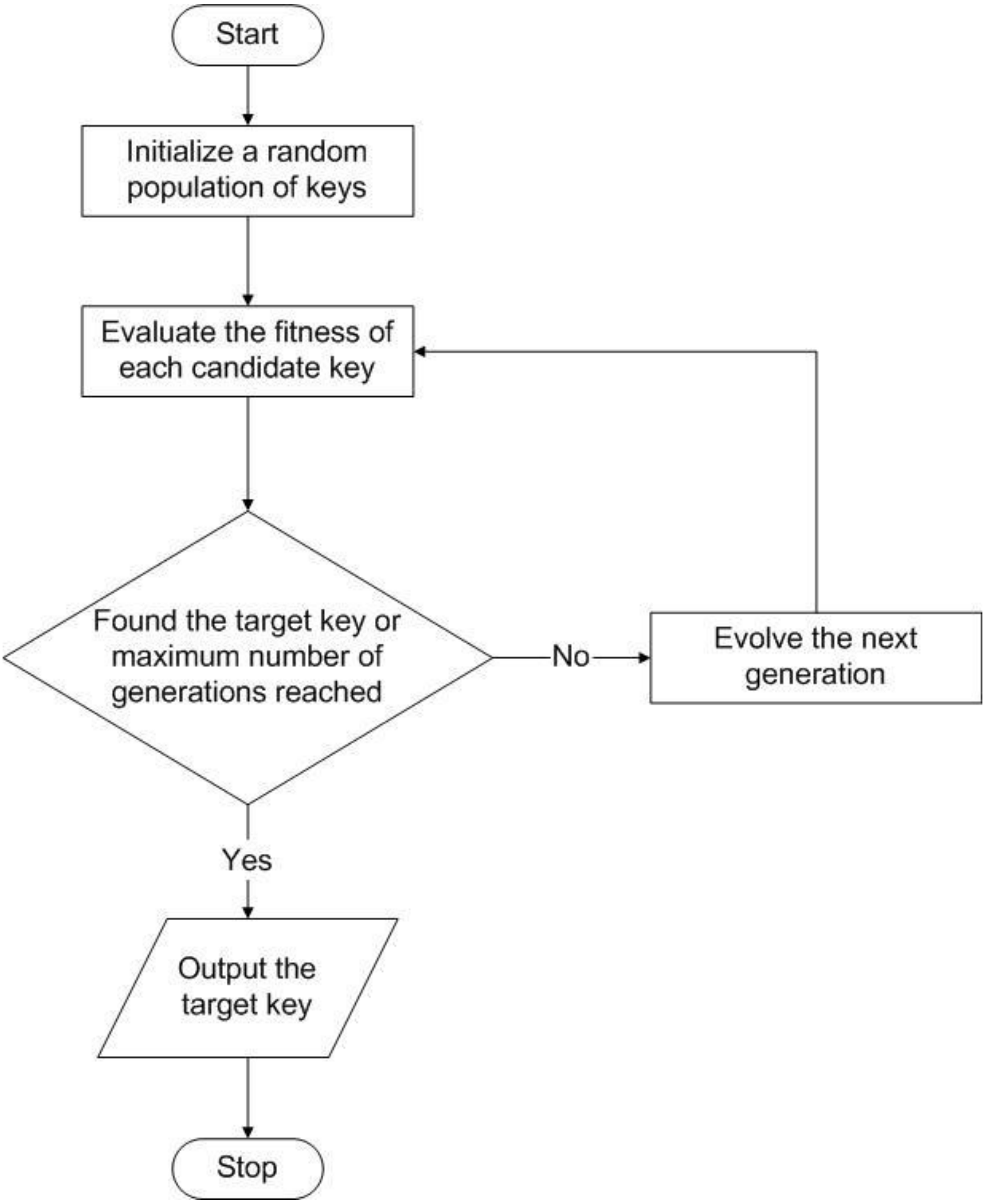

Figure 4.1: CRYPTANALYSIS MODELED AS A GENETIC ALGORITHM PROBLEM

## 4.2 Previous Attempts to Use Genetic Algorithms in Cryptanalysis

Several attempts to use genetic algorithms were made, [39], [40], and [41]. The three attacks were targeted to simple classical symmetric key ciphers (Simple Substitution Cipher, Three Rotor Machine Cipher, and Simple Transposition Cipher respectively) and their fitness measure was based on the statistical analysis of the ciphertext knowing its target language. In all of these simple cases, a population of candidate keys were generated. A fitness measure based on the statistical properties (frequency of letters, diagrams, and trigrams as defined in chapter 2) compared to the target language, English in all three cases, was calculated. Genetic algorithm was then applied for a fixed number of generations and hopefully at the end the target key is found.

As all modern ciphers hide the statistical information of plaintext and they do not propagate to the ciphertext and hence all of the previous mentioned works have no real application actually, then a new fitness measure is required.

## 4.3 Genetic Algorithm Cryptanalysis of the Basic SPN

One of the most powerful cryptanalysis techniques developed was the differential cryptanalysis described earlier. As we concluded, the complexity of the attack was reduced from $2^N$ to $R*2^n$ where N is the entire key length while n is the sub-key length and R is the number of rounds of the cipher. In the example of the SPN in chapter 3, N=80 and n=16, and R=5. Still it is highly complex, since for each candidate sub-key, a huge number of encryption and decryption operations must be carried out. In this work a trial to reduce the complexity of the differential attack is given. Genetic algorithm is used to generate new keys directed to the correct target key by using the match of the candidate sub-key differential characteristics and that of the system as the fitness measure. The technique described after is found to reduce the complexity of the attack by at least 25% due to several experiments on the success of the attack.

### 4.3.1 Problem formulation

The use of such attractive technique, namely the genetic algorithms, has not been studied due to the difficult formulation of the cryptanalysis problem into a genetic algorithm optimization problem.

As mentioned previously, three problems arise in the use of the genetic algorithms in cryptanalysis.

1. Key representation.
2. Fitness Measure.
3. Stopping criteria of the Genetic algorithm search technique.

### 4.3.2 Key Representation

The problem of key representation is inherently solved by using binary genetic algorithm, as the string used to represent a chromosome is a binary stream of bits, and the key is also a binary word. The crossover and mutation processes are applied directly on the candidate keys to generate new keys directed to the correct key via the fitness measure.

### 4.3.3 Fitness Measure

The fitness function chosen is the main factor of the genetic algorithm. The differential characteristic of the cipher is used as a fitness measure, that is a key that gives good differentials, has a better fitness than the others. This characteristic has the advantage of being language independent and no statistical information of the language is anymore required.

So the fitness function can be represented by a number F

$$F = \frac{d}{M} \tag{4.1}$$

Where d is a count that is kept for each value of the target partial sub-key value and M is the total number of plaintext pairs used to evaluate the keys with a fixed difference. The count will be incremented when the difference for the input to the last round corresponds to the value expected from the differential characteristic. This works because it is assumed that the correct partial sub-key value will result in the difference to the last round being as frequent as expected from the characteristic since the characteristic has a high probability of occurring. An incorrect sub-key is assumed to result in a relatively random guess at the bits entering the S-boxes of the last round.

### 4.3.4 Constructing the differential characteristics of the SPN

As described, the differential characteristics of the SPN is used in the measure of fitness in a genetic algorithm cryptanalysis as proposed, and hence it must be constructed first.

Once the differential information has been compiled for the S-boxes in an SPN, useful differential characteristics will be determined for the overall system. This can be done by combining appropriate difference pairs of S-boxes.

Consider a differential characteristic involving $S_{13}$, $S_{22}$, $S_{32}$, $S_{33}$ and $S_{34}$ (visualized in Figure 4.2). The diagram illustrates the influence of non-zero differences in bits as they travel through the network, highlighting the S-boxes that may be considered as active. Note that this develops a differential characteristic for the first 3 rounds of the cipher and not the full 4 rounds.

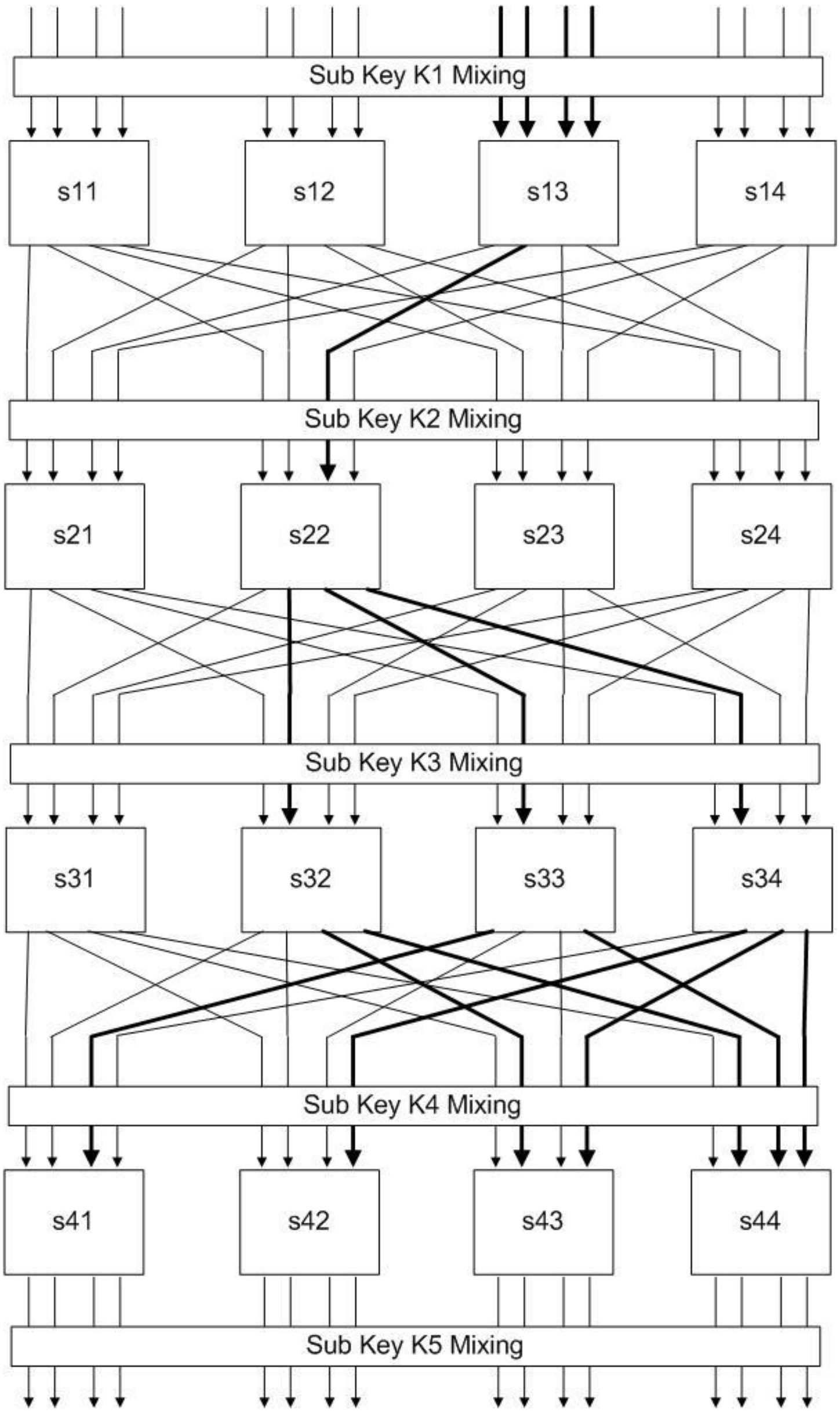

Figure 4.2: DIFFERENTIAL CHARACTERISTIC OF THE SPN

From the diagram, it is apparent that the active S-Boxes are as follows (with difference pairs and probability of occurrence shown)

| Active S-Box | Input Difference | Output Difference | Probability |
|---|---|---|---|
| S13 | F | 4 | 8/16 |
| S22 | 2 | 7 | 4/16 |
| S32 | 4 | 3 | 6/16 |
| S33 | B | 1 | 4/16 |
| S34 | 4 | 7 | 4/16 |

All other S-boxes will have zero input difference and consequently zero output difference. The input difference to the cipher is equivalent to the input difference to the first round and is given by

$\Delta X$= [0000 0000 1111 0000]

Let $\Delta U_i$ *represent the* input difference to the i$^{th}$ round then

$$\Delta X = 00F0$$
$$\Delta U_1 = 00F0$$
$$\Delta U_2 = 0200$$
$$\Delta U_3 = 0444$$
$$\Delta U_4 = 2157$$

In determining the probability, given plaintext difference $\Delta X$, it is assumed that the differential of each round is independent of the differential of other rounds, hence, the probability of occurring is determined by the product of the probabilities.

A probability of 8/16 × 4/16 × (6/16*4/16*4/6) = 3/1024 given plaintext difference $\Delta X$, where again it has been assumed the independence between the difference pairs of S-boxes in the same round. During the cryptanalysis process, many pairs of plaintexts for which $\Delta X$ = [0000 0000 1111 0000] will be encrypted. With high probability, 3/1024, the differential characteristic illustrated will occur. Such pairs for $\Delta X$ are called *right pairs*. Plaintext difference pairs for which the characteristic does not occur are referred to as *wrong pairs*.

### 4.3.5 Stopping Criteria

This is one of the classical problems in genetic algorithms, that is when to stop the process of GA. The problem at hand really needs a stopping criterion that can be used.

The one chosen here, a classical choice, is when the algorithm can generate no more better solutions for a number of generations.

Let $P_D$ be the differential probability of the given SPN then fitness should be $\geq \alpha P_D$ where $\alpha$ is a positive number less than 1, taken 0.8 in this example. The maximum number of generations has been chosen in order that the total number of keys in all generations (S) not to exceed half of all possible sub-key value that is

$$S < \frac{1}{2} 2^n = 2^{n-1} .$$

### 4.3.6 Implementing the Attack

The above mentioned points were taken into consideration and an implementation using object oriented C++ is developed. The idea behind using a native implementation instead of using a ready package for genetic algorithm is the search for optimized implementation of the attack and to get stick with bit-level implementation whenever it is possible. The basic SPN model described earlier is implemented also using object oriented modeling. Each S-Box is an object from a class called CSBox. Each object differs from the other in the lookup array used to get the output of the S-Box.

Then every 4 objects are combined together in a quad-S-Box to represent the substitution of a round. Permutation is implemented as a bit shift operation for each bit of the input word. The next array shows how many positions should a bit be shifted right, negative values means a left shift.

byte shifter[16]={0,3,6,9,-3,0,3,6,-6,-3,0,3,-9,-6,-3,0};

The key is mixed using XOR operation and the SPN is implemented and tested to prove a very high throughput.

The attack is implemented by generating independent 5 sub-keys 16-bit each to represent the target key. Then a difference for 5000 pair of plaintext of 00F0 is selected. The most probable difference is then calculated through the SPN, as in section 4.3.4.

The first generation is generated randomly using a simple uniform random number generator, to cover the bit space of the last round sub-key.

Then for each chromosome, a candidate sub-key, the fitness is set to zero, then the entire test pairs are used and the output difference is compared to the expected differential. If the correct difference is found the fitness value is incremented and finally normalized to the number of pairs.

The genetic algorithm then goes in the normal way to generate new generations. The roulette wheel is used as a selection method. The algorithm is stopped based on the criteria described earlier.

The algorithm has been implemented to get key bits of the last round; essentially the attack shall continue upward to get the entire key bits. See figure 4.3 for a flowchart of the proposed cryptanalysis technique. Note the similarity to figure 4.1 and the extra details given here in the fitness measure and stopping conditions.

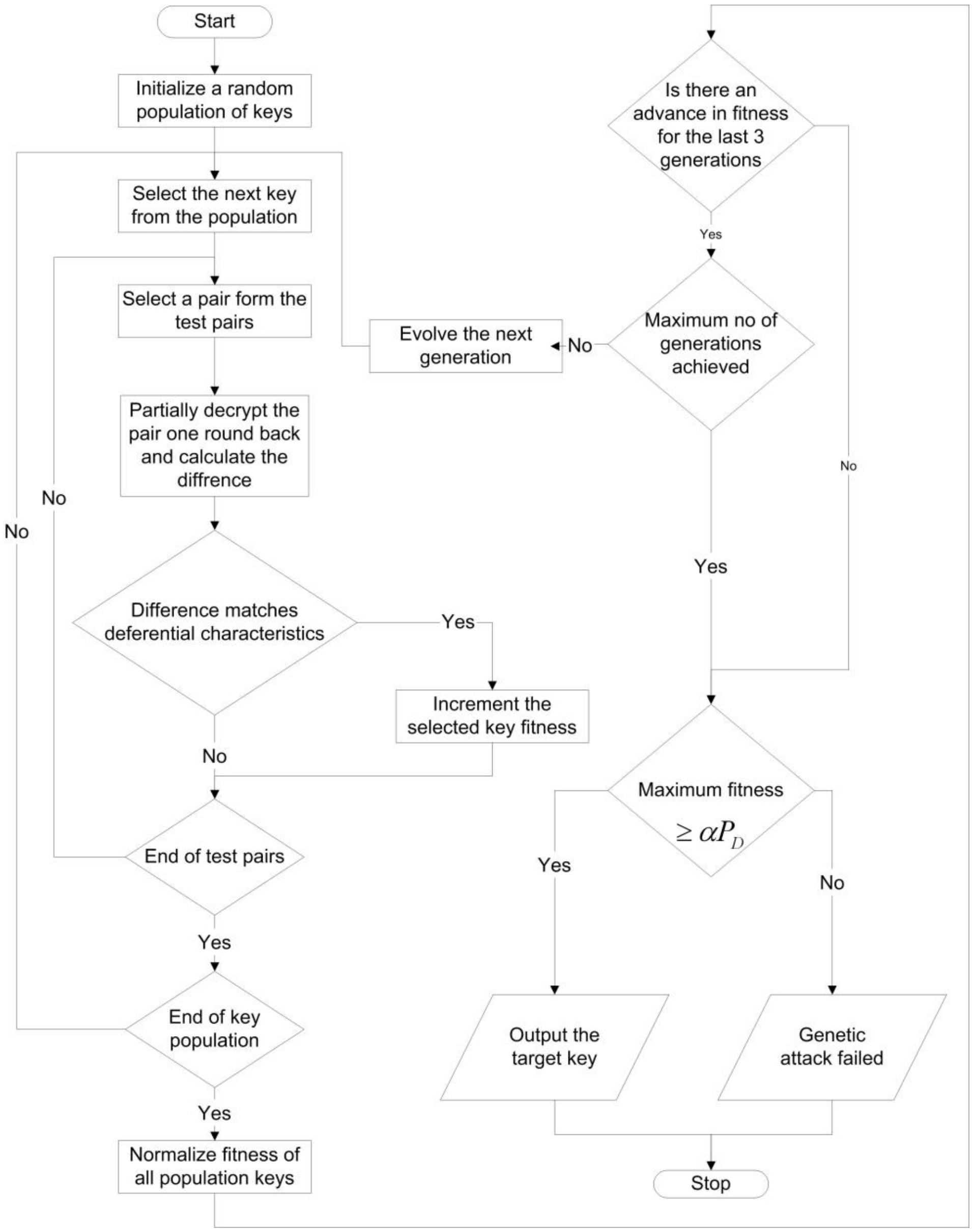

Figure 4.3: GENETIC ALGORITHM CRYPTANALYSIS OF THE BASIC SPN

### 4.3.7 Experimental Results

The attack had been tested several times on different keys.

The following table shows a sample of the attack to get the bits of the last round sub-key, in this example it was 8F6E, corresponding to difference of 00F0 in the input. The differential probability of the SPN with input difference 00F0 is 3/1024. In our example, we have used 5000 pair plaintext means that a correct sub-key should produce about 15 of correct output differences.

The genetic algorithm parameters were

| | |
|---|---|
| Chromosome Size: | 16 bits |
| Population Size: | 1024 chromosomes |
| Maximum Number of generations: | 16 generations |
| Crossover probability: | 25% |
| Mutation probability: | 1% |

The table below shows the generations together with the best solution and its fitness. The test here went through the entire 16 generations while due to our stopping criteria it should have been stopped after generation 4.

Key: 8F6E

| Generation | Best Solution | Fitness Normalized out of 5000 | Percentage of differential probability |
|---|---|---|---|
| 0 | 8F6C | 0.0012 | 40.96% |
| 1 | 8F6C | 0.0012 | 40.96% |
| 2 | 8F6E | 0.0024 | 81.92% |
| 3 | 8F6E | 0.0024 | 81.92% |
| 4 | 8F6E | 0.0024 | 81.92% |
| 5 | 8F6E | 0.0024 | 81.92% |
| 6 | 8F6E | 0.0024 | 81.92% |
| 7 | 8F6E | 0.0024 | 81.92% |
| 8 | 8F6E | 0.0024 | 81.92% |
| 9 | 8F6E | 0.0024 | 81.92% |
| 10 | 8F6E | 0.0024 | 81.92% |
| 11 | 8F6E | 0.0024 | 81.92% |
| 12 | 8F6E | 0.0024 | 81.92% |
| 13 | 8F6E | 0.0024 | 81.92% |
| 14 | 8F6E | 0.0024 | 81.92% |
| 15 | 8F6E | 0.0024 | 81.92% |

Other tests looked as follows

Key: 2EA3

| Generation | Best Solution | Fitness Normalized out of 5000 | Percentage of differential probability |
|---|---|---|---|
| 0 | 2E23 | 0.0018 | 61.44% |
| 1 | 2E23 | 0.0018 | 61.44% |
| 2 | 2EA3 | 0.003 | 102.40% |
| 3 | 2EA3 | 0.003 | 102.40% |
| 4 | 2EA3 | 0.003 | 102.40% |

Key: 48BD

| Generation | Best Solution | Fitness Normalized out of 3000 | Percentage of differential probability |
|---|---|---|---|
| 0 | 78CA | 0.001 | 34.13% |
| 1 | 46BD | 0.002 | 68.27% |
| 2 | 46BD | 0.002 | 68.27% |
| 3 | 46BD | 0.002 | 68.27% |
| 4 | 46BD | 0.002 | 68.27% |
| 5 | 46BD | 0.002 | 68.27% |
| 6 | 46BD | 0.002 | 68.27% |
| 7 | 48BD | 0.002667 | 91.02% |
| 8 | 48BD | 0.002667 | 91.02% |
| 9 | 48BD | 0.002667 | 91.02% |

Huge number of experiments had been carried out and in all experiments, the attack succeeded.

It should be noticed from the tables given before that the differential probability is exceeded in some fitness calculations. This comes from the way the fitness is calculated by counting the right pairs and this count may not always be aligned, some extra pairs may have the fitness back under 100%.

### 4.3.8 Complexity of the Attack

As mentioned earlier, the attack is designed to reduce the complexity of the differential attack on SPN described in chapter 3.

The reduction comes from, not to try all the sub keys but only part of them in each generation, 1024, in our case.

The number of chosen plaintext pairs is a major parameter. As mentioned in chapter 3, this count is given by

$$N_D = c / P_D$$

Where $P_D$ is the differential characteristic probability of the SPN, 3/1024 in this example, and c is a positive integer.

Many values had been tested for the $N_D$ value starting from 1024, adding 1024 each time until $N_D$ reached 8192 or equivalent for c=3,6,9,12,15,18.21 and 24. Figure 4.2 shows the genetic evolving fitness for different $N_D$ normalized to the number of pairs.

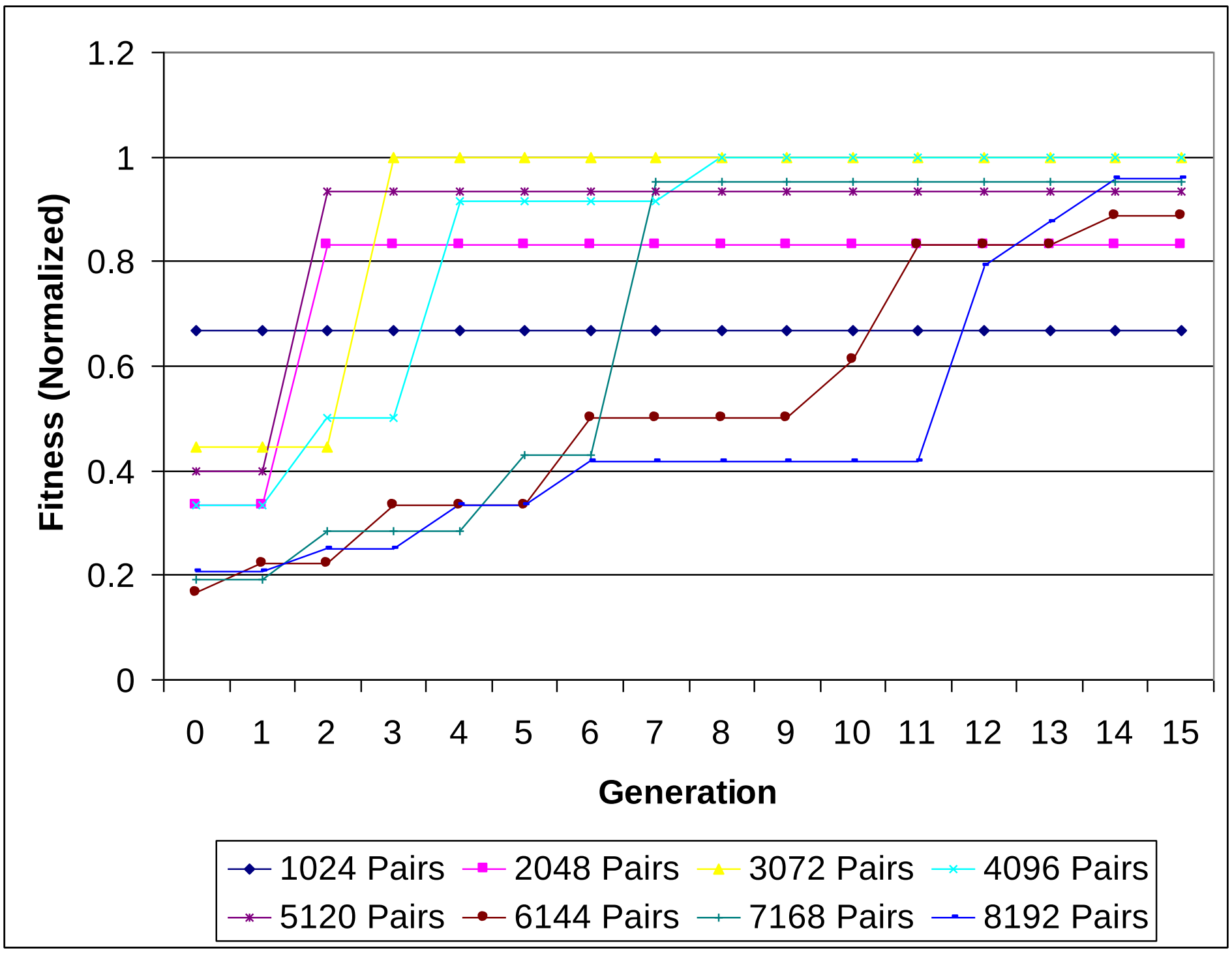

Figure 4.4: FITNESS VALUE AGAINST GENERATION FOR DIFFERENT $N_D$

From the figure, it can be seen that similar behaviour is obtained for different number of pairs, when using 1024 pairs only (c=3); the system does not converge to the correct key value. All experiments were allowed to run to the maximum number of generations.

## 4.4 Genetic Algorithm Cryptanalysis of a Feistel Ciphers

In this section a proposal to reduce the complexity of the differential attack on Feistel ciphers using the genetic algorithms is proposed. GA is used to generate new keys directed to the correct key by the matching the candidate sub-key differential characteristics and that of the system as the fitness measure. Experiments showed that the technique described reduces the complexity of the attack by at least 25%. The Feistel structure is more complicated than the SPN structure in the previous section and hence it is worth to give a detailed description.

### 4.4.1 The Considered Feistel Cipher

A 5 round Feistel cipher similar to figure 2.6 is considered. This structure has 4 sub keys. The round function F of this cipher is chosen to be identical to the round function of the basic Substitution Permutation Network (SPN) described in chapter 3. One round of this cipher is shown in figure 4.5

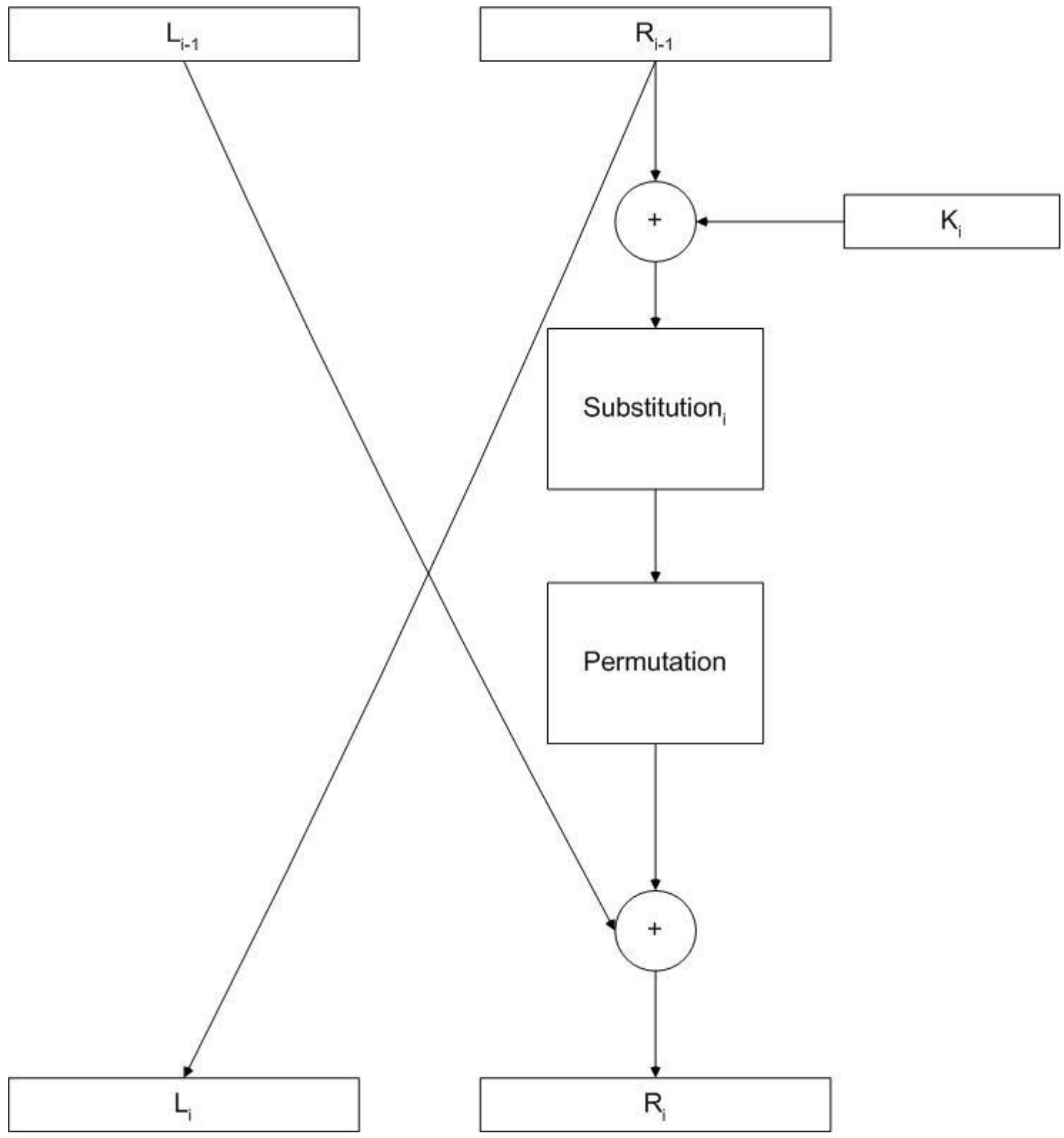

Figure 4.5: ONE ROUND OF THE TEST FEISTEL CIPHER

From the figure, the key is mixed with the input using bitwise XOR function. A substitution, n to n mapping, is performed. This substitution is changed from  round to the other.

Finally a bit permutation is performed on the block. In this case n=16 bits is chosen. For substitution, 4 S boxes from the DES block cipher are selected for each round.

The key is also 16 bits and 4 independent sub keys are used which resuls in a 64 bits length key. The overall block size is 32 bits. This structure is more complicated than the basic SPN described earlier. This structure is called DES like block cipher. Figure 4.6 shows the details of one round function.

In this cipher, the 16-bit data block is broken into four sub-blocks 4-bits each. Each sub-block forms an input to a 4x4 S-box (a substitution with 4 input and 4 output bits), which can be easily implemented with a lookup table of sixteen 4-bit values, indexed by the integer represented by the 4 input bits (see table 2.1). The permutation portion of a round is simply the transposition of the bits or the permutation of the bit positions. Figure 4.7 shows the entire cipher.

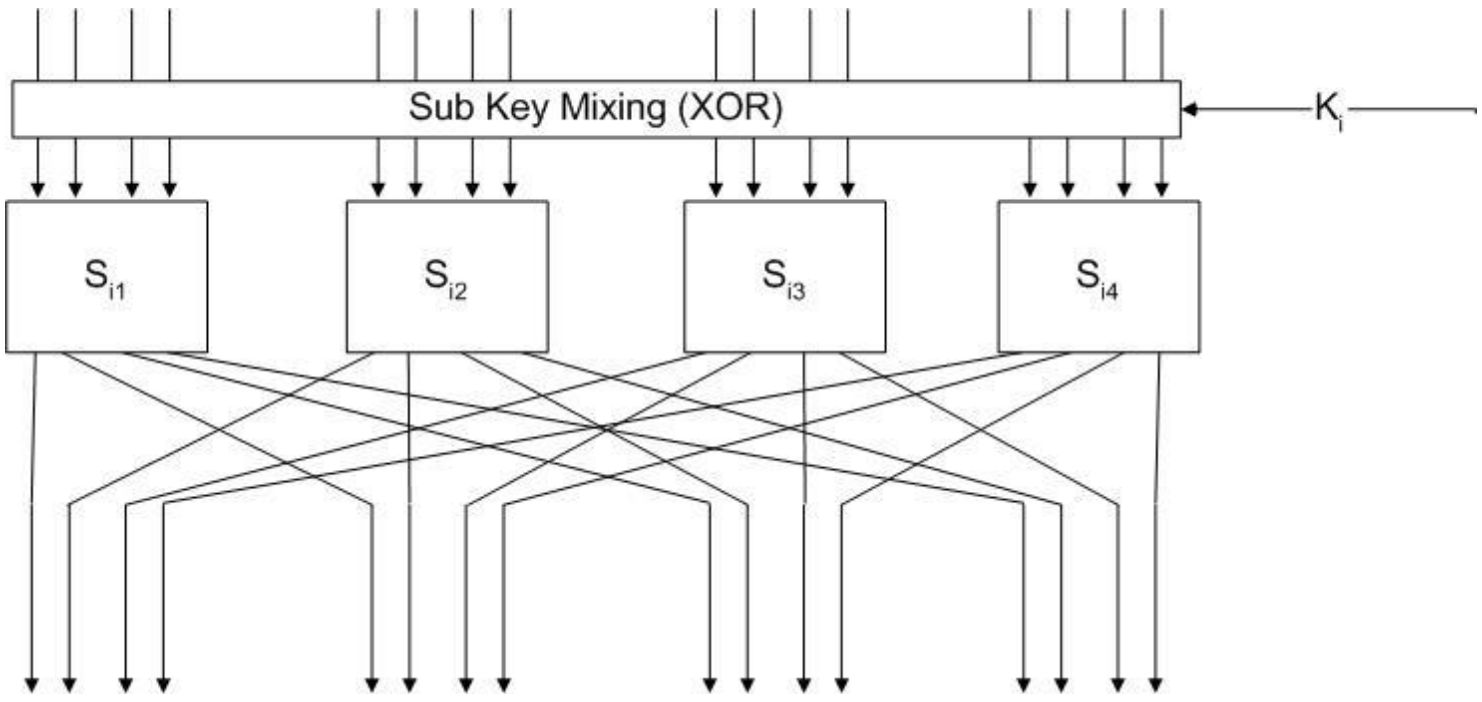

Figure 4.6: CORE FUNCTION F

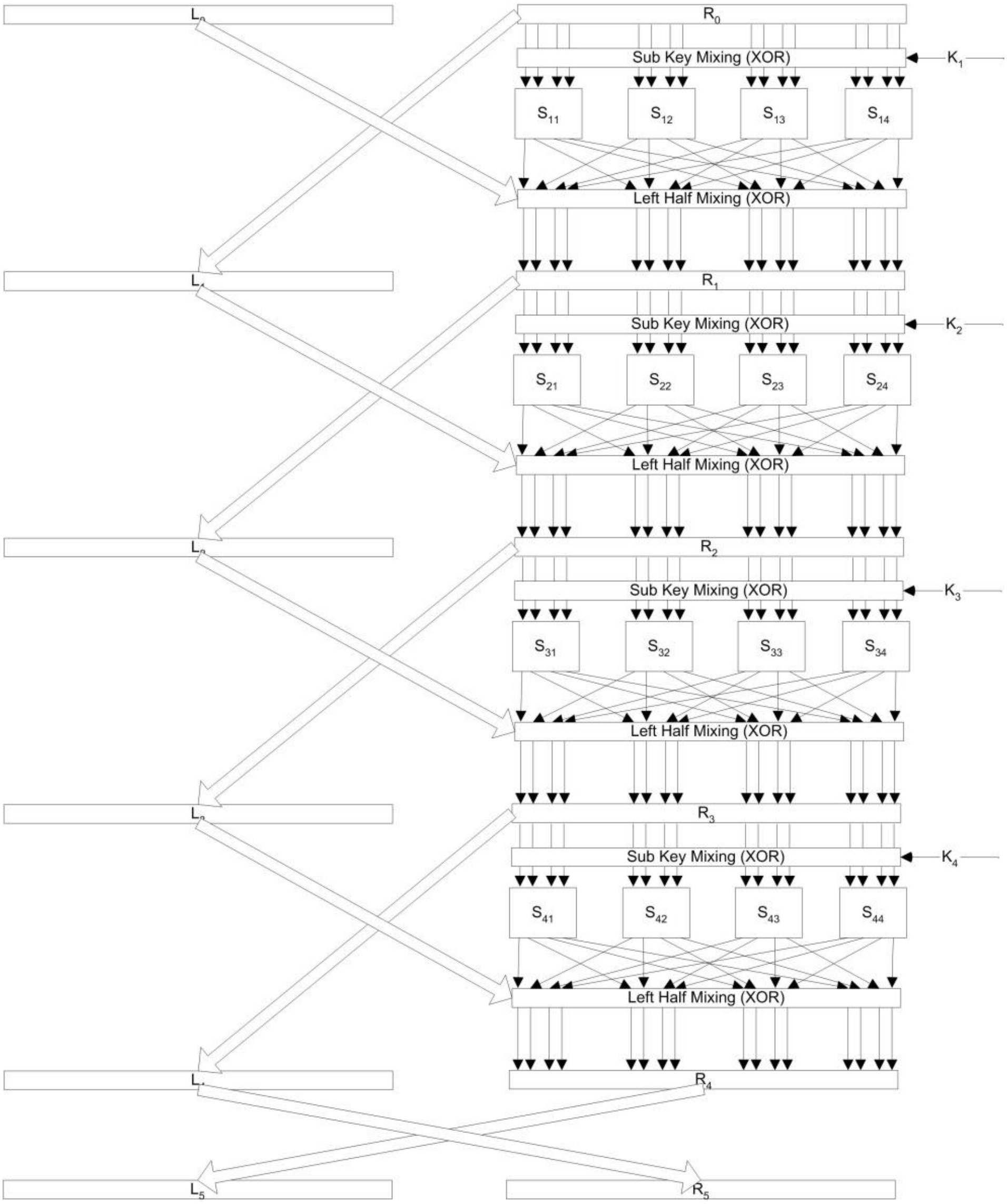

Figure 4.7: FEISTEL CIPHER UNDER ATTACK

### 4.4.2 Differential Cryptanalysis of the Feistel Cipher

The analysis of the basic SPN differential characteristics as described in the chapter 3 can be applied in the same way to the Feistel cipher structure, but the analysis is little bit complicated.

Consider Figure 4.8 of the $i^{th}$ round in Feistel cipher where the key is first mixed with the input to the core function F using bitwise XOR.

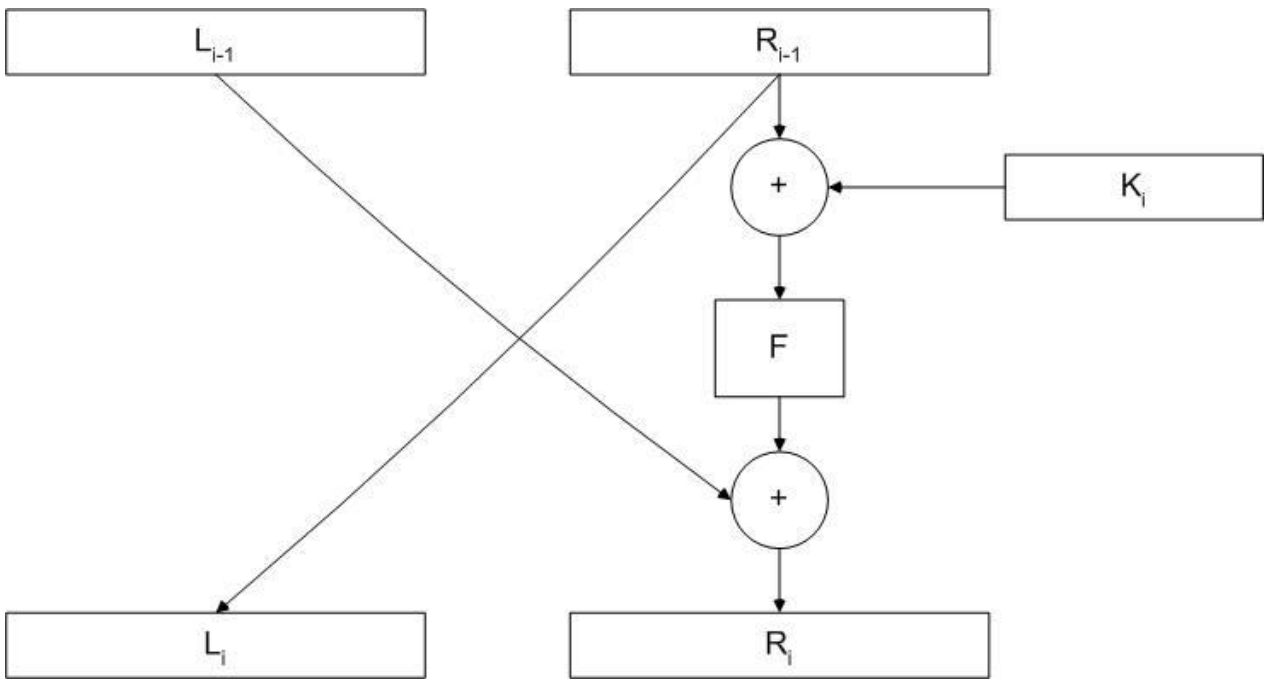

Figure 4.8: FEISTEL CIPHER ROUND WITH KEY MIXED BY XOR

Figure 4.8 is a special case of figure 2.7 where the key is applied to the input and then a non-linear function F is applied to the keyed input. This implementation yields an output dependant on both the input and the key. The implementation of figure 4.8 is the most widely used way in many modern block ciphers, see chapter 2 for DES.

Now the description of how to deduce the differential characteristics of the Feistel cipher is presented.

Consider a Feistel cipher with a round similar to the one given above in figure 4.8, let two inputs X', $L_{i-1}'$ $R_{i-1}'$, and X", $L_{i-1}''$ $R_{i-1}''$, be presented to the inputs of the round using the same key $K_i$.

Then the difference in the input to the round can be calculated by xoring the two inputs X' and X" as described earlier.

$$\begin{aligned} \Delta X^i &= X' \oplus X'' = L_{i-1}' | R_{i-1}' \oplus L_{i-1}'' | R_{i-1}'' \\ &= L_{i-1}' \oplus L_{i-1}'' | R_{i-a}' \oplus R_{i-a}'' = \Delta X_L{}^i \,|\, \Delta X_R{}^i \end{aligned} \tag{4.2}$$

Now turn to the output

$$\Delta Y^i = Y' \oplus Y'' = L_i' | R_i' \oplus L_i'' | R_i'' = \Delta Y_L{}^i \,|\, \Delta Y_R{}^i \tag{4.3}$$

It can be seen that $\Delta Y_L^i = \Delta X_R^i$ directly from the Feistel structure.
For the right half, then

$$\begin{aligned}\Delta Y_R^i &= (F(R_{i-1}'\oplus K_i) \oplus L_{i-1}') \oplus (F(R_{i-1}''\oplus K_i) \oplus L_{i-1}'') \\ &= (F(R_{i-1}'\oplus K_i)) \oplus (F(R_{i-1}''\oplus K_i)) \oplus L_{i-1}'\oplus L_{i-1}'' \\ &= (F(R_{i-1}'\oplus K_i)) \oplus (F(R_{i-1}''\oplus K_i)) \oplus \Delta X_L^{\,i}\end{aligned} \tag{4.4}$$

Now the output difference of the round depends on the output difference of the core function.
Consider a similar analysis to the one given in section 4.3, it can be seen that for the core function with input difference $\Delta X_F^{\,i} = \Delta X_R^{\,i}$ as the key bits vanishes in the XOR operation, the output difference is $\Delta Y_F$ can be obtained with a high probability if the components of the core function F are differentially analyzed.
Let $\Delta Y_F^{\,i}(\Delta X_R^{\,i})$ be the output difference of the core function F given the right half input difference $\Delta X_R^{\,i}$ then the i$^{th}$ round output difference as a function of the round input difference and the core function output difference can be written as

$$\Delta Y^i = \Delta Y_L^{\,i} \mid \Delta Y_R^{\,i} = \Delta X_R^{\,i} \mid \Delta Y_F^{\,i}(\Delta X_R^{\,i}) \oplus \Delta X_L^{\,i} \tag{4.5}$$

The above equation can then be used to construct complete differential characteristics of the Feistel cipher knowing its core function differential characteristics in similar way as in the case of the SPN.

### 4.4.3 Constructing the differential characteristics of the Feistel Cipher

As in SPN, a chosen difference is applied and navigated through the structure,
Let the difference be 000000f0, i.e. the difference in the least significant byte of the right half (all numbers in hexadecimal).

Then the only active S-Box is $S_{13}$ with input difference f and a probable output difference of 4.

For that difference an expected output from the first round is 0040 which becomes 0020 after permutation.

After mixing with the left half (using bit wise XOR), the output difference for the output right half is 0020, as the left half difference is 0000.

Using the same notation used in the SPN, then the differences as propagated are as follows,

$$\Delta X = 000000F0$$
$$\Delta U_1 = 000000F0$$
$$\Delta U_2 = 00F00200$$
$$\Delta U_3 = 020004B4$$
$$\Delta U_4 = 04B40357$$

Active S-Boxes are listed below

| Active S-Box | Input Difference | Output Difference | Probability |
|---|---|---|---|
| S13 | F | 4 | 8/16 |
| S22 | 2 | 7 | 4/16 |
| S32 | 4 | 3 | 6/16 |
| S33 | B | 1 | 4/16 |
| S34 | 4 | 7 | 4/16 |

Then the expected input difference to the last round is 04B40357 with a probability (as before multiplying all the probability assuming independence) is 8/16*4/16*6/16*4/16*4/16=3/1024. The construction is visualized in figure 4.9.

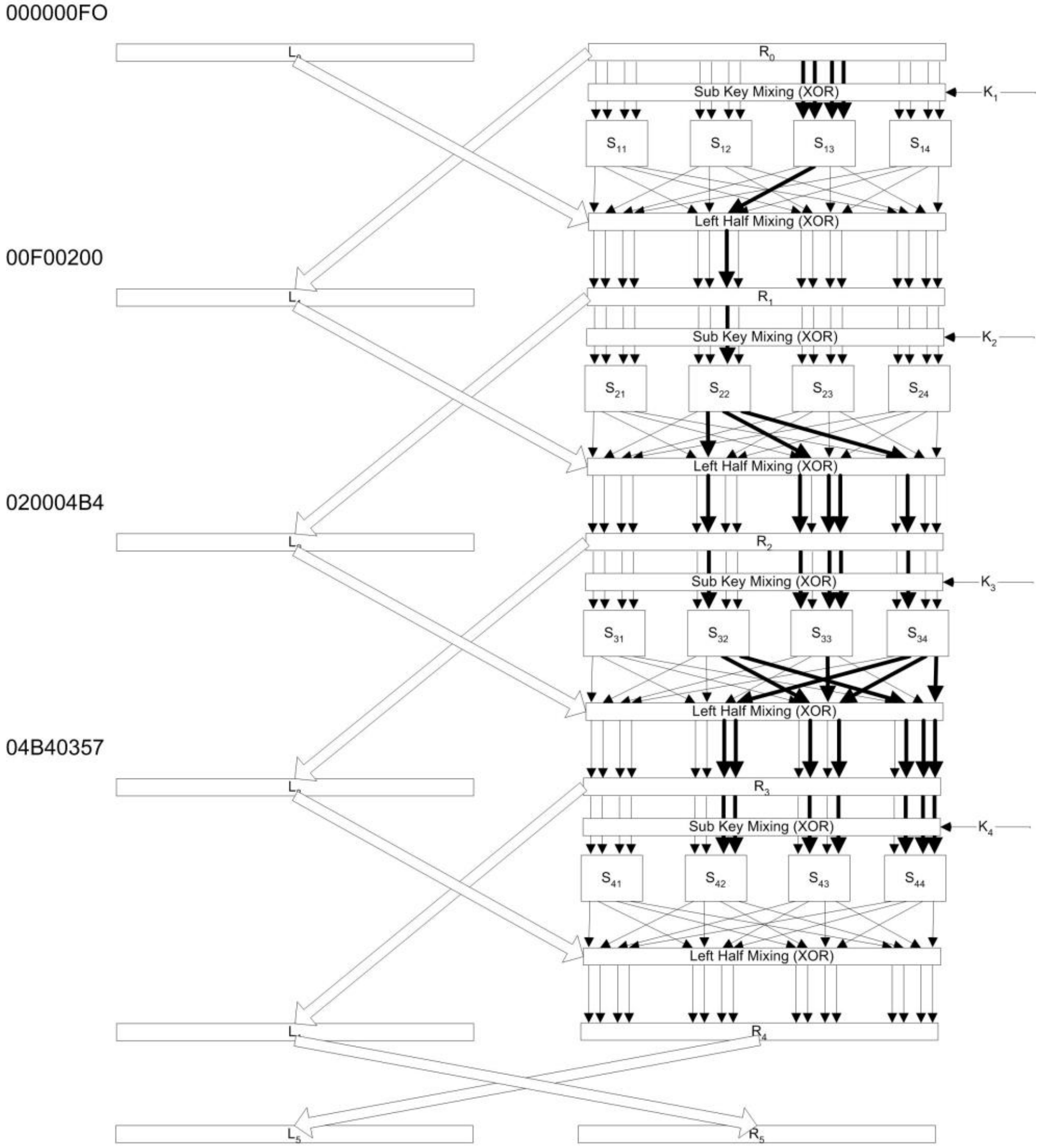

Figure 4.9: DIFFERENTIAL CHARACTERISTIC OF THE TEST CIPHER

### 4.4.4 Problem Formulation in Genetic Form

As in the case of the SPN attack using genetic algorithm, three problems arise

1. Key representation.
2. Fitness Measure.
3. Stopping criteria of the Genetic algorithm search technique.

In the previous section solutions to these three problems have been proposed, and these proposals are used here.

### 4.4.5 Implementing the Attack

The attack is implemented by generating independent 4 sub-keys 16-bit each to represent the target key. Then a difference is selected and 5120 (5*1024) pairs of plaintext with that difference are generated, in our case it is 000000F0, the most probable difference to the last round is then calculated through the cipher as described previously and is expected to be 04B40357.

Only the right half of the difference is considered during the attack as the candidate sub-key only affects that half, while the left half is independent on the key and so it is not considered.

The first generation is generated randomly using a simple uniform random number generator, to cover the bit space of the last round sub-key. Then for each chromosome, a candidate sub-key, at start the fitness is set to zero, then the entire test pairs are used and the output difference is compared to the expected differential. If the correct difference is found the fitness value is incremented and finally normalized to the number of pairs.

The genetic algorithm then goes in the normal way to generate new generations. The roulette wheel is used as a selection method. The algorithm is stopped based on the criteria described earlier. That is when the algorithm can generate no more better solutions for a number of generations.

Let $P_D$ be the differential probability of the given SPN then fitness should be $\geq \alpha P_D$ where $\alpha$ is a positive number less than 1, taken 0.8 in this example. The maximum number of generations has been chosen in order that the total number of keys in all generations (S) not to exceed half of all possible sub-key value that is

$$S < \frac{1}{2} 2^n = 2^{n-1}$$

The algorithm has been implemented to get key bits of the last round; essentially the attack shall continue upward to get the entire key bits.

### 4.4.6 Experimental Results

The attack had been implemented several times,

The genetic algorithm parameters were

| | |
|---|---|
| Chromosome Size: | 16 bits |
| Population Size: | 1024 |
| Maximum Number of generations: | 16 |
| Crossover probability: | 25% |
| Mutation probability: | 1% |

The table below shows the generations together with the best solution and its fitness. (Full 16 generation results are shown instead of stopping at generation 8)

Key: 8BC4

| Generation | Best Solution | Fitness normalized out of 5120 | Percentage of differential probability |
|---|---|---|---|
| 0 | 9BC5 | 0.000781 | 26.66% |
| 1 | 8B75 | 0.000977 | 33.35% |
| 2 | 8BC5 | 0.001758 | 60.01% |
| 3 | 8BC5 | 0.001758 | 60.01% |
| 4 | 8BC5 | 0.001758 | 60.01% |
| 5 | 8BC5 | 0.001758 | 60.01% |
| 6 | 8BC4 | 0.003125 | 106.67% |
| 7 | 8BC4 | 0.003125 | 106.67% |
| 8 | 8BC4 | 0.003125 | 106.67% |
| 9 | 8BC4 | 0.003125 | 106.67% |
| 10 | 8BC4 | 0.003125 | 106.67% |
| 11 | 8BC4 | 0.003125 | 106.67% |
| 12 | 8BC4 | 0.003125 | 106.67% |
| 13 | 8BC4 | 0.003125 | 106.67% |
| 14 | 8BC4 | 0.003125 | 106.67% |
| 15 | 8BC4 | 0.003125 | 106.67% |

Other tests looked as follows

73F7

| Generation | Best Solution | Fitness normalized out of 5120 | Percentage of differential probability |
|---|---|---|---|
| 0 | 83F7 | 0.001367 | 46.66% |
| 1 | C3F7 | 0.001563 | 53.35% |
| 2 | E3F7 | 0.002148 | 73.32% |
| 3 | 23F7 | 0.002148 | 73.32% |
| 4 | 33F7 | 0.002148 | 73.32% |
| 5 | 73F7 | 0.003125 | 106.67% |
| 6 | 73F7 | 0.003125 | 106.67% |
| 7 | 73F7 | 0.003125 | 106.67% |

B3CC

| Generation | Best Solution | Fitness normalized out of 5120 | Percentage of differential probability |
|---|---|---|---|
| 0 | A34C | 0.001758 | 60.01% |
| 1 | A34C | 0.001758 | 60.01% |
| 2 | A3CC | 0.002734 | 93.32% |
| 3 | A3CC | 0.002734 | 93.32% |
| 4 | B3CC | 0.003125 | 106.67% |
| 5 | B3CC | 0.003125 | 106.67% |
| 6 | B3CC | 0.003125 | 106.67% |

11D2

| Generation | Best Solution | Fitness normalized out of 5120 | Percentage of differential probability |
|---|---|---|---|
| 0 | 16DE | 0.000781 | 26.66% |
| 1 | 11D0 | 0.001172 | 40.00% |
| 2 | 11D2 | 0.002734 | 93.32% |
| 3 | 11D2 | 0.002734 | 93.32% |
| 4 | 11D2 | 0.002734 | 93.32% |

The results obtained earlier are similar to that of the SPN.

It is worth some comments on the result tables given before. Again the fitness may exceed the differential probability due to the way of calculating the fitness by counting.

In the second table (of 73F7 key) the fitness was constant for generations 2, 3, and 4 but the process did not stop because the fitness didn't exceed the 80% of the differential probability. The GA is stopped after the $7^{th}$ generation instead. Also in the third table (of B3CC key), The fitness exceeded the 80% percentage (93.32%) at the second generation but only for two generations, 2 and 3, so the GA didn't stop until a fixed fitness for 3 generations (4, 5, and 6) is observed.

It is concluded from both results that the proposed genetic algorithm attack can be applied on any cipher vulnerable to differential attack based on that its differential characteristics is discovered. The proposed attack outperforms the differential attack as the number of keys being tried is much less than that of the differential attack. The reduction is guaranteed to be at least by 25% if the attack succeeds (16 generations 1024 sub-key each results a total of 16384 sub keys tried out of all sub-keys of $2^{16}$=65536)

# Chapter 5: Neural Network Cryptanalysis

Neural networks, see appendix A, have proven to be very efficient in many problems. One of these problems is function approximation. Unlike interpolation, neural networks can approximate large degree algebraic functions through simple iterative process called learning, while in interpolation, a huge number of equations have to be solved or evaluated. This criterion of neural networks inspired its use in cryptanalysis as a natural extension to interpolation attack. In this chapter a hypothetical simplified cipher using a Feistel structure and using the Rijndael S-Box, known to be secured against interpolation attack, as its core function is used to address the neural network cryptanalysis.

A short comparison between the proposed attack and interpolation attack is given.

## 5.1 Universal Approximation

Neural Networks can be used as universal mapping approximator. A fully connected feed forward network as shown in figure 5.1, trained with the back propagation algorithm can be viewed as a practical nonlinear input-output mapping. This is inherited from a theory called Universal Approximation Theory, which can be viewed as a natural extension to the Weirstrass theorem developed in 1885, that states that any continuous function over a closed interval on the real axis can be expressed in that interval as an absolutely and uniformly convergent series of polynomials. Fourier series representation is a special type of that universal approximation. The Universal approximation theory is described as follows,

Let $\phi(.)$ be a non constant bounded and a monotone-increasing continuous function. Then given any function f(x) where x is a vector in m dimensional hyper space, then there exists set of real numbers $a_i$, $b_i$ and $w_{ij}$ where i=1,2,….,n and j=1,2,….m such that

$$F(x) = \sum_{i=1}^{n} a_i \phi\left(\sum_{j=1}^{m} w_{ij_j} x + b_i\right) \tag{5.1}$$

as an approximation realization of the function f(.) where

$$|F(x) - f(x)| < \varepsilon \tag{5.2}$$

for all x in the input space and $\varepsilon$ is a small positive constant.

The universal approximation theory is directly applicable to the feed forward multi layer neural network with back propagation algorithm [42]. The same concept can be used to approximate any block cipher which is considered as a n-n discrete mapping between the input plaintext space and the output cipher-text space.

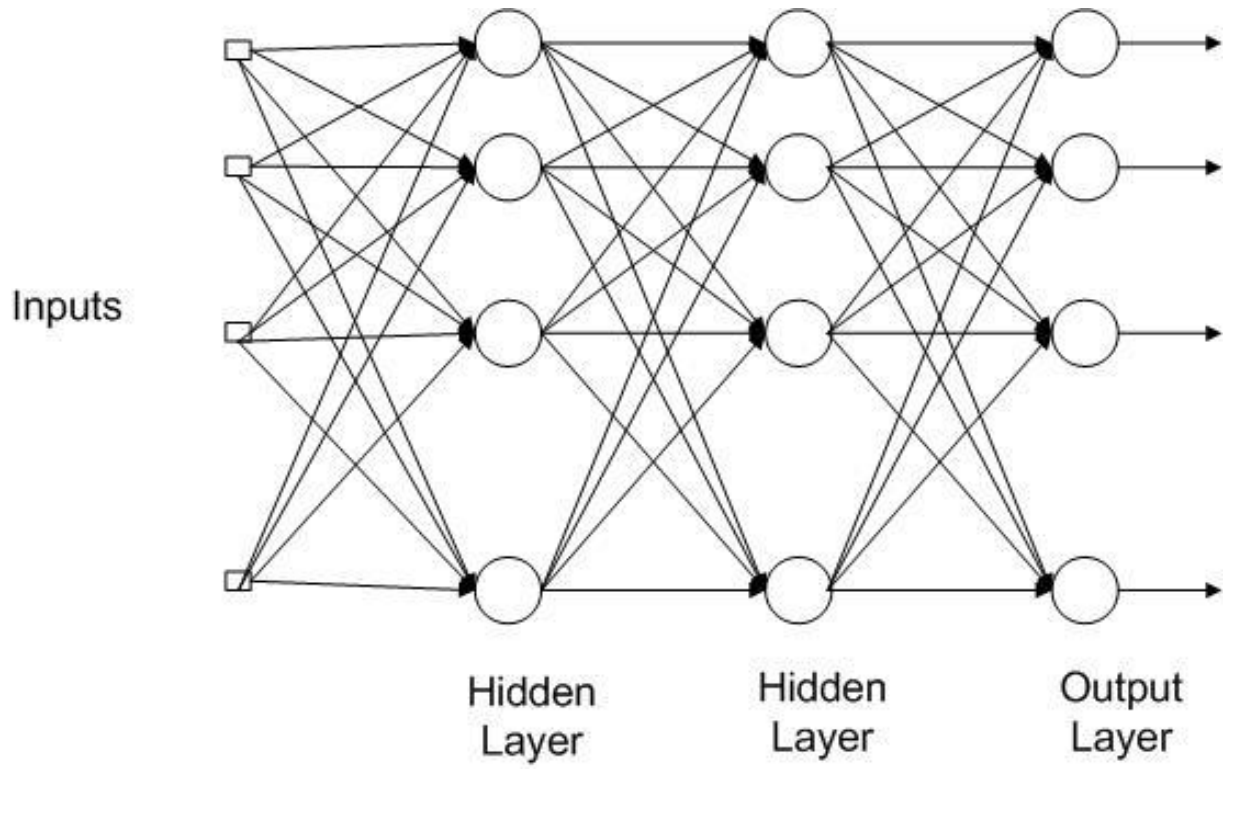

Figure 5.1: THREE LAYERS FULLY CONNECTED FEED FORWARD NEURAL NETWORK

## 5.2 HypCipher

In this section, a hypothetical Feistel type cipher is used where the round function f is chosen from the AES [5], [6] cipher, the round function is a mapping in the GF($2^8$) where this mapping is chosen of degree 8 and is non vulnerable to either differential, linear or interpolation cryptanalysis. The AES is not structured like a Feistel cipher; rather it is a Substitution Permutation Network structure (SPN). The same concepts of the attack can be applied on SPNs. One round of the HypCipher is shown in figure 5.2

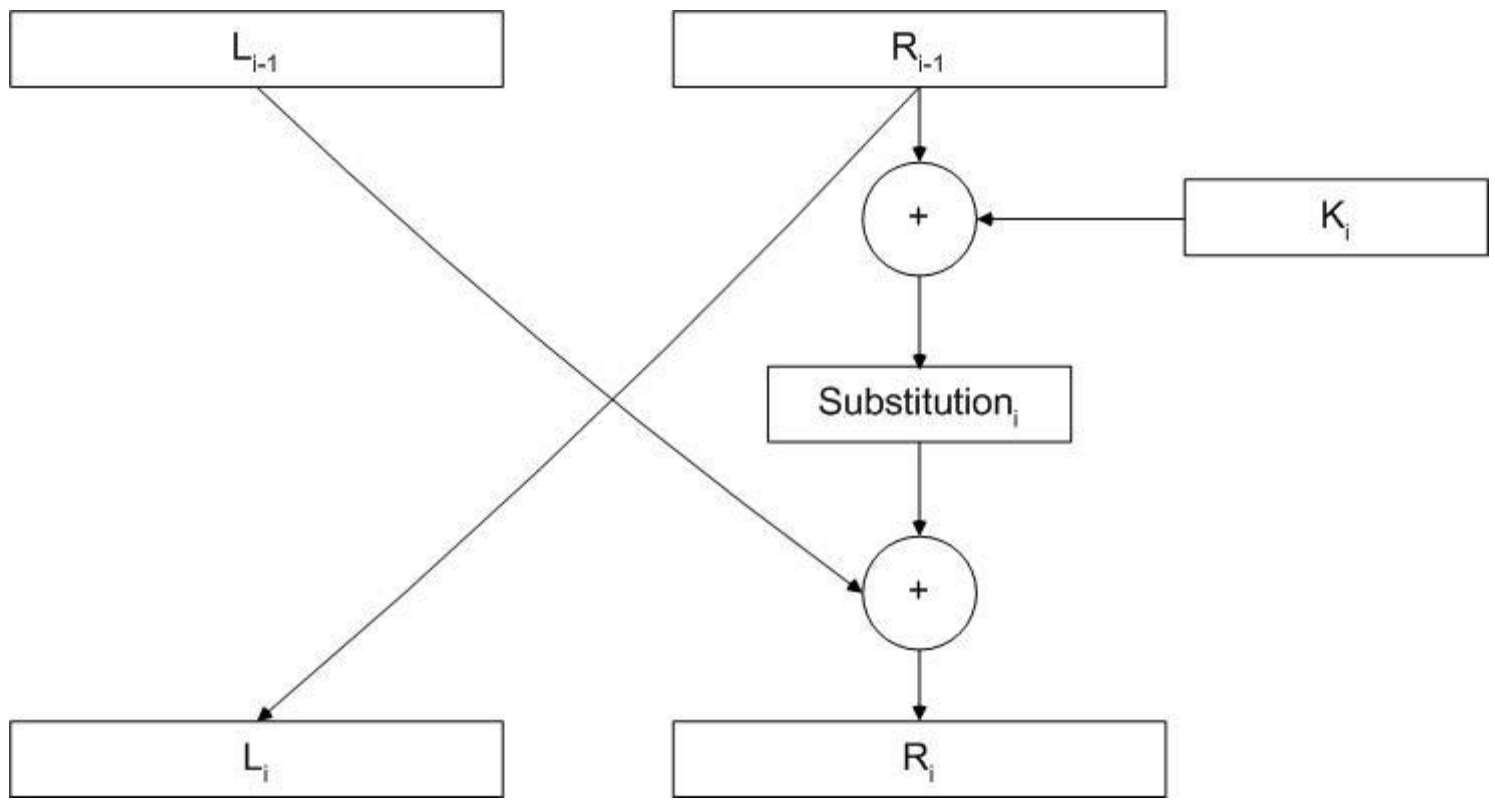

Figure 5.2: ONE ROUND OF THE HYPOTHETICAL CIPHER.

As can be seen from the figure, the key is mixed with the input using bitwise XOR function. A substitution, n to n mapping, is performed. We used the S box from the AES block cipher which is defined to get $x^{-1}$ in $GF(2^8)$ with polynomial basis as

$$a(x) = x^{-1} \bmod (x^8 + x^4 + x^3 + x + 1)$$

Then an affine transform is performed to rise up the degree of the overall system to overcome the interpolation attack [32].

The transform used is

$$b(x) = (x^7 + x^6 + x^2 + x) + a(x)(x^7 + x^6 + x^5 + x^4 + 1) \bmod (x^8 + 1) \ .$$

See chapter 2 for details.

The structure used here is more straight forward with respect to the Rijndael cipher and also has the same prosperities, that is still non vulnerable to differential, linear, and interpolation attacks.

The described structure is going to be called HypCipher in the rest of this work.

The round key is 8 bits and independent round sub-keys are used. The overall block size is 16 bits. The HypCipher described earlier is non vulnerable to all the mentioned attacks.

### 5.3 Neural Network Attacks

In this section, the proposed neural network attack is described. Let the description begins with the following theorem.

*Theorem 1: Consider an iterated block cipher with block size m, Express the output of the round just before the last as an m-m mapping and use a neural network to approximate this mapping using a set of known plaintext – partially decoded cipher-text pairs obtained from a guessed value of the last round key. Further assume that the trained neural network obtained from a wrongly guessed values of the last round key will produce a random output and uniformly distributed over the output bits when tried with other known plaintext – partially decrypted text pair, then there exist a neural attack to the cipher at hand such that the last round key value is obtained by dropping out the values of the wrong key.*

***Proof*.**

Let y be the output of the cipher given a particular input x and $\tilde{y}$ be the partially decrypted cipher-text with the correct value of the key K and let $\tilde{y}'$ be the one obtained by a wrong guess of the key, and assume that the last round function mapping is g(.) then

$$\tilde{y} = g^{-1}(k, y) \tag{5.3}$$

$$\tilde{y}' = g^{-1}(k', y) = g^{-1}(k', g(k, \tilde{y})) \quad (5.4)$$

Then the difference between the two values $(\tilde{y} - \tilde{y}')$ is the difference between applying the last round function twice, one with a correct key and the inverse with a wrong key, then from the nature of the cipher round function and its strong confusion and diffusion characteristics, this difference $(\tilde{y} - \tilde{y}')$ is expected to occur randomly and uniformly distributed over the output bits and hence one will unlikely be able to find an approximating neural network that finds the correct mapping for the (R-1) rounds block cipher.

In other words, a trained neural network with wrong guessed key value will actually be a random mapping that is not likely to guess values that were not trained by in a uniformly distributed environment.

On the other hand, the correct key value will produce a robust mapping resulting in much lower errors when trying new values not used to train the network.

### 5.3.1 Extracting Key Bits

The attack is now quit obvious, a set of known plaintext – cipher-text pairs is constructed where part of that set is used to train a neural network, the training is performed between the input and the partially decrypted cipher-text to the last round using an assumption of the last round key.

The rest of the set is used to test the output of the network and a record of the error between the actual output of the neural network and the real output is recorded.

Each sub-key value is then tried and the minimum error key value is assumed to be the correct key or a local search can be done on the set of the minimum error values.

### 5.3.2 Example of applying the attack:

The attack was implemented on several variants of the HypCipher described earlier,

The network was chosen to be with one hidden layer. The Input size and the output layer size are inherited from the problem, recall that the neural network is going to learn the mapping produced by R-1 round of an R round iterated cipher. The input size is 16 and the output is only 8 since the right half of the cipher-text is the only half needed to be partially decrypted.

The hidden layer size was taken to be also 16 neurons with sigmoidal activation function. Many researches have been carried out in the area of selecting the hidden layer size and the universal approximation theory tells a little about that. Many heuristics goes around selecting something between the input and output sizes like the multiplication or the addition average. The hidden layer can be viewed as a feature extraction or domain remapping, for that reason the hidden layer size was chosen to be 12 neurons after some trails.

A set of known plaintext cipher-text pairs was generated and used to implement the attack. The number of pairs used increases with the number of rounds, and the following linear relation

$$M = kR + c \qquad (5.5)$$

can be used where M is the needed number of examples, R is the number of rounds and k, c are positive integers. The value of k depends on the algebraic degree of the cipher being attacked and a value of 1.5 of the degree of the cipher core function, 8 in our case, is found to give good results after some experiments.

The extra examples (c) are used to evaluate the trained network for possible correct key values.

The attack succeeded in all the cases of 2 rounds variant to find the correct key value. For 3 and 4 rounds variants, a second hidden layer was found to make the attack successful.

The following figures show the error produced by the neural network evaluating the extra examples using different key values. In figure 5.3, the correct last round key was (01100000) =96. The error value is calculated as the summed square error between the neural network output and the desired output.  As can be seen from the figure, the minimum error occurs at the correct key value

Sigmoidal function followed by hard limiter where the final output is considered 1 if the neuron output exceeds 0.8 and 0 if the neuron output falls below 0.2 was used to represent the binary mapping at hand

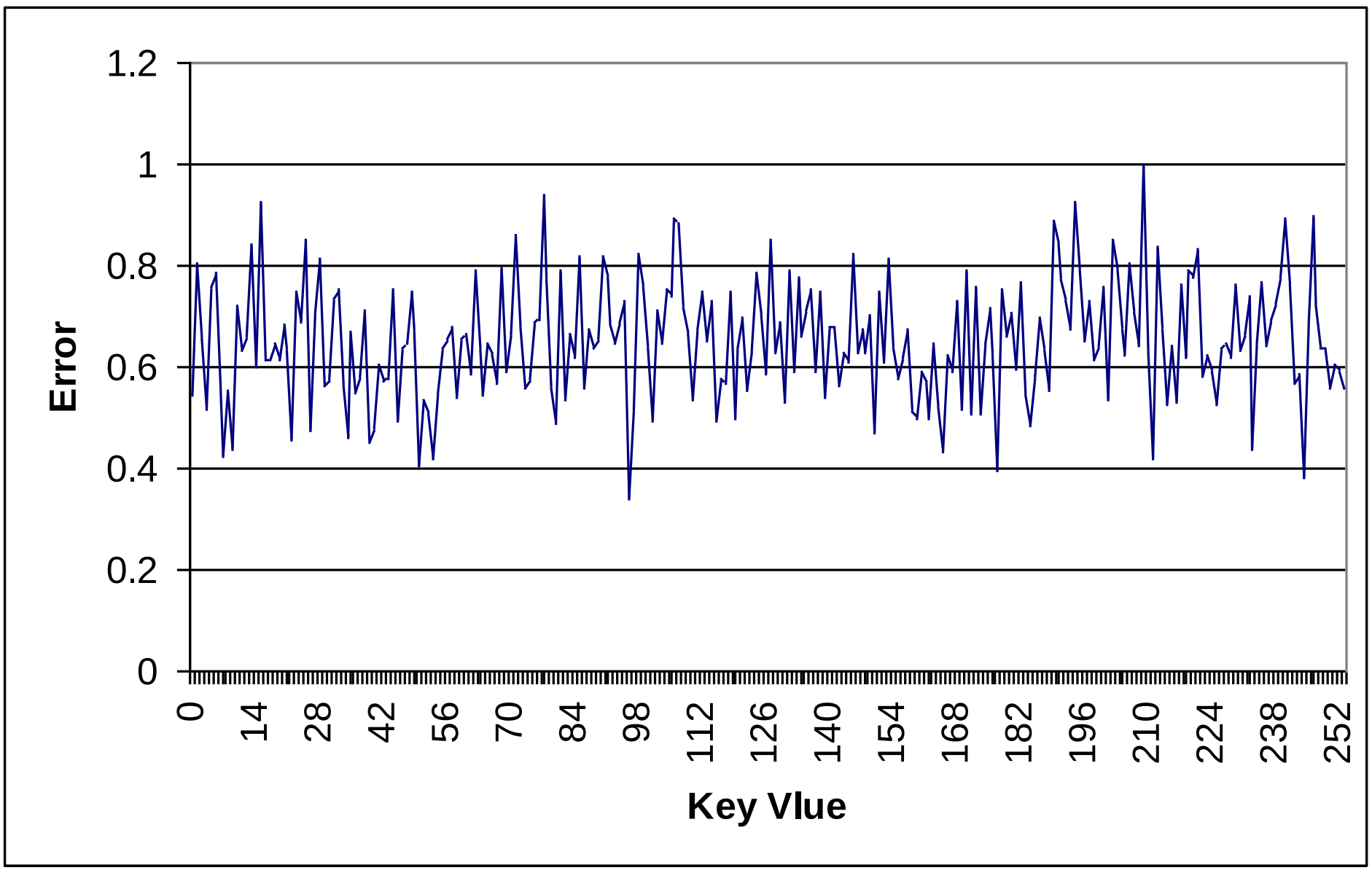

Figure 5.3: RESULTS FOR 2 ROUNDS, KEY VALUE = 96.

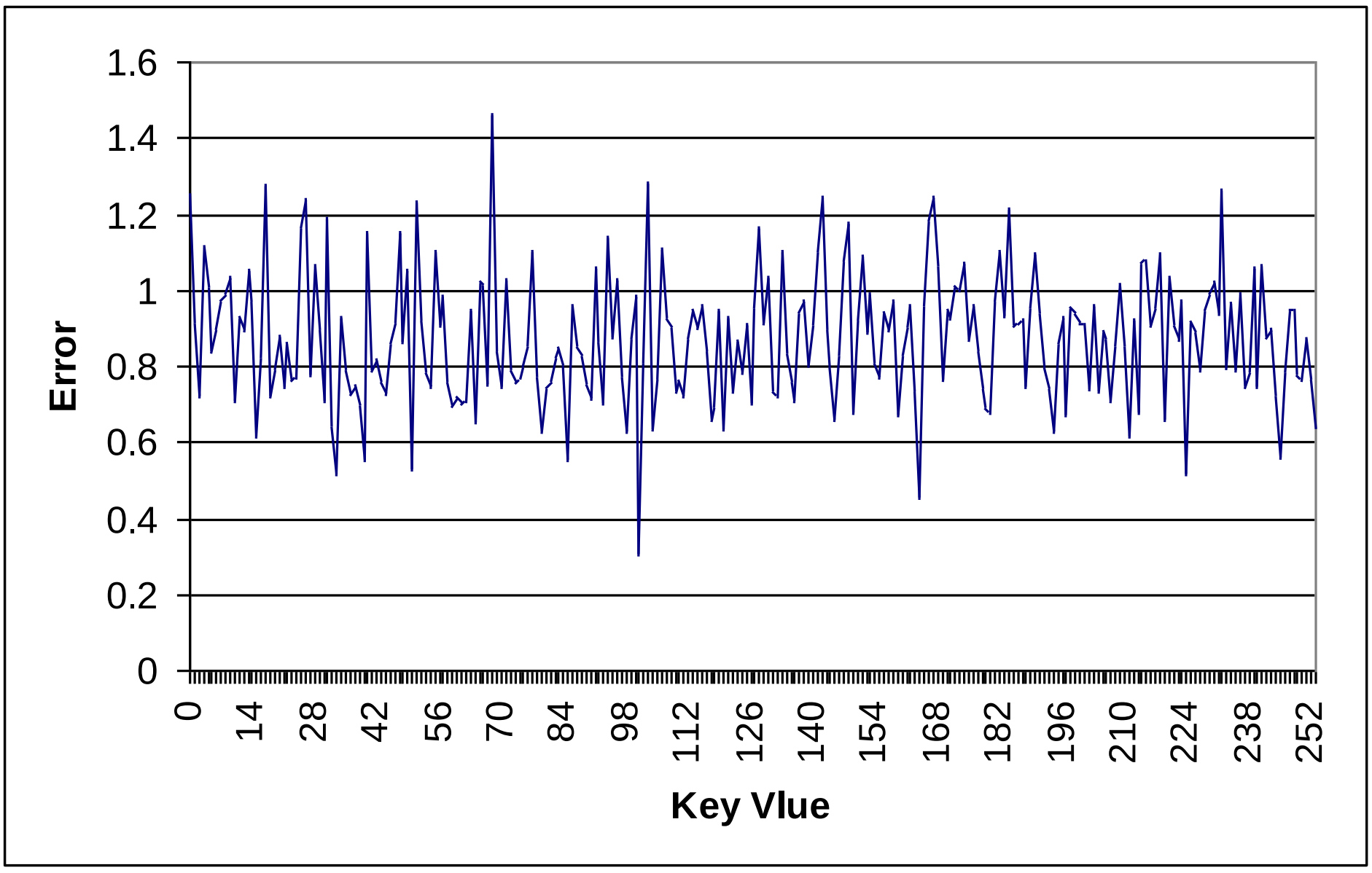

Figure 5.4: RESULTS FOR 2 ROUNDS, KEY VALUE = 101.

The results shown above in figure 5.3 and figure 5.4 were plotted as the neural network was trained for each sub-key value using the training set in random sequence.

Arranging the training set in ascending order results a better response as shown in figures 5.5 and 5.6. This is another analogy between neural network and interpolation where an ordered set of data points eases the calculations in interpolation. Also it was noted from experiments that the correct key value mapping is learned much faster by the neural network than the other wrong values as a consistent mapping is at hand.

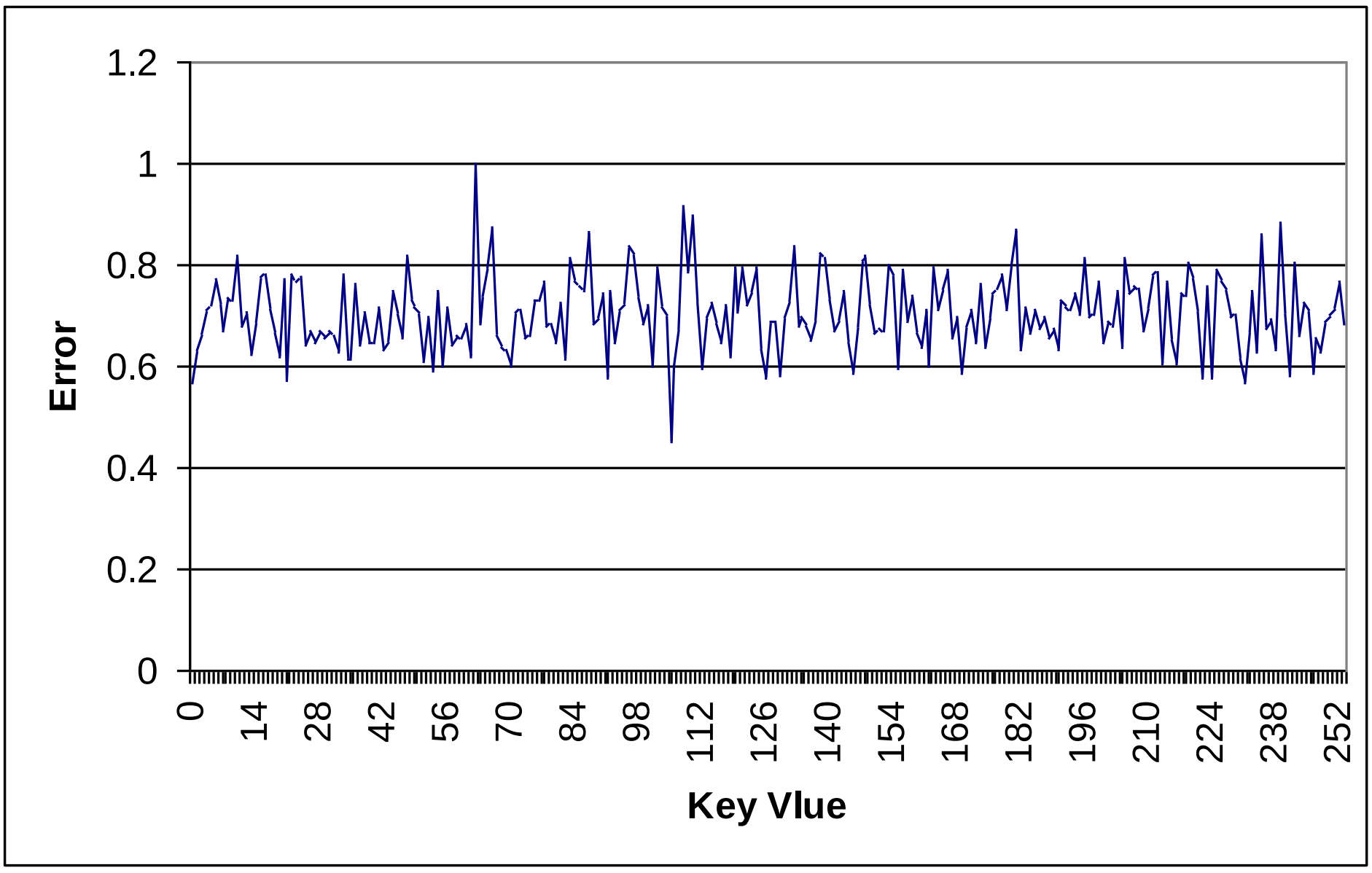

Figure 5.5: RESULTS FOR 2 ROUNDS, KEY VALUE = 105.

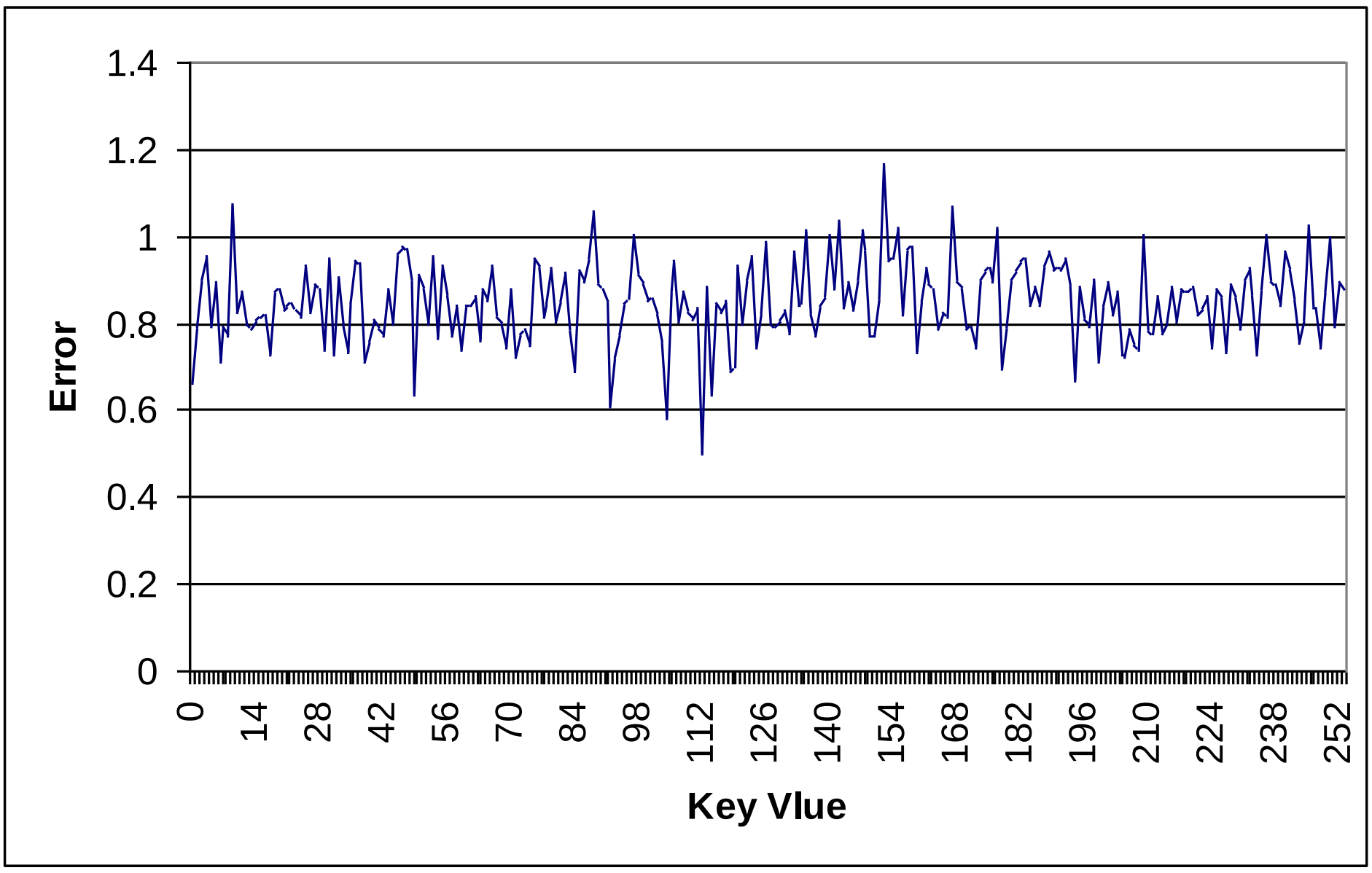

Figure 5.6: RESULTS FOR 2 ROUNDS, KEY VALUE = 112.

The attack on 3 and 4 rounds variants was not successful using the above neural structure, a more complicated structure using 2 hidden layers was successful. The reason is that using more rounds, complicates the mapping and reduces the common features a neural network to find a universal approximation to the mapping. Also more examples were needed to perform a correct mapping, from equation 5.5, about 53 examples were needed for 4 rounds.

The number of training examples is a little bit low with the neural attack than other attacks, but the absolute value should not be tricky as we are using here a very small variant with only 16 bit block size and only 8 bit round key for proving the concept. Figure 5.7 and figure 5.8 show the result for 3 and 4 rounds cipher respectively.

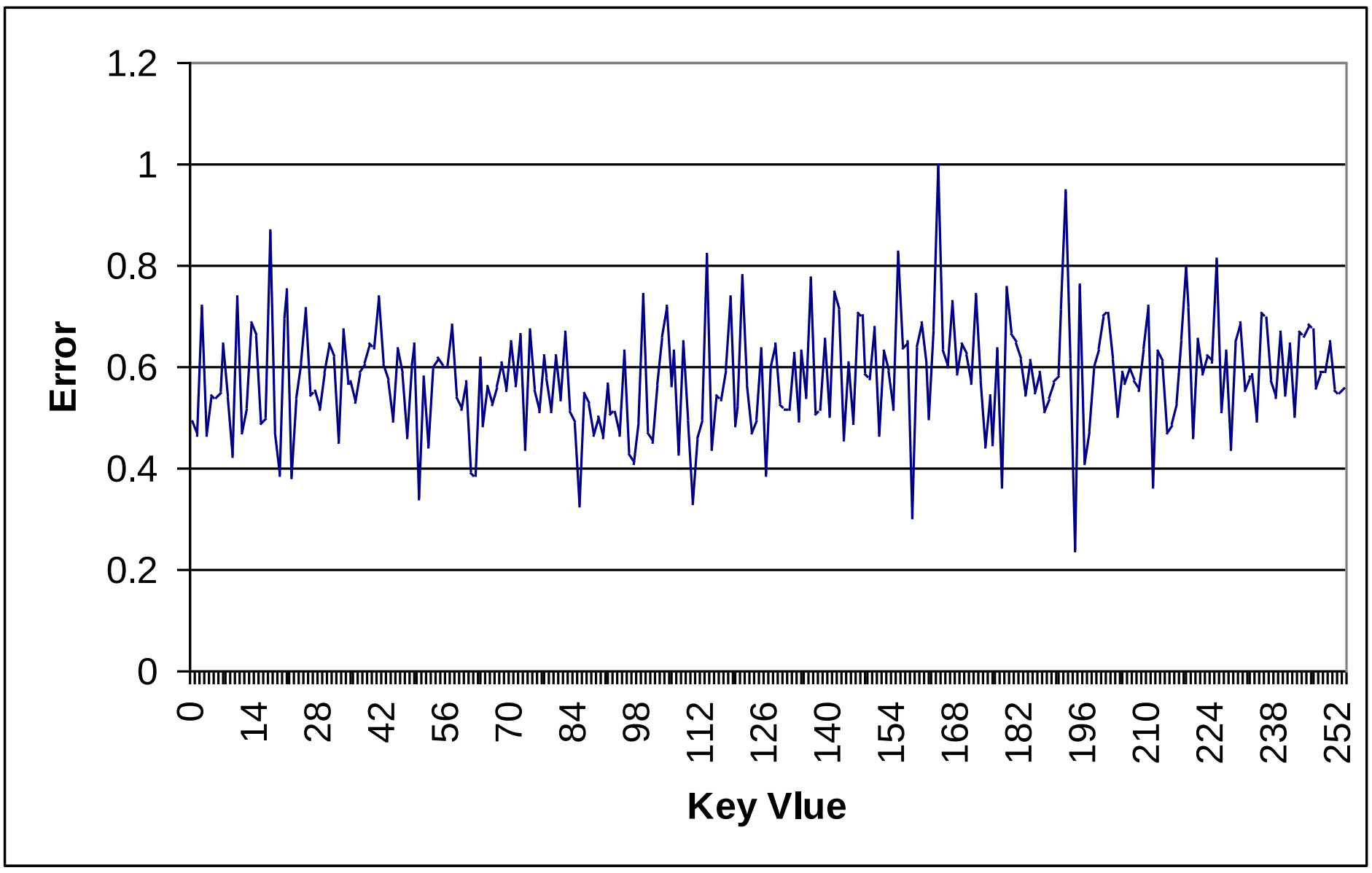

Figure 5.7: RESULTS FOR 3 ROUNDS, KEY VALUE = 194.

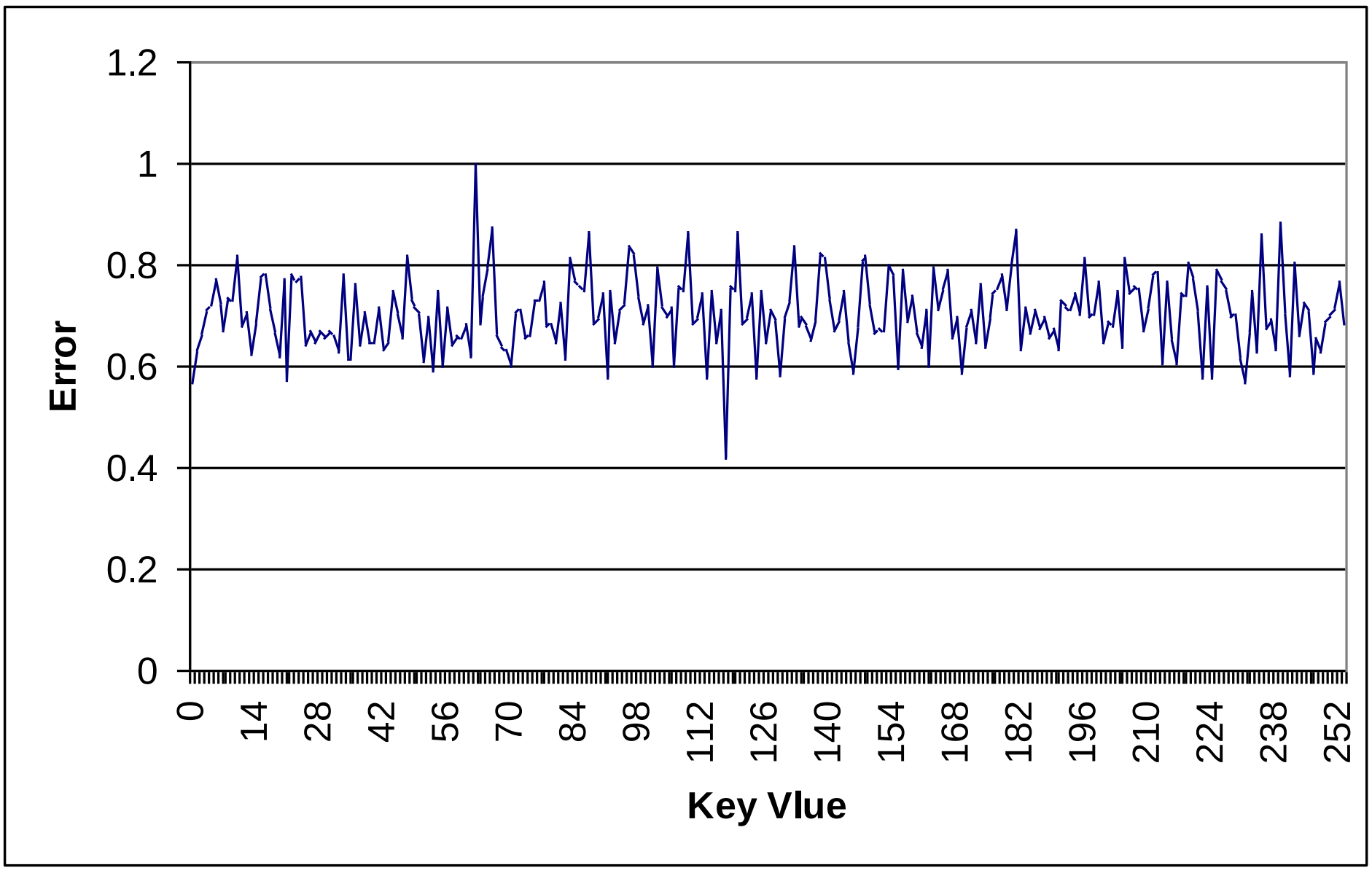

Figure 5.8: RESULTS FOR 4 ROUNDS, KEY VALUE = 117.

The attack is believed to be extendable to further rounds, but due to the limited computational power of a single PC, the capability of applying the attack for larger ciphers is limited. For example, the 2 hidden layer neural used for the last experiment needed about 15 hours to complete testing all the possible key value for the 4 rounds variant using a P IV 2.8 GHz with hyper threading technology micro processor. But due to the inherited parallel nature of neural networks, the attack is promising to be modified to a distributed version for larger ciphers and real ciphers.

## 5.4 Comparison

The proposed attack use a similar concept to the interpolation attack as both of them are using a way to approximate the R-1 rounds cipher and then try to find the correct last round key value.

The proposed attack has the benefit of being parallel by nature and can be easily extended to a distributed version. Also there are no equations to be solved or evaluated like the case of using Lagrange or Newton interpolation.

# Chapter 6: Neural Network Based Block Cipher

Neural networks have many interesting characteristics like parallelism, non linearity, and flexibility of their function mapping with high degree of freedom due to large number of weights. The above mentioned three properties of neural networks, are attractive to block cipher designers to be used as a confusion producing element. In this chapter, an approach of using neural networks as building block in a block cipher is proposed. The new cipher is called I-CRYPT. The proposed block cipher is based on the Feistel network structure with a neural network as the cipher core function. The core function in many block ciphers is implemented using S-Boxes that perform some sort of non-linear mapping between its input and its output domains. So the design parameters characterizing S-Boxes, and generally cipher core functions, are detailed in the next sections. Each design parameter is studied and its effect on the overall cipher security and performance is detailed. Different design methodologies are also presented and finally a proposal of a novel block cipher is given and studied.

## 6.1 Neural Networks for Domain Mapping

One application of neural networks is to be used as a domain mapping for complex functions. Neural networks can learn this mapping using several pairs of input - output from the mapping without knowing the original function.

In the approach of designing I-CRYPT, a neural network is used as 32 x 32 S-Box which is believed to be secured due to its large size. The approach of designing the S-Box neural network is to use a Random initialization with a key dependent mapping similar to the approach used in the Blowfish cipher, see chapter 2 for details.

## 6.2 Feistel Cipher Design parameters

Many design parameters characterize the Feistel block cipher. Modern block ciphers differ from each other in those parameters and basically they use the same model or some variations of it. In this section those parameters and their effect on the overall cipher are described.

### 6.2.1 Block Size

Larger block size means greater security keeping all other parameters fixed. This is because the statistical information of bigger chunks of the plaintext are confused using the core Function.

But the bigger the block size the slower the encryption or decryption processes. A block size of 64 bits is an accepted size for most block ciphers and recently a 128 bit is the preferred size.

### 6.2.2 Key Size and key scheduling

Key length is also a major component in measuring the security. In the case of lack of any other attack than brute force, key length is the only measure of security. Larger keys may also result in degraded encryption/decryption performance. A key of at least 128 bit is now believed to be secure enough at least against exhaustive search attacks.

Key scheduling is the process of generating the round sub-keys from the original key. The most secured algorithm is to use statistically independent sub-keys for each round, but this means a very large key. Most modern block ciphers use a key scheduling algorithm based on some criteria and many considerably secured ciphers use the cryptographic core itself in generating the sub-keys. Greater complexity in key scheduling usually means more security and the price of longer setup time. There are a lot of studies and works on attacking the key scheduling algorithm itself and hence degrading the security of the overall cipher.

### 6.2.3 Number of rounds

The basic criterion of Feistel cipher is to apply the product cipher more than one time to produce greater security. More rounds mean more security. From another point of view more rounds means higher order cipher. The order of the ciphers is an expression found in some attack literature namely interpolation attacks [32] and similar ones. In these attacks, the cipher is approximated into an algebraic function, polynomials in many cases. The higher the order of the cipher the more secure cipher against those attacks since higher order algebraic functions are harder to be solved and need more points to evaluate. The idea of product cipher means that if a round of the cipher is of degree n, then an r round cipher is of r x n degree which results in a harder cipher to be attacked.

### 6.2.4 Round Function F

The round function is the main component of the block cipher of Feistel type. It is usually called the core function. The basic purpose of F is to produce confusion and it is the main component indicating the algebraic degree of the cipher as described in the previous section. In general F must be as complex as possible to make cryptanalysis harder.

In the Rijndael proposal to NIST for a new block cipher to replace DES, a great work in mathematics especially the works of Nyberg [31] made some guidelines for the choice of the core S-Box. But the S-box chosen is still a small one. Schneier [28] in his blowfish cipher used a key dependent S-Box. The S-boxes used are generated based on the key and has a large setup time. The S-boxes are 8 x 32, still with small input size but large output size.

S-boxes are usually implemented by lookup tables and input size of 8 bits is most preferred as the lookup table can be implemented using small memories with 256 words size.

In this thesis, a neural network based round function is proposed which is a generalization of S-Boxes with larger inputs and outputs. The core neural network is key dependant but with a small setup time.

### 6.2.5 Design Criteria for F [2]

#### *6.2.5.1 Avalanche effect*

Avalanche linguistically means a huge unexpected action happening due to a small cause. In block cipher design the avalanche effect is a measure of how many bits are changed in the cipher-text if one bit of the plaintext or the key is changed. It is clear that more avalanche ratio means more security. One design criterion of DES S boxes was to introduce about 50% avalanche effect. A close look to each DES S-box, one can see that if a single bit of the input is changed, two bits of four output bits are changed.

#### *6.2.5.2 Strict Avalanche Criterion*

This is a more rigid definition of the avalanche effect. It was suggested by Webster [2] that any output bit j of an S-box should change with probability 1/2 if any single input bit i is inverted for all i, and j. A similar criterion can be easily adapted for any core function not using S-box scheme.

#### *6.2.5.3 Bit Independence Criterion*

Also Webster [2] suggested that output bits j and k should change independently when a single input bit i is inverted for all i, j, and k.

#### *6.2.5.4 Guaranteed Avalanche*

Heys proposed another criterion [2] that is guaranteed avalanche. The criterion is as follows, An S-box satisfies the Guaranteed Avalanche of order $\gamma$ if for one input bit change, at least $\gamma$ output bits are changed. A conclusion of $\gamma$ being between 2 and 5 produces strong confusion.

#### *6.2.5.5 Input-Output size of F*

A function F is n x m if it has n input bits and m output bits. In DES a 6 x 4 S-box is used. Blowfish has 8 x 32 size. Schneier [27] states that Larger S-boxes are more resistant to linear and differential cryptanalysis. But as mentioned before, the most preferred and easiest way of realizing F is by using S-boxes implemented as lookup table, this enforces small input size at least. In this thesis neural network as a realization of the core function F is proposed which overcomes the size limitation of the lookup table method.

#### *6.2.5.6 Design methodologies of F*

The core function F is basically a mapping between two binary domains with sizes of m x n. As mentioned earlier S-boxes are the most widely used realization technique. As Robshaw quoted in [43], there are four approaches to design an S-box

- **Random:** using some pseudo random number generator to generate the s-box lookup table. This approach may result in an undesired properties for small s-boxes but it should be acceptable for larger ones.
- **Random with testing:** where the previous approach is used to construct an initial mapping, then by using the measuring criteria like those mentioned before, the design may be kept or thrown away.
- **Human-made:** where manual techniques with simple mathematics are used to construct the mapping. This was the approach used in the DES design and it is very difficult to be used for large S-boxes design.
- **Math-made:** where a solid mathematical foundations are used to produce a mapping with a predetermined characteristics. This approach was used in designing the S-box of Rijndael.

**Random and key dependent:** this approach is used by Schneier in the design of Blowfish [28]. The s-boxes are generated randomly then the key is used to tune them. This approach is not stated in [43] but it is widely accepted.

The last approach can be used to develop some how large S-boxes and the key dependence hardens the analysis of the entire cipher. In the case of differential cryptanalysis, the number of core function mapping to be analyzed are very large and it is very hard or approximately impossible to build a key recovery attack in this case as the structure itself varies with the key.

Another example where the core function is composed of smaller fixed components but with different possible structures is the Tornado cipher proposed by Abo Elfotouh in [44]. In that work a key dependent structure of the cipher is proposed and seems immune to linear and differential cryptanalysis.

## 6.3 The I-CRYPT 64 Block Cipher

The I-CRYPT block cipher is basically a Feistel cipher with a neural network as its core function. The basic cipher is a 64 bit block cipher with the ability to be extended to other sizes. The key size is 64 bits and the number of rounds must be even. A sub-key is generated for each round using a carefully chosen key scheduling algorithm.

The main benefits of the I-CRYPT over other block ciphers especially blowfish is its larger core function and its fast setup time compared to blowfish. The cipher can be used in applications where the key can be rapidly changed.

### 6.3.1 I-CRYPT 64 Core Function F

The core function of the I-CRYPT cipher is a neural network. The design of the network is as follows.

Number of layers: 1 layer
Number of neurons: 32 neurons
Neurons activation function: Hard Limiter
Network topology: Fully connected
Input layer size: 32
Each neuron has an extra bias input b. This structure performs a 32 x 32 bit mapping depending on the values of the weights and the inputs to the bias links.
In the I-CRYPT weights are chosen randomly at design time and round sub-key is applied as inputs to the bias links, see figure 6.2 (a). The value b is also chosen randomly at design time. The output of the neural network is hence key dependent.
The neuron activation function is highly non linear and is very hard to be approximated to a linear function. The overall structure can be implemented using only a 32*32+1 memory to hold all the weights and bias values resulting in approximately 1 k word memory. The weights are restricted to discrete values and a range form -128 to 127 is a good one. The overall memory required is 1025 bytes. In the case of a lookup S-Box a memory of $2^{32}$ words each 32 bits is needed or 16 Giga bytes.
The core neural network introduces a great confusion and diffusion effects. Each output bit is a function of all input bits and all the key bits. Permutation is inherited in the full connection scheme used in the construction of the neural network. Avalanche effect is also very high due to the structure of the cipher.
An interesting feature that requires a lot of future work is the ability to adapt the weights of the neural network to perform a certain mapping. This point involves a lot of work in selecting the learning algorithm, the quality measuring factors and so on. It is suggested that discrete weights and some sort of reinforcement learning with Hebbian learning method can be used.

### 6.3.2 I-CRYPT 64 Cipher Structure

The I-CRYPT cipher is a Feistel cipher. This choice was made due to that neural network as described earlier is not reversible by itself. It can not be verified that it is reversible until a complete mapping is performed and a one to one feature is shown. Generally this is not possible and is not needed as a Feistel cipher is reversible independently of the reversibility of the core function.

Figure 6.1 shows one round of the I-CRYPT cipher.

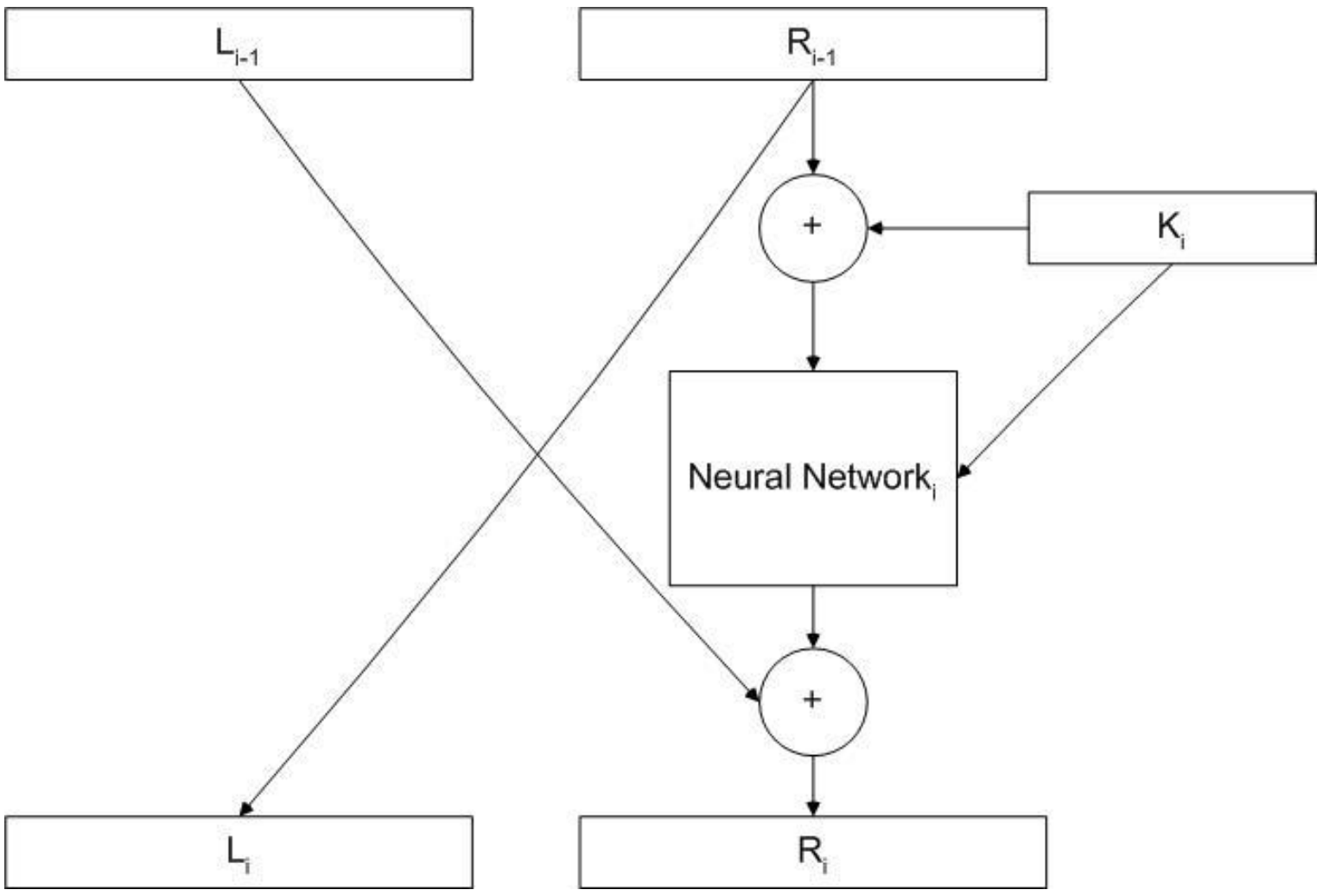

Figure 6.1: I-CRYPT ROUND

As can be seen from the figure the I-CRYPT round is a normal Feistel round where the key is mixed with the right half using bitwise XOR and is also used as an input to the core neural network. The neural network is then key dependent and for a single bit change, a change in the output occurs.

Another way of introducing the key is to use a 64 input neural network with the key applied as inputs to the neural network. Figure 6.2 (a) shows the first possibility where Figure 6.2 (b) shows the other one.

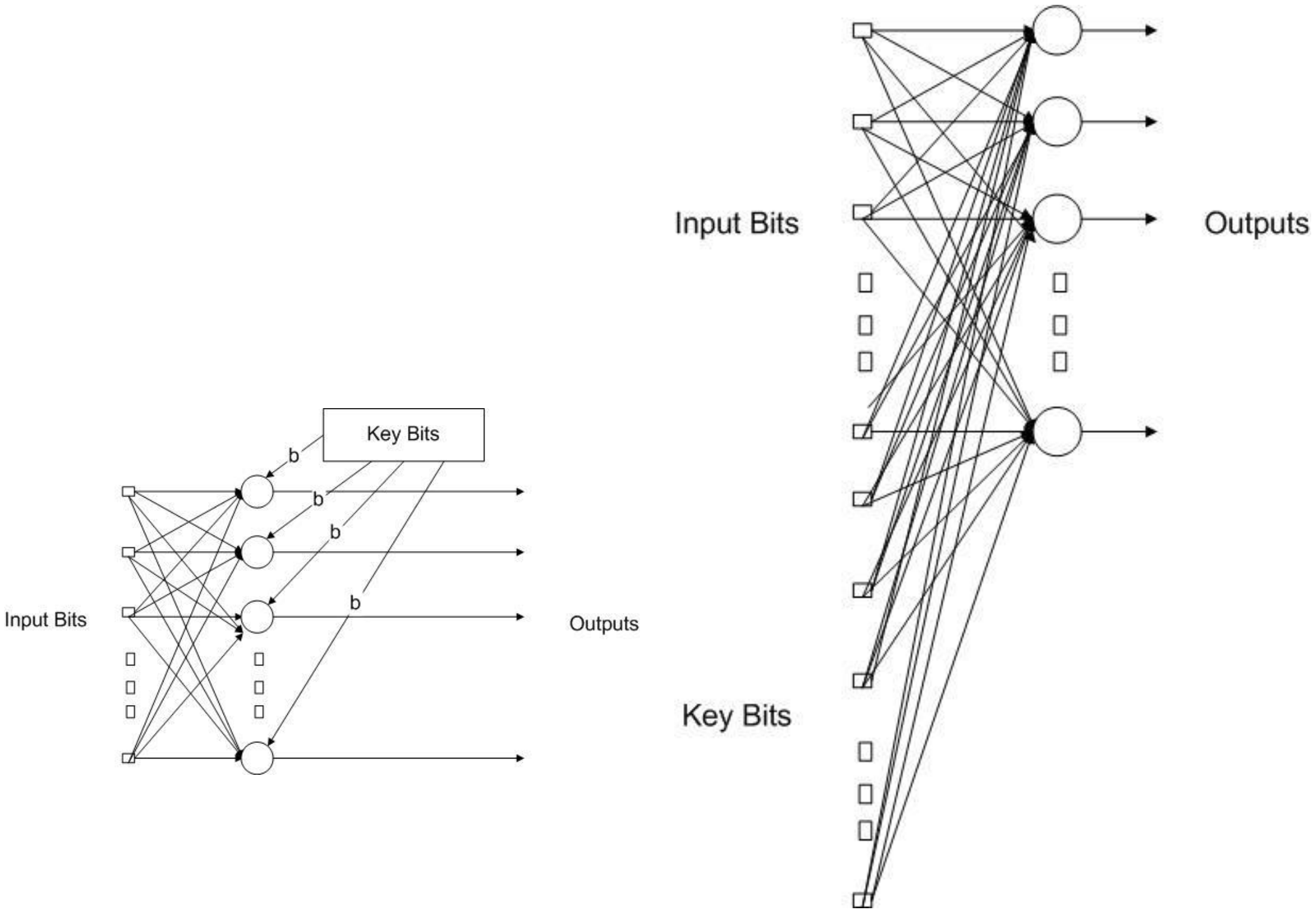

(a) (b)

Figure 6.2: POSSIBILITIES OF APPLYING THE KEY BITS IN THE I-CRYPT

The difference between the two options is in the way the output is affected by the key. Both neural networks are key dependent, but the later one have a greater avalanche effect for key change than the first one and hence it is better, but the number of weights is doubled and also the memory required.

### 6.3.3 Input Output Interpretation.

Input bits and key bits are presented to the network in there binary form but 0 bits are interpreted as -1 instead of just zeros. This is better because the zero value does not participate in the output of the network, it just leaves the decision to other one bits, while negative values tries to inhibit the output of each neuron it is connected to.

It is a mater of interpretation and some math justification will do the job quit good.

### 6.3.4 I-CRYPT 64 Number of Rounds

The proposed cipher Number of rounds is dependent on some design choices like performance, key scheduling algorithm, and of course security. In this version of the cipher, a 64 bit key is used. Each round needs only 32 bits. Simply the design requires an even number of rounds, each pair of rounds share a 64 bit round key half by half.

For the performance and security reasons, 10 rounds is believed to be secure and efficient in performance after several experimentation.

### 6.3.5 I-CRYPT 64 Key Scheduling

The I-CRYPT cipher uses its cryptographic core in key scheduling. The key scheduling algorithm is then as secure as the cipher itself. Schneier suggested using this technique in his proposal for TOWFISH cipher. The key schedule is very simple and is as follows.

1. Divide the user supplied 64 bit key into two halves.
2. Pass each half as an input to the neural network using the other half as the key.
3. XOR each of the output 32 bit block with the input block.
4. If an all-zero half appears, encrypt it using a two round I-CRYPT with the user supplied key as the two rounds key.
5. Replace the key used in 1 with the resulting key and repeat again until all rounds sub-keys are obtained.

### 6.3.6 Other I-CRYPT Variants

The I-CRYPT is very flexible and can be used with any block size. Variants of 128, 192, and 256 bits have been implemented and tested. The 64 bit version described earlier is solely intended for cryptanalysis purpose and for the cryptography community as a proof of concept.

### 6.3.7 Cryptanalysis of I-CRYPT

It is believed that most cryptanalysis of I-CRYPT is very difficult due to its key dependent core function. The large size of the function which performs 32 bit to 32 bit mapping with high nonlinearity in the hard limiter activation function and huge avalanche due to the structure of the network are all element that increase the security of the cipher to a high extent. Differential and Linear cryptanalysis are believed not to be possible on the 10 rounds cipher.

Further investigations are required on the class of neural based block ciphers.

### 6.3.8 Performance of the I-CRYPT

Neural networks are parallel by nature and hence a parallel architecture is the best to implement the I-CRYPT cipher, but on normal processors it is still efficient to implement neural networks in software.

The most time consuming calculation is the calculation of activation of each neuron involving multiplying the input to the weights and summing all of them.

In a very simplified version of I-CRYPT weights are restricted to -1 and 1 and hence all the multiplication process can be simplified to an XNOR and summation can be implemented using a byte lookup table where each byte is looked up in a 256 entry table that contains in each row the sum of all the bits where zeros are considered -1 as described earlier, see figure 6.3 and figure 6.4 for description. This version of I-CRYPT still has all its security and strength with the merit of extreme performance on normal PCs.

| X | Y | | X' | Y' | X'Y' | | XY |
|---|---|---|---|---|---|---|---|
| 0 | 0 | 0 as -1 | -1 | -1 | 1 | -1 as 0 | 1 |
| 0 | 1 | → | -1 | 1 | -1 | → | 0 |
| 1 | 0 | | 1 | -1 | -1 | | 0 |
| 1 | 1 | | 1 | 1 | 1 | | 1 |

Figure 6.3: MULTIPLICATION PERFORMED AS AN XNOR

| Number | Summation of bits (0 as -1) |
|---|---|
| 0 | -8 |
| 1 | -7 |
| 2 | -7 |
| 3 | -6 |
| 4 | -7 |
| . | |
| . | |
| 255 | 8 |

Figure 6.4: SUMMATION OF BITS OF 8 BITS BYTES WHEN 0 BITS INTERPRETED AS -1

# Chapter 7: Conclusion and Future Work

This thesis was concerned with the use of intelligent systems such as genetic algorithms and neural networks in information security cryptography mechanism. Three proposals were given in this research work and they were evaluated. Those proposals are

1. A proposed extension to the differential attack using cryptanalysis which resulted in a general extension that can be applied to any cipher vulnerable to differential cryptanalysis and at one quarter of the complexity of the original attack.
2. A proposed extension to the interpolation attack using back propagation neural networks which outperforms the original attack in that id does not need to solve or evaluate a huge number of equations and yet the novel proposal can attack more complicated ciphers than the interpolation cryptanalysis can attack
3. A novel block cipher based on a neural network as its core function is proposed. The proposed core function can be much larger than the S-boxes for their input output sizes, and hence more secure than smaller realizations, and yet outperforms the S-box lookup process and requires much lower memory for the same size.

## 7.1 Concluding Remarks

The following points summarize the concluding remarks

1. Cryptography and especially block ciphers importance in information security was cleared out.
2. It was shown how statistical techniques are limited in cryptanalysis for modern block ciphers.
3. Differential attacks was also described and detailed on the basic Substitution Permutation Network.
4. Also Interpolation attacks were introduced.

5. Genetic algorithms are shown to improve the performance of differential attacks and it was tested on the basic two models of most modern block ciphers, the Substitution Permutation Network and the Feistel network.
6. Proposed Neural networks cryptanalysis is shown to work with ciphers non vulnerable to differential cryptanalysis.
7. Finally a neural network block cipher is proposed and we claim it is very secure due to its large size and highly non linear core function realized by the neural network.

## 7.2 Future Work

Many points still open for future work, for examples:

1. Using parallel competing genetic algorithms on both the key space and chosen plaintext space to further optimization of differential attack
2. Using the genetic algorithm and neural network in a cooperative attack on complicated ciphers
3. Applying and extending the proposed attack to more ciphers.
4. Increasing the performance of the proposed I-CRYPT cipher.
5. Studying the point of applying reinforcement learning to tune the weights of the I-CRYPT core neural network to achieve one or more of the design criteria of core functions.
6. Studying other neural network models, especially the Hopfield optimization network, for cryptanalysis
7. Studying hardware software co design for implementing the proposed I-CRYPT cipher.

# Appendix (A): Artificial Intelligence

## A.1 Genetic Algorithm

Genetic algorithms are considered as one of the most efficient search techniques. Although they do not offer an optimal solution, their ability to reach a suitable solution in considerably short time gives them their respectable role in many artificial intelligence and searching techniques. The following sections introduce genetic algorithms and describe their characteristics.

### A.1.1 Genetic algorithms

In this section a brief description of genetic algorithm and its characteristics is introduced [45]. Holland (1969, 1975) proposed the use of genetic algorithm as an efficient search mechanism in artificially adaptive systems. This procedure was designed to incorporate specific operators, emphasizing especially crossover and inversion.

The easiest way to consider a GA is in the context of a function optimization problem. The goal is to search through a space of function inputs (represented by some binary string) to find the input which maximizes a given target function. A genetic algorithm begins with a randomly selected population of function inputs represented by strings. The GA uses the current population of strings to create a new population such that the strings in the new population are on average "better" than those in the current population.

Three processes which have a parallel in human genetics are used to make the transition from one population generation to the next. They are selection, crossover, and mutation. The basic GA cycle based on these three processes is shown in Figure A.1.

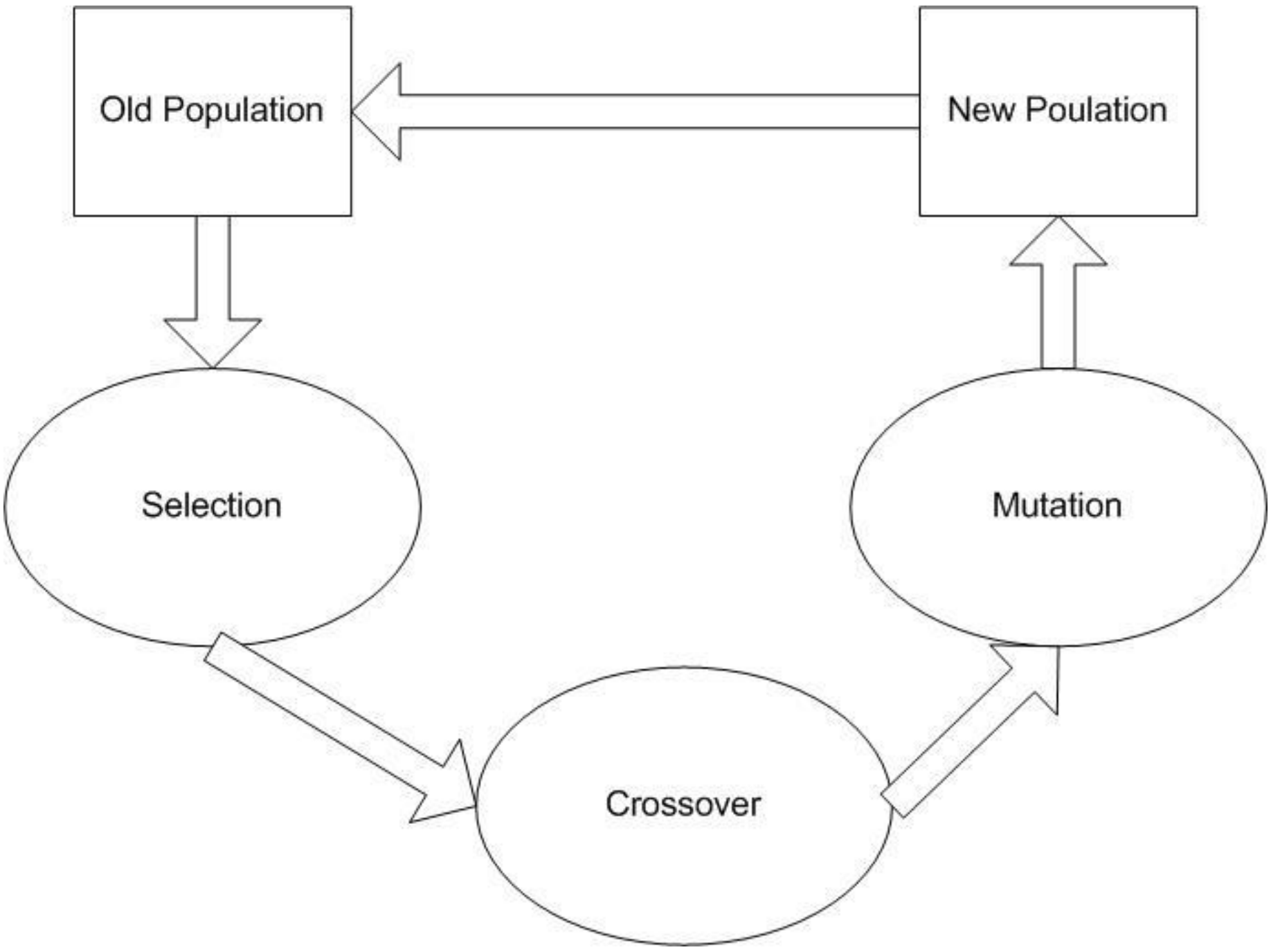

FIGURE A.1 THE BASIC GENETIC ALGORITHM CYCLE.

The genetic algorithm has proven to be very effective in some search problems [45]. While it may seem to be a random search, in fact, the improvement in each generation indicates that the algorithm provides an effective directed search technique.

### *A.1.1.1 Implementation of GA*

Genetic algorithms are typically implemented as follows,

The problem to be addressed is defined and captured in an objective function that indicates the fitness of any potential solution.

A population of candidate solutions is initialized subject to certain constraints. Typically, each trial is coded as a vector x, termed a chromosome, with elements being described as genes and varying values at specific positions called alleles. All solutions should be represented by binary strings. For example, if it is desired to find the scalar value x that maximizes

$F(x) = -x^2$,

Then a finite range of values for x would be selected and the minimum possible value in the range would be represented by the string [0... 0], with the maximum value being represented by the string [1... 1]. The desired degree of precision would indicate the appropriate length of the binary coding.

Each chromosome, $X_i$, i = 1,….,P, in the population is decoded into a form appropriate for evaluation and a fitness score is assigned to each chromosome, $\mu(x_i)$, according to the objective.

a probability of reproduction, pi, i = 1,..,P, is assigned to each chromosome, so that its likelihood of being selected is proportional to its fitness relative to the other chromosomes in the population. If the fitness of each chromosome is a strictly positive number to be maximized, this is traditionally accomplished using roulette wheel selection, Figure A.2.

| % total | Function | String | Number |
|---|---|---|---|
| 14.4 | 169 | 011110110111 | 1 |
| 49.2 | 576 | 011010010111 | 2 |
| 5.5 | 64 | 101101111011 | 3 |
| 30.9 | 361 | 111010010111 | 4 |
| 100 | 1170 | | Total |

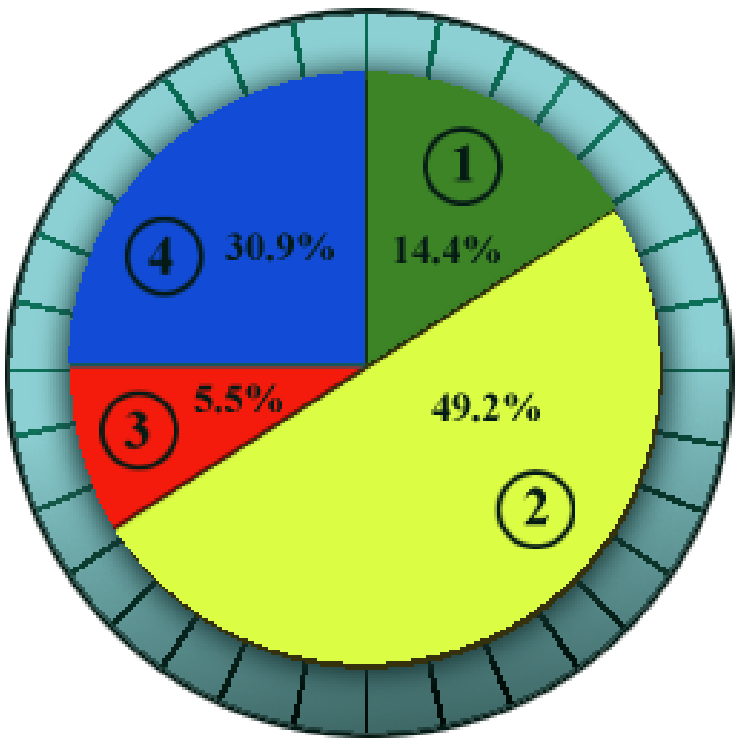

FIGURE A.2 ROULETTE WHEEL SELECTION MECHANISM

According to the assigned probabilities of reproduction, p, i= I, P, a new population of chromosomes is generated by probabilistically selecting strings from the current population. The selected chromosomes generate offspring via the use of specific genetic operators, such as crossover and bit mutation. Crossover is applied to two chromosomes (parents) and creates two new chromosomes (offspring). This is done by selecting a random position along the coding and by exchanging the section that appears before the selected position in the first string with the section that appears after the selected position in the second string, and vice versa. Figure A.3 shows this process. Bit mutation simply offers the chance to flip some bits in the coding of a new solution, see figure A.4.

| Crossover Point | | | |
|---|---|---|---|
| Parent #1: | 1101 | 0111101 | Offspring #1: 10100111101 |
| Parent #2: | 1010 | 0000100 | Offspring #2: 11010000100 |

FIGURE A.3 THE ONE-POINT CROSSOVER OPERATOR APPLIED TO TWO PARENTS.

| | | | | | | | | | | |
|---|---|---|---|---|---|---|---|---|---|---|
| Old String | 0 | 1 | 0 | 1 | 0 | 0 | 1 | 0 | 1 | 1 |
| New String | 0 | 0 | 0 | 1 | 1 | 0 | 0 | 1 | 1 | 1 |

FIGURE A.4 MUTATION PROCESS.

Typical values for the probabilities of crossover and bit mutation range from 0.25 to 0.95 and 0.001 to 0.01, respectively.

The process is halted if a suitable solution is found or if the available computing time is expired. Otherwise, the process goes to step 3, where the new chromosomes are scored and the procedure iterates.

For example, suppose the task is to find a vector of 100 bits {0, 1} such that the sum of all the bits in the vector is maximized. The objective function could be written as

$$\mu(x) = \sum_{i=1}^{100} x_i ,$$

Where x is a vector of 100 symbols from {0, 1}. Such vector x could be scored with respect to μ(x) and would receive fitness ranging from zero to 100. Let an initial population of 100 parents be selected completely at random and subjected to roulette wheel selection in light of μ(x), (fitness function) with the probabilities of crossover and bit mutation being 0.8 and 0.01, respectively. The process will rapidly converge to a vector of all l's. A number of issues must be addressed when using a genetic algorithm [45].

- Using of binary representation versus floating point representation especially in real values optimization.
- Selection in proportion to fitness can be problematic. Since (1) Roulette wheel selection depends on positive values, and (2) simply adding a large constant value to the objective function can eliminate selection. Other problems of fairness may arise for this selection technique.
- Premature convergence is another important concern in genetic algorithms. This occurs when the population of chromosomes reaches a configuration such that crossover no longer produces offspring that can outperform their parents, as must be the case in a homogeneous population. Under such circumstances, all standard forms of crossover simply regenerate the current parents. Any further optimization relies solely on bit mutation and can be quite slow.

Although many open questions remain, genetic algorithms have been used to address diverse practical optimization problems successfully.

### A.1.2 Example of applying GA to function optimization.

In this section, the results from the GA simulator, which has been developed as a part of the thesis supplementary work, are introduced. A complex function of two variables is used to test the process. Consider the following function,

$$f(x_1, x_2) = 21.5 + x_1 . \sin(4\pi x_1) + x_2 . \sin(20\pi x_2)$$

and the ranges for x1,x2 are –3<x1<12.1 and 4.1<x2<5.8,

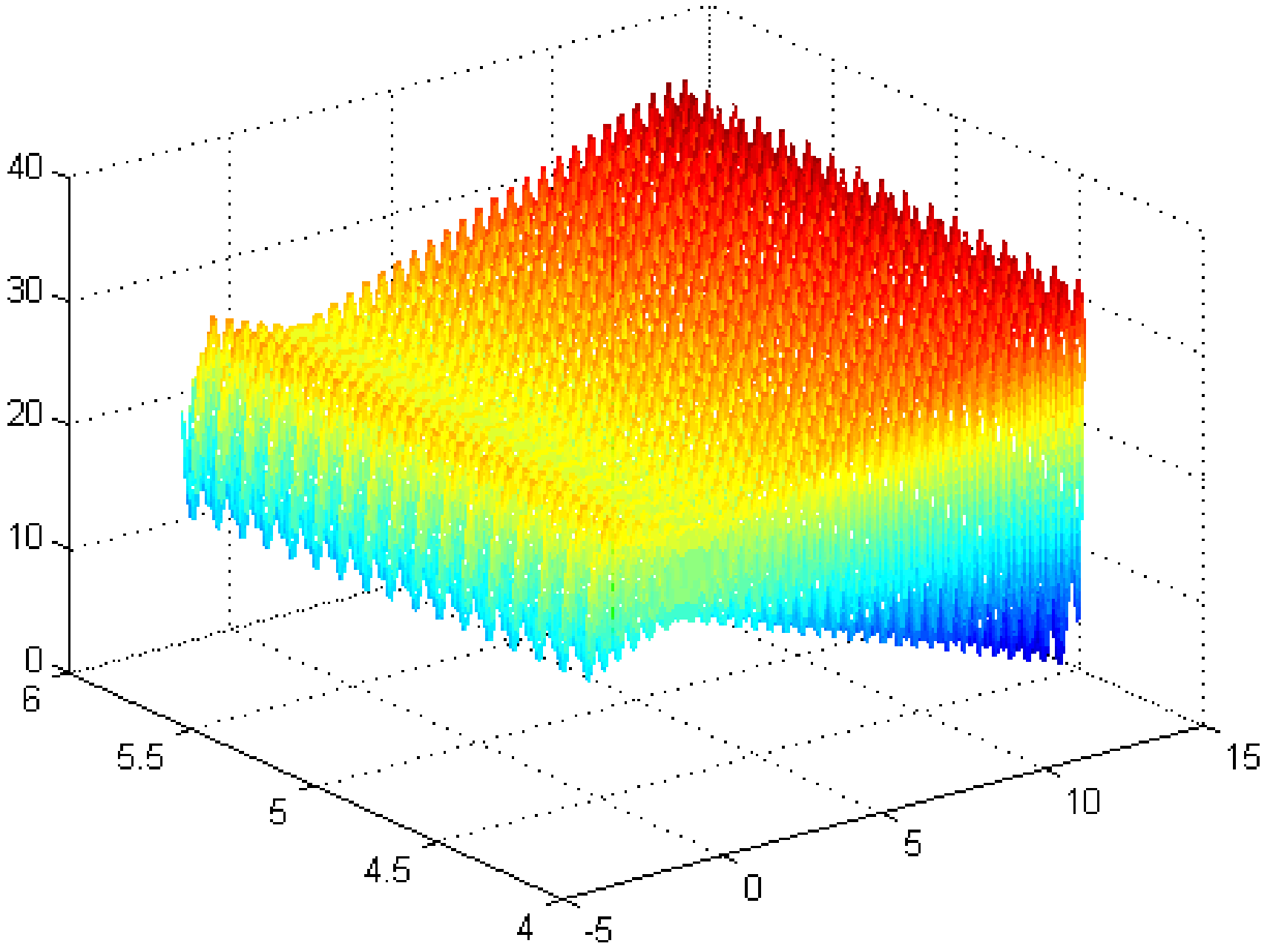

FIGURE A.5 A GRAPH OF A FUNCTION TO BE MAXIMIZED

Figure A.5 shows a graphical representation of the given function

Let us consider finding the values of x1 and x2 such that the function is maximized. Assume that the required precision is four decimal places for each variable. That means the range of x1 should be divided into 151000 equal size ranges and also for x2 17000 slides are required. Resulting in 2567 * 106 possible values for the combination of the two variables. If we try to find the values for x1 and x2 giving maximum value of the function f, we find the maximum value of f equals 38.9354 for x1=12.0999 and x2= 5.7227

This result was found after 01 hours, 04 minutes, and 59 seconds using a Pentium 233 PC and a simple C++ program written for this purpose. The code of the program is as follows,

```
for (float x1=-3.0;x1<=12.1;x1+=0.0001)
      for (float x2=4.1;x2<=5.8;x2+=0.0001)
      {
      fit=21.5+x1*sin(88.0/7.0*x1)+x2*sin(440.0/7.0*x2);
                  if(fit>maxfit)
                  {
                        maxfit=fit;
                        bestx1=x1;
                        bestx2=x2;
                  }
      }
```

Now let us try to find a solution for the same problem using GA. First the solutions to be found must be decoded into binary strings that represent the different values for x1 and x2. For x1 we need 18 bits and for x2 we need 15 bits resulting a total of 33 bits which is going to be the length of our chromosome. The decoding formulas for x1 and x2 are as follows,

$$X_1 = -3.0 + decimal(chromo[1:18]) * \frac{12.1-(-3.0)}{2^{18}-1}$$

$$X_2 = 4.1 + decimal(chromo[19:33]) * \frac{5.8-(4.1)}{2^{15}-1}$$

Now let us assume the following parameters for our GA problem

| | |
|---|---|
| Population size: | 50 chromosome |
| Chromosome size: | 33 bits |
| Maximum number of generations: | 200 generations |
| Crossover probability: | 0.25 |
| Mutation probability: | 0.1 |

Running the simulator gives the following results,

Maximum value obtained was 37.5385 for generation no. 160 at values x1=10.6324 and x2=5.51564. The time consumed for the 200 generations was 2.884 sec

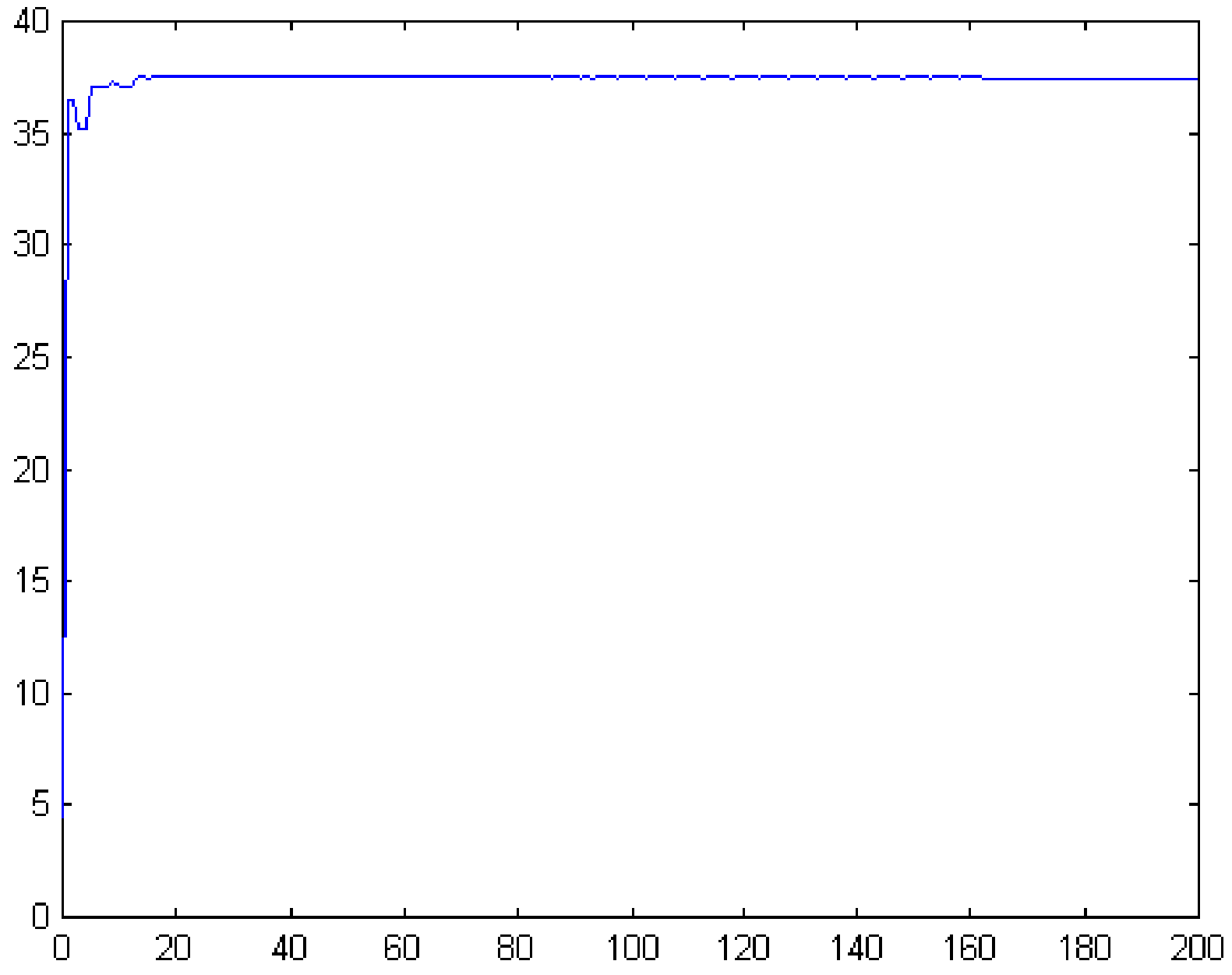

FIGURE A.6 CHANGE OF FITNESS VS. GENERATION

Figure A.6 shows the fitness value variation for each generation. The best chromosome is the only one plotted.

Figure A.7 focuses on the changes around the optimal value found

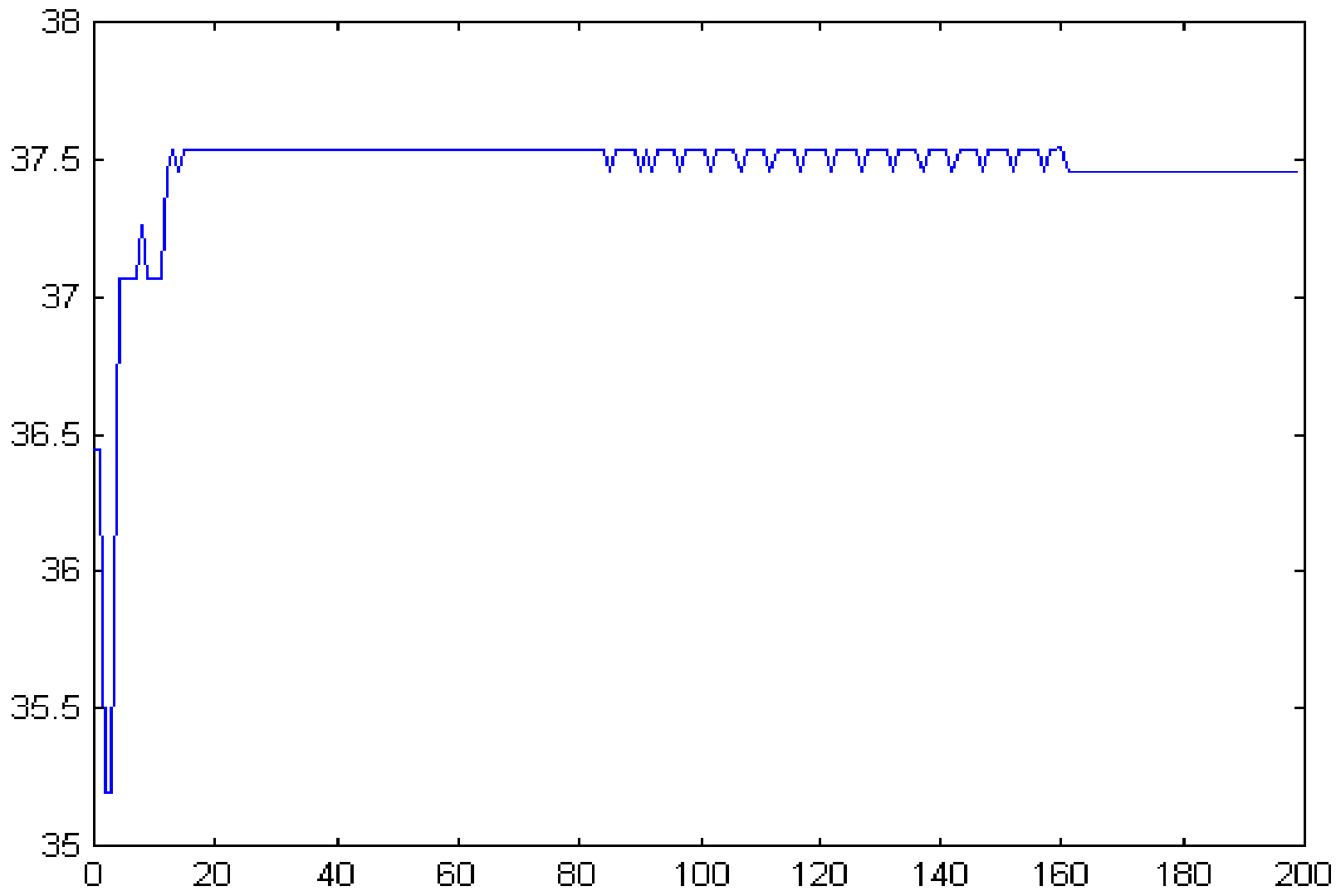

FIGURE A.7 VARIATION OF THE BEST FITNESS VALUE.

As seen from the figure, GA actually oscillates around the best value. The algorithm in its basic description does not hold the best value from a generation to another, and so the best value may be conserved in any generation during the process.

### A.2 Neural Networks

Artificial neural networks (ANNs) [46]-[50] are highly parallel interconnections of simple processing elements or neurons that function as a collective system. There exist various problems in pattern recognition that humans seem more efficient in solving as compared to computers. Neural networks may be seen as an attempt to emulate such human performance. These networks can be broadly categorized as those that learn adaptively by updating their connection weights during training and whose parameters are time-invariant [51].

In the following Section, we will give a general idea about ANNs.

### A.2.1 Neural Networks Approaches

Artificial neural networks are physical cellular systems that can acquire, store, and utilize experimental knowledge [52]. An artificial neural network is a structure that tries to mimic the human brain. There are different types of networks; each is designed to solve a specific problem. In this chapter Back Propagation Networks (BPN) are reviewed in details.

### A.2.2 Processing Elements

Neurons perform as summing and nonlinear mapping junctions. In some cases they can be considered as threshold limits that fire when their total input exceeds certain bias level. Neurons usually operate in parallel and are configured in regular architectures. They are often organized in layers, and feedback connections both within the layer and toward adjacent layers are allowed. Each connection's strength is expressed by a numerical value called a weigh, which can be modified [52].

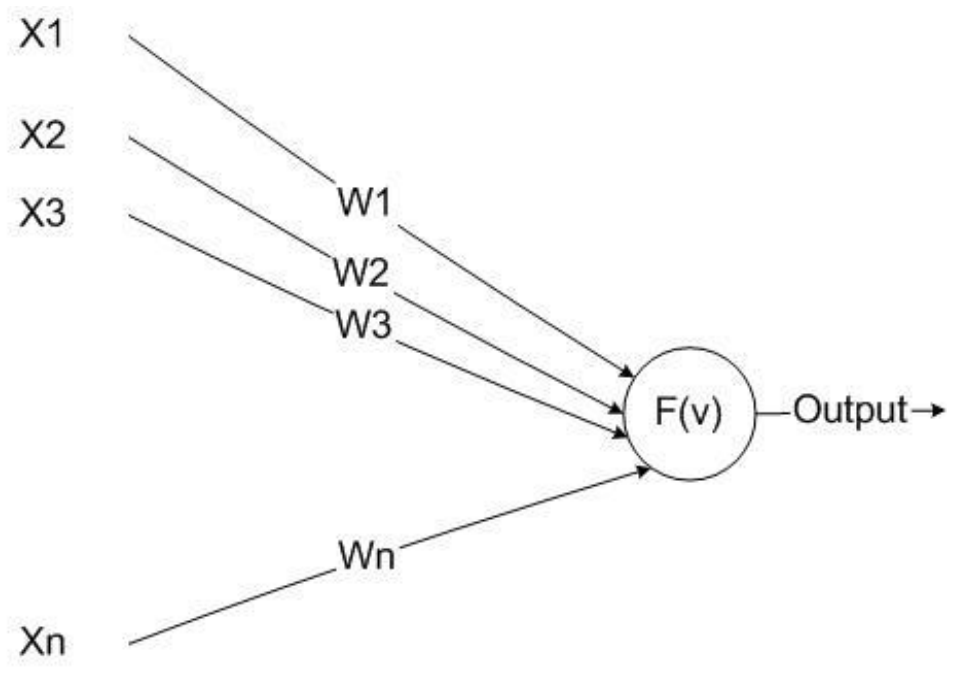

FIGURE A.8 PROCESSING NODE AND SYNAPTIC CONNECTIONS.

Every neuron model consists of a processing element (PE) with synaptic input connections and a single output. The signal flow of neuron inputs is considered to be unidirectional as from input to output indicated by arrows. Figure A.8 shows a general neuron symbol. This symbolic representation shows a set of weights and the neuron's processing unit. The neuron output signal is given by the following relationship:

$$o = f(v) \tag{A.1}$$

The variable net is defined as:

$$\text{v} = \sum_j w_j i_j \tag{A.2}$$

The function f(v) is often referred as an activation function.

Observe from Equation (A.1) that the neuron as a processing node performs the operation of summation of its weighted inputs, or the scalar product of both the input and weight vectors to obtain net. Subsequently, it performs the nonlinear operation f(v) through its activation function. Typical activation functions used are:

$$f(v) = \frac{1}{1 + e^{-\lambda v}} \tag{A.3}$$

Or

$$f(v) = \begin{cases} 1, & v > 0 \\ 0, & v < 0 \end{cases} \tag{A.4}$$

Where λ is proportional to the neuron gain determining the steepness of the continuous function f(net) near v = 0. The continuous activation function is shown in Figure A.9 for λ=1

The soft-limiting activation function (A.3) is often called sigmoidal characteristics, as opposed to the hard-limiting activation function given in (A.4). The hard-limiting activation function describes the discrete neuron model.

Essentially, any function f(v) that is monotonically increasing and continuous such that $f(v) \in (0,1)$ can be used instead of (A.3) as an activation function.

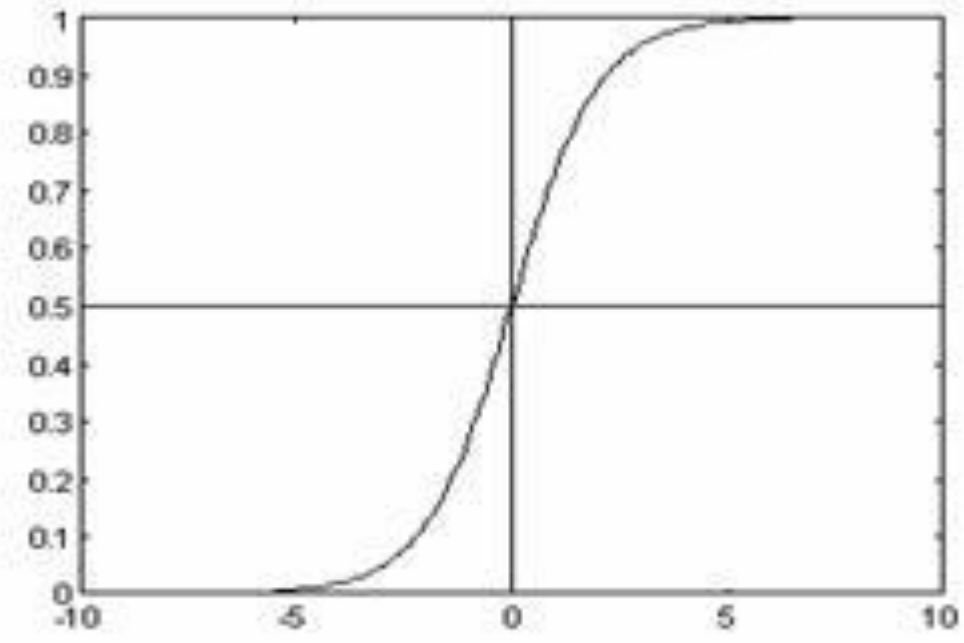

FIGURE A.9 UNI-POLAR ACTIVATION FUNCTION OF A NEURON.

### A.2.3 Network Topology

There are several network topologies. Each topology is designed to solve a specific problem or to represent certain point of view. Each topology has some drawbacks and benefits. These topologies can be categorized in three main categories: Fully interconnected, feed forward connection, and mixed connectivity.

In Feed forward connection, the neurons are organized in multiple layers, so that a neuron in a certain layer is connected only to neurons on the next layer as shown in Figure A.10. Multilayer perceptrons [53] belong to this category.

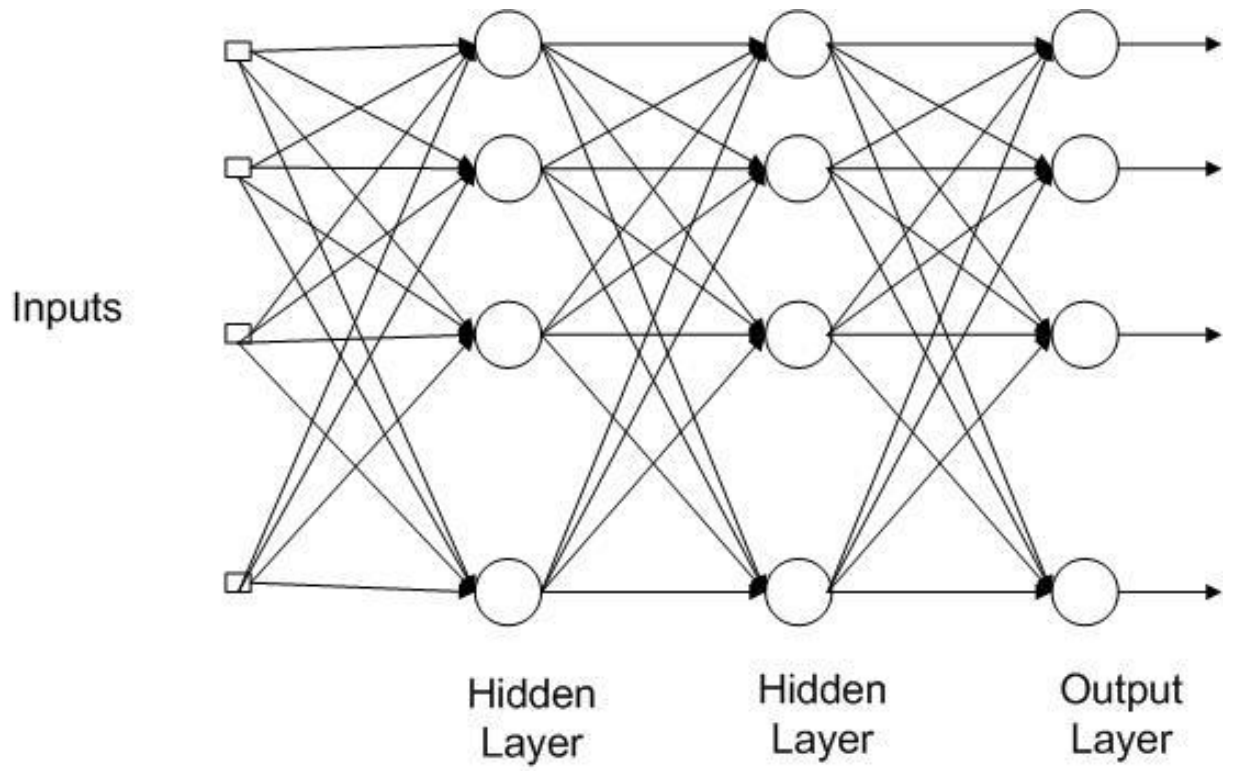

FIGURE A.10 A THREE-LAYER FULLY INTERCONNECTED NETWORK.

### A.2.4 Learning Algorithm

The learning algorithm in neural networks is the process of adjusting the weights of neurons (PEs), so the network can perform a specific task. The learning process is divided into two categories: Supervised learning and unsupervised learning. In supervised learning, the network is given both the input and the output, and it tries to adjust its weight in a way to minimize the error between its normal output and the desired output. In unsupervised learning, which is more difficult to implement, the network is only exposed to different inputs and it tries to extract the similarities between them.

There are different learning algorithms developed for neural networks. Those algorithms have been developed to be suitable to certain network topologies and to overcome the drawbacks of other algorithms. Most algorithms are based on the hill-climbing optimization or steepest descent concept, but as the problems tend to be more complexes, new concepts of learning were proposed. Many new algorithms are based on using Genetic Algorithm, Simulated Annealing and Random optimization [54]-[57].

### A.2.6 Back Propagation Network

Back-propagation [48] is the most widely algorithm used for classification. The back propagation network (BPN) [48], [42] is a layered, feed forward network that is fully interconnected by layers. Thus, there are no feedback connections and no connections that bypass one layer to go directly to a later layer. It employs the supervised mode of learning.

#### *A.2.6.1 BPN Operation*

The network learns a predefined set of input-output example pairs by using a two-phase propagate-adapt cycle. After an input pattern has been applied as a stimulus to the first layer of network units, it is propagated through each upper layer until an output is generated. This output is then compared to the desired output, and an error signal is computed for each output unit. The error signals are then transmitted backward from the output layer to each node in the intermediate layer contributing directly to the output. However, each unit in the intermediate layer receives only a portion of the total error signal, based roughly on the relative contribution of the unit made to the original output. This process repeats layer by layer, until each node in the network has received an error signal that describes its relative contribution to the total error. Based on the error signal received, connection weights are then updated by each unit to cause the network to converge toward a state that allows all the training patterns to be encoded.

The significance of this process is that, as the network trains, the nodes in the intermediate layers organize themselves such that different nodes learn to recognize different features of the total input space. After training, when presented with an arbitrary input pattern that is noisy or incomplete, the units in the hidden layers of the network will respond with an active output if the new input contains a pattern that resembles the feature, the individual units learned, to recognize during training.

Conversely, hidden-layer units have a tendency to inhibit their outputs if the input pattern does not contain the feature that they were trained to recognize.

As the signals propagate through the different layers in the network, the activity pattern presented at each upper layer can be thought of as a pattern with features that can be recognized by units in the subsequent layer. The output pattern generated can be thought of as a feature map that provides an indication of the presence or absence of many different feature combinations at the input. The total effect of this behaviour is that the BPN provides an effective means of allowing a computer system to examine data patterns that may be incomplete or noisy, and to recognize subtle patterns from the partial input [42].

### *A.2.6.2 The BPN Algorithm*

For the described network, the back propagation algorithm is summarized. For details refer to [42], [57]. Figure A.11 shows the multi-layer BPN architecture.

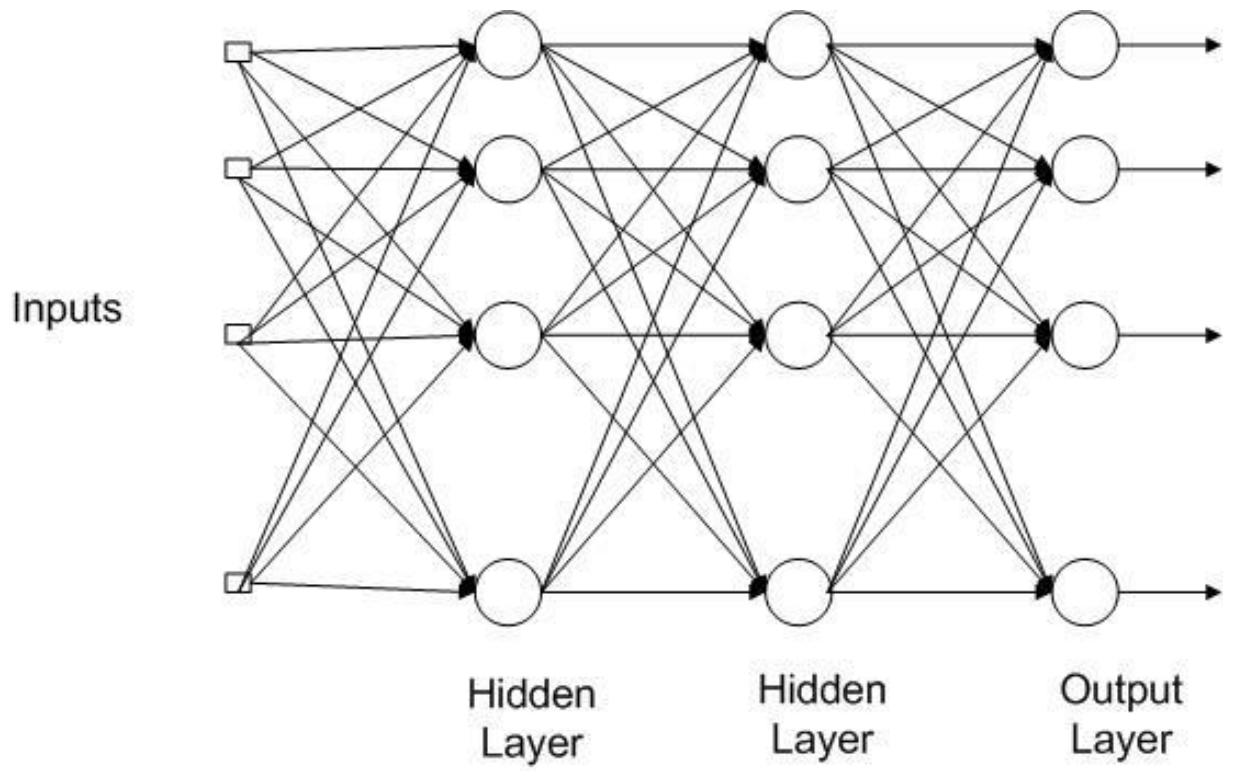

FIGURE A.11 3 LAYERS BACK-PROPAGATION NETWORK

Step 1 Apply the input vector $x_p = (x_{p1}, x_{p2}, \ldots, x_{pn})$ to the input units.
Step 2: Calculate the net input values to the hidden layer units:

$$net_{pj}^{h} = \sum_{i=1}^{n} w_{ji}^{h} x_{pi} + \theta_{j}^{h}$$

Step 3: Calculate the output from the hidden layer

$$i_{pj} = f_j^{h}(net_{pj}^{h})$$

Step 4: Move to the output layer, calculate the net-input values to each unit

$$net_{pj}^{o} = \sum_{j=1}^{L} w_{kj}^{o} i_{pj} + \theta_{k}^{o}$$

Step 5 Calculate the outputs:

$$o_{pk} = f_k^{o}(net_{pk}^{o})$$

Step 6 Calculate the gradient terms for the output units

$$\delta_{pk}^{o} = f_k^{`o}(net_{pk}^{o}).(y_{pk} - o_{pk})$$

Step 7: Calculate the gradient terms for the hidden units

$$\delta_{pj}^{h} = (f_j^{`h}(net_{pj}^{h}))\sum_{k} \delta_{pk}^{o} w_{kj}^{o}$$

Note that the error terms on the hidden units are calculated before the connection weights to the output-layer units have been updated.
Step 8: Update weights on the output layer:

$$w_{kj}^{o}(t+1) = w_{kj}^{o}(t) + \eta \delta_{pkj}^{o} i_{pj}$$

Step 9: Update weights of the hidden layer:

$$w_{ji}^{h}(t+1) = w_{ji}^{h}(t) + \eta \delta_{pj}^{h} x_{pj}$$

The order of the weight updates on an individual layer is not important.
Be sure to calculate the error term

$$E_p = \frac{1}{2}\sum_{k}(y_{pk} - o_{pk})^2$$

Since this quantity is the measure of how well the network is learning when the error is acceptably small for each of the training vector pairs training can be discontinued.

Back Propagation Neural Network Training Summary.

Where:
xp is the Input vector,

$net^h_{pj}$ is the net input to the $j^{th}$ hidden unit,

$w^h_{ji}$ is the weight on the connection from the $i^{th}$ input unit,

$\theta^h_j$ is the bias term,

The "h" superscript refers to quantities on the hidden layer.

$i_{pj}$ is the output of this node is,

$net^o_{pj}, w^o_{kj} i_{pj}, \theta^o_k$ and $o_{pk}$ have the same meaning but for the output layer,

where the "o" superscript refers to quantities on the output layer,

$e_{pk} = (y_{pk} - o_{pk})$ is the error at a single output unit,

where subscript "p" refers to the $p^{th}$ training vector.

and "k" refers to the $k^{th}$ output unit.

$y_{pk}$ is the desired output value,

and $o_{pk}$ is the actual output from the $k^{th}$ unit,

The factor $\eta$ is called the learning-rate parameter, $0 < \eta < 1$.

### *A2.6.3 Practical Considerations*

The back propagation algorithm is now widely recognized as a powerful tool for learning input-output mappings, with many applications in such areas as pattern recognition, time series forecasting, and control. However, all users of BPN have been confronted to two main drawbacks:

1. The slowness of the learning process, especially when large training sets, or large networks have to be used, and
2. The absence, so far, of any theoretical result, or heuristics, allowing for a reliable apriori determination of an optimal network architecture for a given task.

For the first point, the training may take long time and convergence behaviour may depend on:

1. The learning rate[59].
2. The number of hidden units [60] – [62].
3. The number of layers.
4. The nature of the training patterns.
5. Weights initialization [62] – [64].

# Appendix (B): Mathematical Background

## B.1 Interpolation

It happens occasionally that we need to estimate intermediate values between precise data points. The most common method used for this purpose is polynomial interpolation. The general form of $n^{th}$-order polynomial is

$$P_n(x) = a_0 + a_1 x + a_2 x^2 + ......+ a_n x^n \qquad \text{(B.1)}$$

For n+1 data points there is one and only one polynomial of order n passing through these points. For example there is only one straight line passing through 2 points and only one parabola passing thought 3 points and so on, see figure b.1

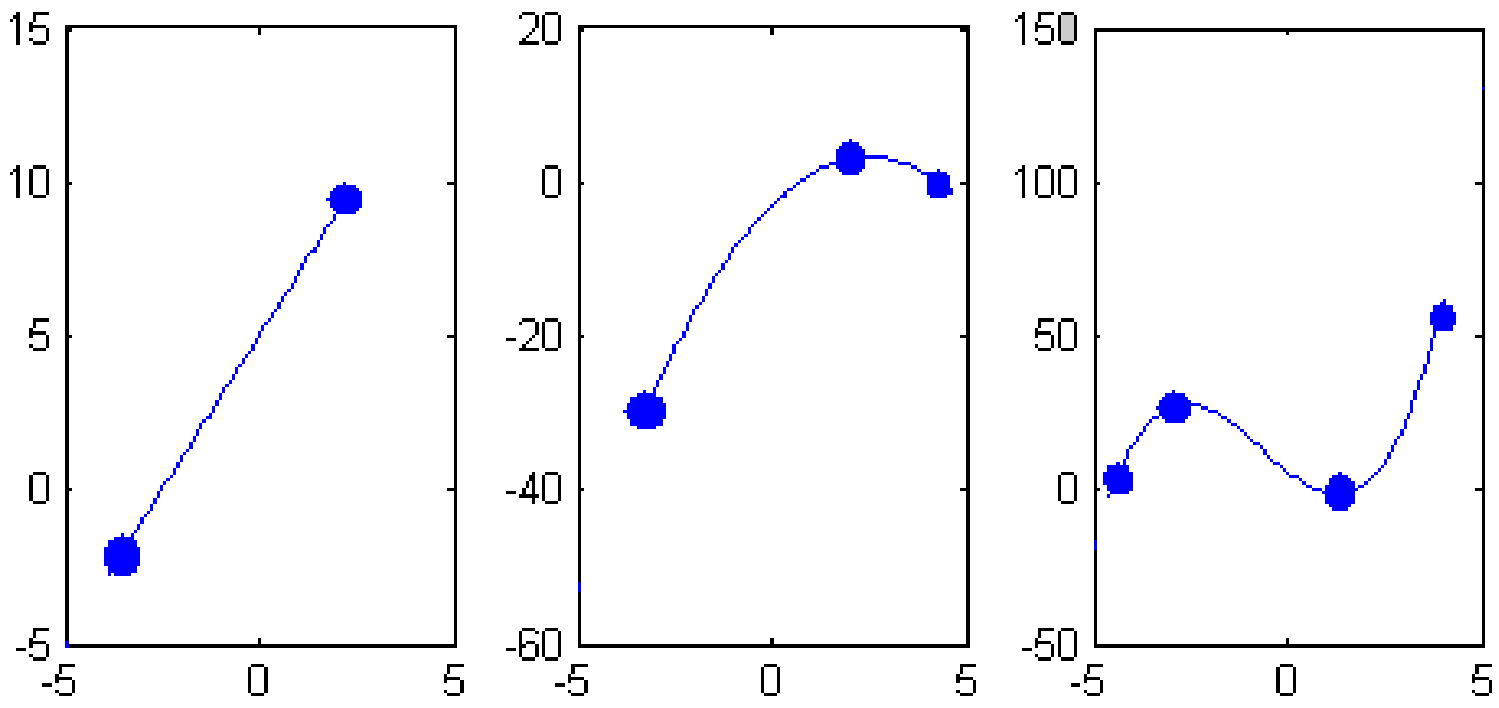

FIGURE B.1: POSSIBLE POLYNOMIAL PASSING THROUGH 2, 3, AND 4 POINTS

Polynomial interpolation consists of finding the unique $n^{th}$-order polynomial that fits the n+1 points. This polynomial provides a formula to find the intermediate points. The need for n+1 points comes from the fact that an $n^{th}$-order polynomial has n+1 unknowns, these are the coefficients of the successive power of x, $a_0$ through $a_n$ and so n+1 equations are needed and these equations can be found from the substitution of the n+1 data points.

### *b.1.1 Lagrange Interpolation*

The Lagrange interpolating polynomials are very straight forward to calculate. The method is characterized by the equations

$$f_n(x) = \sum_{i=0}^{n} L_i(x) f(x_i) \tag{B.2}$$

where

$$L_i(x) = \prod_{\substack{j=0 \\ j\neq i}}^{n} \frac{x - x_j}{x_i - x_j} \tag{B.3}$$

or if we combined them it becomes

$$f_n(x) = \sum_{i=0}^{n} \prod_{\substack{j=0 \\ j\neq i}}^{n} \frac{x - x_j}{x_i - x_j} f(x_i) \tag{B.4}$$

For example the linear version (n=1) is

$$f_1(x) = \frac{x - x_1}{x_0 - x_1} f(x_0) + \frac{x - x_0}{x_1 - x_0} f(x_1) \tag{B.5}$$

and the second order version is

$$\begin{aligned} f_2(x) = {} & \frac{(x - x_1)(x - x_2)}{(x_0 - x_1)(x_0 - x_2)} f(x_0) + \frac{(x - x_0)(x - x_2)}{(x_1 - x_0)(x_1 - x_2)} f(x_1) \\ & + \frac{(x - x_0)(x - x_1)}{(x_2 - x_0)(x_2 - x_1)} f(x_2) \end{aligned} \tag{B.6}$$

Notice that for equation b.6 each term is a second order curve that passes at specific point of the 3 points and is 0 at the 2 others, see figure b.2, and hence the summation of the 3 curves is the only second order curve passing through the 3 points

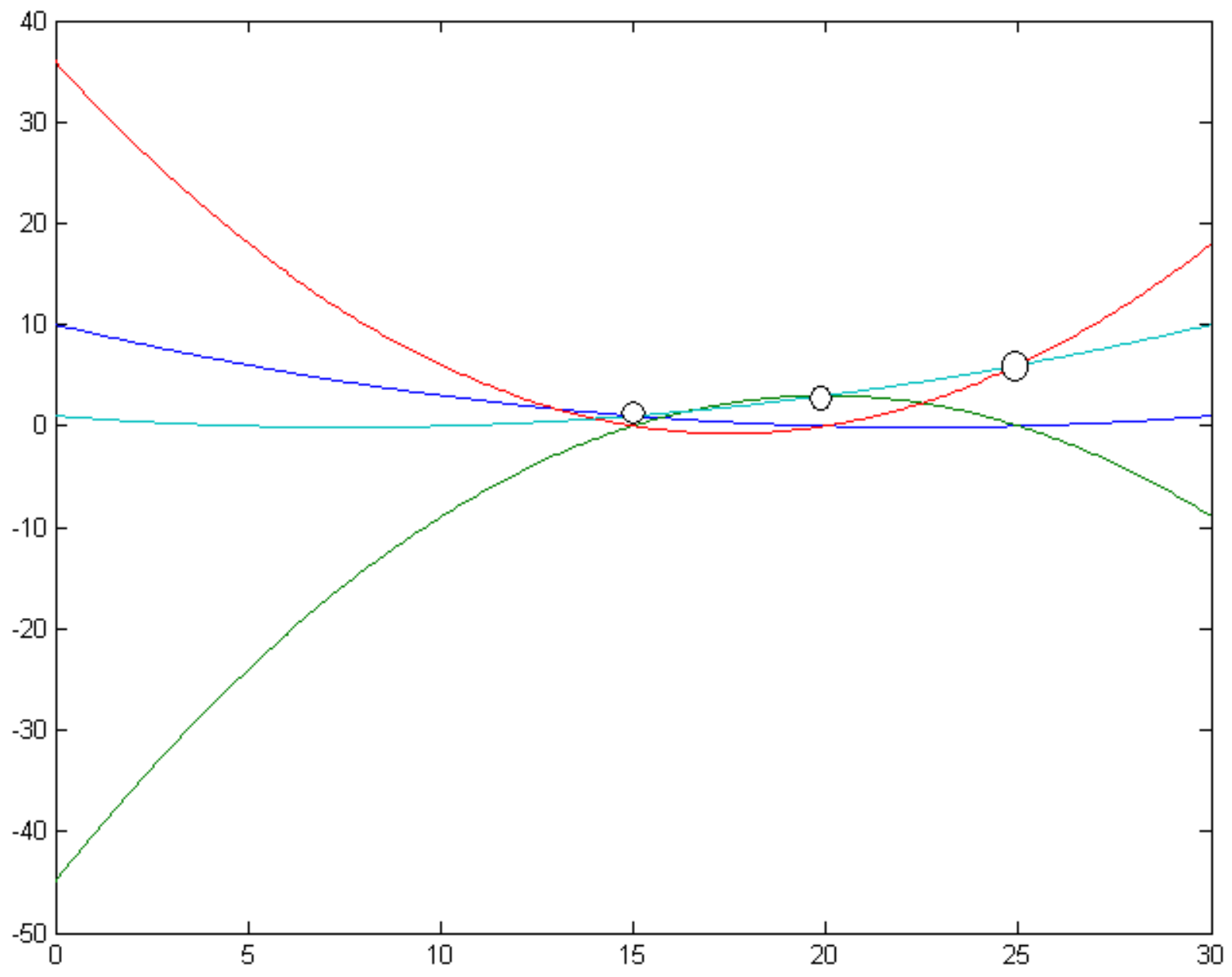

FIGURE B.2: CONSTRUCTION OF A SECOND ORDER CURVE USING 3 POINTS

### *b.1.2 Newton Interpolation*

A practically convenient for of the interpolating polynomial for n+1 data points, that is an n$^{th}$ order polynomial, is

$$f_n(x) = b_0 + b_1(x - x_0) + b_2(x - x_0)(x - x_1) + ..... + b_n(x - x_0)(x - x_1).....(x - x_{n-1}) \quad (B.7)$$

The above equation is equivalent equation b.1. n+1 data points are needed to calculate the coefficients $b_0, b_1, \ldots, b_n$. We use the following equations to evaluate the coefficients.

$$\begin{aligned} b_0 &= f(x_0) \\ b_1 &= f[x_1, x_0] \\ b_2 &= f[x_2, x_1, x_0] \\ &. \\ &. \\ b_n &= f[x_n, x_{n-1}, \ldots\ldots, x_1, x_0] \end{aligned} \tag{B.8}$$

Where the bracket function evaluation is called the divided difference. For example the first divided difference is

$$f[x_i, x_j] = \frac{f(x_i) - f(x_j)}{x_i - x_j} \tag{B.9}$$

and the second is

$$f[x_i, x_j, x_k] = \frac{f[x_i, x_j] - f[x_j, x_k]}{x_i - x_k} \tag{B.10}$$

and similarly

$$f[x_n, x_{n-1}, \ldots, x_1, x_0] = \frac{f[x_n, x_{n-1}, \ldots, x_1] - f[x_{n-1}, x_{n-2}, \ldots, x_0]}{x_n - x_0} \tag{B.11}$$

using these equations, yields the Newton divided difference interpolating polynomial that is

$$f(x) = f(x_0) + (x - x_0) f[x_1, x_0] + (x - x_0)(x - x_1) f[x_2, x_1, x_0] + ......$$
$$+ (x - x_0)(x - x_1).......(x - x_{n-2})(x - x_{n-1}) f[x_n, x_{n-1}, ......., x_1, x_0] \quad \text{(B.12)}$$

The equation b.12 seems a little bit complicated, but this complication can be reduced if a table is used to evaluate the differences.

***b.1.3 Evaluating Newton Equation using Horner's Method***

The Newton interpolation form was written as

$$f(x) = f(x_0) + (x - x_0) f[x_1, x_0] + (x - x_0)(x - x_1) f[x_2, x_1, x_0] + ......$$
$$+ (x - x_0)(x - x_1).......(x - x_{n-2})(x - x_{n-1}) f[x_n, x_{n-1}, ......., x_1, x_0] \quad \text{(B.13)}$$

Now let us consider rearranging the equation in the following form

$$\text{Let} \quad \begin{aligned} a_0 &= f(x_0) \\ a_1 &= f[x_1, x_0] \\ &. \\ &. \\ a_n &= f[x_n, x_{n-1}, ......, x_1, x_0] \end{aligned}$$

Found on the ladder described earlier. Then the Newton equation can be written as

$$f(x) = a_0 + a_1(x - x_0) + a_2(x - x_0)(x - x_1) + ......$$
$$+ a_n(x - x_0)(x - x_1)(x - x_2)....(x - x_{n-1}) \quad \text{(B.14)}$$

which can be written as

$$f(x) = (((((.....(a_n(x - x_{n-1}) + a_{n-1})(x + x_{n-2}) + a_{n-2}).....+ a_1)(x - x_0) + a_0 \quad \text{(B.15)}$$

Then the equation can be written as

$$\begin{aligned}
s_n &= a_n \\
s_{n-1} &= s_n(x - x_{n-1}) + a_{n-1} \\
&. \\
&. \\
s_1 &= s_2(x - x_1) + a_1 \\
s_0 &= s_1(x - x_0) + a_0, \\
f(x) &= s_0
\end{aligned} \tag{B.16}$$

Which is easier in calculation.

**B.2 Finite Fields**

Finite fields have become very important in cryptography. In the scope of this thesis, Rijndael cipher was illustrated and in this cipher, the core function design is based on properties of finite fields. In this section a brief introduction to finite fields is given. It is referred to [2], [65], and [66] for details.

**B.2.1 Groups, Rings, and Fields**

Before defining fields, groups and rings must be first defined.

A group is a set of element and a mathematical operation usually called addition, and + is used as a symbol, with the following properties.

- A 1. Closure under addition
- A 2. Associativity of addition
- A 3. Existence of additive identity
- A 4. Existence of additive inverse

Consider that three elements, a, b and c as members of the group, then for all a, b, and c

a+b results a member in the group (A 1)

a+(b+c)=(a+b)+c (A 2)

There exists a member called zero (0) such that

a+0=0+a=a (A 3)

For each element a in the group there exists another element –a such that

a+(-a)=(-a)+a=0 (A 4)

A group is called an Abelian if addition is commutative that is

a+b=b+a (A 5)

If we add another operation that is called multiplication, then with the following properties a ring is defined

A 6. Closure under multiplication

A 7. Associativity of multiplication

A 8. Distributive low

As previous consider a, b, c elements of a ring then for all a, b, and c

ab is a member of the ring (A 6)

a(bc)=(ab)c (A 7)

a(b+c)=ab+ac and (a+b)c=ac+bc (A 8)

A ring is commutative if

ab=ba (A9)

With theaddition of a multiplicative inverse and the no zero divisions, an Integral domain is constructed

There exists a member called (one) such that

a1=1a=a (A10)

if ab=0 then either a=0 or b =0

Finally a Field is an integral domain with the existence of a multiplicative inverse for all elements except zero

If a≠0 then there exists an element a-1 such that

$aa^{-1}=a^{-1}a=1$ (A11)

If the elements of the field are finite, then the field is called a finite field. There are lot of finite fields, but a class of them of importance to cryptography are the Galois Fields named after Evariste Galois and noted as GF. Galois Fields are finite fields of integers that all the mathematical operations mentioned above are done in a modular fashion.

### B.2.2 Modular Arithmetic

First let us introduce modular arithmetic

Given integers a, b, $n \neq 0$, a is congruent to b modulo n. written as $a \equiv_n b$, if and only if a-b=kn for some integer k.

For example $19 \equiv_4 3$ as 19-3=16=4*4.

For any integer $n \neq 0$, all the element {0,1,……,n-1} are called the set of residues modulo n.

Also $a \equiv_n b$ if and only if a mod n = b mod n

Integers modulo n with addition and multiplication forms a commutative ring as described earlier.

The addition and multiplication mathematical operations can be done modulo n by applying the normal operation and reducing the result by taking the remainder of dividing by n.

#### *B2.2.1 .Divisors*

A non-zero number b divides a if for some m have a=mb where a,b, and m are all integers. Or in another words, b divides into a with no remainder. This is denoted as b|a and pronounced b divides a or b is a divisor of a. For example 1,2,3,4,6,8,12,24 divide 24

The set of all numbers that divides an integer is called the set of divisors so for the previous example {1,2,3,4,6,8,12,24} is the set of divisors of 24.

Modular Arithmetic is some what a 'clock arithmetic' which uses a finite number of values, and loops back from either end.

As mentioned before, modular arithmetic can be done in the normal way and then reduce answer modulo the base. The reduction can be done in any point of the operation, ie

a+b mod n = [a mod n + b mod n] mod n

When reducing, we "usually" want to find the **positive** remainder after dividing by the modulus. For positive numbers, this is simply the normal remainder. For negative numbers we have to "overshoot" (i.e. find the next multiple larger than the number) and "come back" (i.e. add a positive remainder to get the number); rather than have a "negative remainder".

Define the set $Z_n$ to be the set of integers less than n where n is a positive integer that is $Z_n$= {0, 1…, (n-1)} the following properties hold

(a+b) mod n= (b+a) mod n

(a X b) mod n= (b X a) mod n

[(a+b)+c] mod n= [a+(b+c)] mod n

[(a X b) X c] mod n= [a X (b X c)] mod n

[a X (b+c)] mod n= [(a X b) + (a X c)] mod n

(a+0) mod n= a mod n

(a X 1) mod n= a mpd n

For a $\in Z_n$ there exists and element b $\in Z_n$ such that

a + b mod n= 0 mod n

Table B.1 shows the addition operation mod 6 and table B.2 shows the multiplication operation mod 6 while table B.3 shows the additive inverses and the ***available*** multiplicative inverses.

| + | 0 | 1 | 2 | 3 | 4 | 5 |
|---|---|---|---|---|---|---|
| 0 | 0 | 1 | 2 | 3 | 4 | 5 |
| 1 | 1 | 2 | 3 | 4 | 5 | 0 |
| 2 | 2 | 3 | 4 | 5 | 0 | 1 |
| 3 | 3 | 4 | 5 | 0 | 1 | 2 |
| 4 | 4 | 5 | 0 | 1 | 2 | 3 |
| 5 | 5 | 0 | 1 | 2 | 3 | 4 |

**Table B.1 Addition modulo 6**

| X | 0 | 1 | 2 | 3 | 4 | 5 |
|---|---|---|---|---|---|---|
| 0 | 0 | 0 | 0 | 0 | 0 | 0 |
| 1 | 0 | 1 | 2 | 3 | 4 | 5 |
| 2 | 0 | 2 | 4 | 0 | 2 | 4 |
| 3 | 0 | 3 | 0 | 3 | 0 | 3 |
| 4 | 0 | 4 | 2 | 0 | 4 | 2 |
| 5 | 0 | 5 | 4 | 3 | 2 | 1 |

**Table B.2 Multiplication modulo 6**

| a | -a | $a^{-1}$ |
|---|---|---|
| 0 | 0 | - |
| 1 | 5 | 1 |
| 2 | 4 | - |
| 3 | 3 | - |
| 4 | 2 | - |
| 5 | 1 | 5 |

**Table B.3 Inverses modulo 6**

### B.2.3 Greatest Common Divisors and Inverses

Modular arithmetic as shown above on finite integers modulo n is not a field in general as multiplicative inverse is not always available. It can be shown that a multiplicative inverse of a number a modulo nm exists if and only if a and n are relatively prime, that is if they exist no number that divides both of a and n.

If an integer divides two other integers, it is called a common divisor. The largest divisor of two integers is called the grates common divisor of the two numbers. The greatest common divisor of two numbers can be found based on the Euclidian theory that is

GCD(a,b) = GCD(b, a mod b)

So an algorithm can be constructed to calculate the greatest common divisor of two numbers as follows,

```
GCD(a,b)
begin
      A=a, B=b
      while B>0 do
      begin
            R = A mod B
            A = B, B = R
      end
      return A
end
```

Euclid's Algorithm is derived from the observation that if a and b have a common factor d (i.e. a=m.d and b=n.d) then d is also a factor in any difference between them, a-p.b = (m.d)-p.(n.d) = d.(m-p.n). Euclid's Algorithm keeps computing successive differences until it vanishes, at which point that divisor has been reached. Following is an example of applying the algorithm

Find GCD(1512,797)
GCD(1512,797)=GCD(797,1512%797)=GCD(797,715)
GCD(797,715)=GCD(715,797%715)=GCD(715,82)
GCD(715,82)=GCD(82,715%82)=GCD(82,59)
GCD(82,59)=GCD(59,82%59)=GCD(59,23)
GCD(59,23)=GCD(23,59%23)=GCD(23,13)
GCD(23,13)=GCD(13,23%13)=GCD(13,10)
GCD(13,10)=GCD(10,13%10)=GCD(10,3)
GCD(10,3)=GCD(3,10%3)=GCD(3,1)
GCD(3,1)=GCD(1,3%1)=GCD(1,0)=1
Then GCD(1512,797)=1.

An extension to the Euclidean algorithm can be performed in order to calculate the inverse of a modulo n. The result of the algorithm is $a^{-1}$ modulo n if GCD(a,n)=1.

The extended Euclidean algorithm is as follows

```
EXTENDED EUCLID(n, a)      (a^-1 modulo n)
(A1, A2, A3)=(1, 0, n);
(B1, B2, B3)=(0, 1, a)
while (B3≠0 and B3≠1)
begin
      Q = A3 div B3
      (T1, T2, T3)=(A1 – Q B1, A2 – Q B2, A3 – Q B3)
      (A1, A2, A3)=(B1, B2, B3)
      (B1, B2, B3)=(T1, T2, T3)
end
if B3 = 0
      return A3 = gcd(n, a); no inverse
if B3 = 1
      return B3 = gcd(n, a); B2 = a^-1 mod n
```

Following is an example,

Find $550^{-1}$ modulo 1759

| Q | A1 | A2 | A3 | B1 | B2 | B3 |
|---|---|---|---|---|---|---|
| - | 1 | 0 | 1759 | 0 | 1 | 550 |
| 3 | 0 | 1 | 550 | 1 | -3 | 109 |
| 5 | 1 | -3 | 109 | -5 | 16 | 5 |
| 21 | -5 | 16 | 5 | 106 | 339 | 4 |
| 1 | 106 | -339 | 4 | -111 | 355 | 1 |

Then $550^{-1}$ modulo 1759 = 355

### B.2.4 Galois Fields

From the previous discussion, $Z_n$ can be a field if all elements of ring have a multiplicative inverse. This is only possible if n is a prime. In that case $Z_p$ where p is prime is a field and is called Galois Field of P or GF(P).

Also Galois proposed a field on the form of GF($p^n$) where p is also a prime and n is an integer can also be a field if the elements of the field are treated as polynomials not as integers. Where each element is presented as a successive power summation of elements in Zp and n-1 denotes the largest degree a polynomial can have. That is each number can be a polynomial in the form

$f(x) = a_m x^m + a_{m-1} x^{m-1} + .... + a_1 x + a_0$ where $m \le n-1$ and $a_i \in Zp$ for all 0<i<m

In the case of GF(p) where n=1, normal modular arithmetic can be performed and p is the modulus.

But in GF($p^n$) and n>1, another modulus has to be found. In this case an irreducible polynomial of degree n is used as the modulus. That is a polynomial that can not be factorized into a product of lower degree polynomials. For example $X^8+1$ is an irreducible polynomial of degree 8.

Polynomial arithmetic is instead used and all the results are reduce modulo the irreducible polynomial. Euclidean algorithm and extended Euclidean algorithm are applicable in the case of polynomials. The major field of interest in cryptography is the GF($2^n$) as binary numbers, normal representation of digital data inside a computer, can be interpreted as polynomials in this field.

Consider the following example

Get $\{15\}^{-1}$ in GF($2^8$) with the irreducible polynomial $x^8+x^4+x^3+x+1$

Here the number 15 is considered as a polynomial in the field GF($2^8$) that is a polynomial of successive power summation of elements in $Z_2$. This can be simply achieved by getting the binary representation of 15 as an 8 bits number, that is 00001111. This presentation can be converted into a polynomial for as $x^3+x^2+x+1$. Now using extended Euclidean algorithm

| Q | A1 | A2 | A3 | B1 | B2 | B3 |
|---|---|---|---|---|---|---|
| - | 1 | 0 | $x^8+x^4+x^3+x+1$ | 0 | 1 | $x^3+x^2+x+1$ |
| $x^5+x^4+1$ | 0 | 1 | $x^3+x^2+x+1$ | 1 | $x^5+x^4+1$ | $x^2$ |
| x+1 | 1 | $x^5+x^4+1$ | $x^2$ | x+1 | $x^6+x^4+x$ | x+1 |
| x+1 | x+1 | $x^6+x^4+x$ | x+1 | $x^2$ | $x^7+x^6+x^2+x+1$ | 1 |

Then the multiplicative inverse is {11000111}={199}

To show how the first Q is calculated

Consider the following long division of polynomials

| | | | |
|---|---|---|---|
| $x^3+x^2+x+1$ | $x^8+\qquad x^4+x^3+\quad x+1$ | | |
| $x^5+x^4+1$ | $\underline{x^8+x^7+x^6+x^5}$ | (subtracting in $Z_2$) | |
| | $x^7+x^6+x^5+x^4+x^3+\quad x+1$ | | |
| | $\underline{x^7+x^6+x^5+x^4}$ | (subtracting in $Z_2$) | |
| | $x^3+\quad x+1$ | | |
| | $\underline{x^3+x^2+x+1}$ | (subtracting in $Z_2$) | |
| | $x^2$ | | |

So the result of division is $x^5+x^4+1$ while the remainder is $x^2$

# List of publications

# مستخلص

**أيمن محمد بهاء الدين صادق البصال**

**النظم الذكية في أمان المعلومات**

**رسالة دكتوراة**

تهدف الرسالة لاستخدام النظم الذكية لتوسيع وتحسين الأداء والأمان لطرق التشفير الحالية والتي تعتبر المكون الرئيسي في تحقيق أمان المعلومات.

تبدأ الرسالة بدراسة للأهداف المطلوب تحقيقها للوصول لأمان المعلومات. ثم يتبع ذلك شرح مفصل لنظم التشفير المتماثلة حيث يتم عرض مبادئ التصميم وكيفية قياس مدى أمن نظام تشفير معين. بعد ذلك تعرض الرسالة لعدد من الأمثلة لنظم التشفير الحديثة الأكثر استخداما.

تعرض الرسالة بعد ذلك طرق مهاجمة نظم التشفير وكيفية تصنيفها ثم يتبع ذلك شرح تفصيلي لنظم الهجوم التفارقية ونظم الهجوم البينية.

ثم تعرض الرسالة لإطار استخدام الخوارزميات الجينية كوسيلة لمهاجمة نظم التشفير المتماثلة.

تم اقتراح توسيع جديد لنظم الهجوم التفارقية باستخدام الخوارزميات الجينية وتم اقتراح استخدام الخواص التفارقية المحتملة لنظام التشفير كأداة لقياس الملائمة. تم بعد ذلك تنفيذ الهجوم المقترح على النماذج الأساسية لنظم التشفير المتماثلة الحديثة بدلا من تنفيذها على نظام تشفير بعينه وذلك لإثبات إمكانية تنفيذ التقنية المقترحة على أي نظام تشفير معرض للهجوم بالنظم التفارقية.

تم اقتراح نظام هجوم جديد على نظم التشفير المتماثلة يعتمد على قدرة الشبكات العصبية على تقريب الانتقال من مدى لمجال مقابل محكوما بدالة محددة. وقد تم تنفيذ الهجوم وتجربته على نظام تشفير مقترح لا يمكن حله باستخدام النظم التفارقية أو الخطية.

تم بعد ذلك اقتراح نظام تشفير جديد يعتمد على الشبكات العصبية كمحور لنظام التشفير وذلك لقدرتها عل القيام بتحقيق مقابلة بين مدخلات ومخرجات كبيرة الحجم بأداء مرتفع وباستخدام مساحة صغيرة من الذاكرة مقارنة بطريقة جداول التعويض. وقد تم اقتراح نظام لتوليد المفاتيح الفرعية من المفتاح الأصلي كجزء من نظام التشفير الكامل. وقد تم عرض اقتراحين لتنفيذ شبكة عصبية تعتمد على مفتاح التشفير لزيادة أمان النظام.

هذا وقد تم تدعيم المقترحات المقدمة بواسطة عدد كبير من التجارب وتم عرض النتائج كل في موقعه

جامعة عين شمس
كلية الهندسة

**صفحة العنوان**

| | | |
|---|---|---|
| **اسم الباحث** | **:** | **أيمن محمد بهاء الدين صادق البصال** |
| **اسم الدرجة** | **:** | **دكتوراة الفلسفة** |
| **القسم التابع له** | **:** | **هندسة الحاسبات والنظم** |
| **اسم الكلية** | **:** | **الهندسة** |
| **سنة التخرج** | **:** | **1995** |
| **سنة المنح** | **:** | **2004** |

جامعة عين شمس
كلية الهندسة

**تعريف بمقدم الرسالة**

| | | |
|---|---|---|
| **اسم الباحث** | : | **أيمن محمد بهاء الدين صادق البصال** |
| **تاريخ الميلاد** | : | **1972/10/16** |
| **محل الميلاد** | : | **القاهرة** |
| **الدرجة العلمية الأولى** | : | **البكالوريوس** |
| **الجهة المانحة لها** | : | **كلية الهندسة جامعة عين شمس** |
| **تاريخ المنح** | : | **1995** |
| **الوظيفة الحالية** | : | **مدرس مساعد بقسم هندسة الحاسبات والنظم- كلية الهندسة – جامعة عين شمس** |
| **اسم مقدم البحث** | : | **أيمن محمد بهاء الدين صادق البصال** |
| **التوقيع** | : | |
| **التاريخ** | : | |

جامعة عين شمس

كلية الهندسة

قسم هندسة الحاسبات والنظم

**رسالة دكتوراة**

**اسم الباحث : أيمن محمد بهاء الدين صادق البصال**

**عنوان الرسالة : النظم الذكية في أمان المعلومات**

**الدرجة : دكتوراة الفلسفة**

**لجنة الإشراف**

| **الاسم** | **الوظيفة** |
| --- | --- |
| **أ.د./ عبدالمنعم عبدالظاهر وهدان** | أستاذ بقسم هندسة الحاسبات والنظم – كلية الهندسة – جامعة عين شمس |

**تاريخ البحـــث: / /**

**الدراسات العليا**

| **ختم الإجازة** | **أجيزت الرسالة بتاريخ** |
| --- | --- |
| | **/ /** |

| **موافقة مجلس الكلية** | **موافقة مجلس الجامعة** |
| --- | --- |
| **/ /** | **/ /** |

# شـــكر

أولا وقبل كل شئ أحمد الله العظيم على عونه وتوفيقه

ثم أود أن أعبر عن عميق امتناني وشكري لأستاذي أ.د. عبد المنعم وهدان وذلك لاقتراحه نقطة البحث ونصائحه الغالية ومجهوده العظيم أثناء إعداد ومراجعة الرسالة. ولا يمكنني أن أعبر عن شكري له على رعايته واهتمامه بي منذ أن كنت طالبا بالقسم ولا يمكن أن أنسى أنا أو زملائي توجيهاته المستمرة لنا. لقد كان خلال السنوات التي عملت بها معيدا ثم مدرسا مساعدا بقسم هندسة الحاسبات والنظم مثلا أعلى وقدوة يتمثل فيه المعنى الحقيقي لأستاذ الجامعة. شكرا لمجهوده المتواصل ومن قبل أثناء إشرافه على رسالتي للماجستير فلقد أعطاني الفرصة كاملة لإظهار إمكانياتي ووضع أولى خطواتي في طريق البحث العلمي. كل ما أتمناه أن أستطيع أن أتبع خطاه.

كما أود أن أشكر أ.د. محمد أديب رياض الغنيمي لمجهوداته العظيمة أثناء إعداد الرسالة منذ أن كانت مجرد أفكار واقتراحات. أود أن أشكره على دعمه ومساعدته وإعطائي المواد العلمية والمراجع الهامة.

كما أود أن أشكر أساتذتي من قسم هندسة الحاسبات والنظم أولا على كل ما علموني فالفضل يعود إليهم فيما تعلمته ثم على رعايتهم و دعمهم وكذلك جميع زملائي على دعمهم وتشجيعهم

كما أود أن أشكر والدي ووالدتي ووالدة زوجتي وجميع أهلي على دعائهم وتشجيعهم أثناء إعداد الرسالة.

أخيرا أود أن أشكر زوجتي الحبيبة على صبرها و دعمها الكامل وتشجيعها في أهم مراحل إعداد الرسالة

**كلية الهندسة**

**قسم هندسة الحاسبات والنظم**

---

**اسم الباحث : أيمن محمد بهاء الدين صادق البصال**

**عنوان الرسالة : النظم الذكية في أمان المعلومات**

**الدرجة : دكتوراة الفلسفة**

<u>لجنة الحكم</u>

**الاسم : (مناقشا/مشرفا)**

**الوظيفة : ...............**

تاريخ المناقشة: / /

كلية الهندسة

قسم هندسة الحاسبات والنظم

# *النظم الذكية في أمان المعلومات*

مقدمة من

## أيمن محمد بهاء الدين صادق البصال

ماجستير الهندسة الكهربية
(هندسة الحاسبات والنظم)
جامعة عين شمس – **1999**

رسالة
مقدمة للحصول على درجة دكتوراة الفلسفة في الهندسة الكهربية
(هندسة الحاسبات والنظم)

تحت إشراف

**الأستاذ الدكتور./ عبد المنعم عبد الظاهر وهدان**

القاهرة – مصر
**2004**